\documentclass{article}
\usepackage{ijcai21}

\usepackage[T1]{fontenc}    
\usepackage{algorithm, algorithmic}
\usepackage{amsmath, amsthm, amsfonts, amssymb}
\usepackage{array}
\usepackage{booktabs}
\usepackage{enumitem}
\usepackage{flushend}
\usepackage{graphicx}
\usepackage{microtype}
\usepackage{multirow}
\usepackage{nicefrac}       
\usepackage{printlen}
\usepackage{soul}
\usepackage{svg}
\usepackage{url}
\usepackage{wrapfig}
\usepackage{lipsum}  

\newcommand{\com}[1]{\iffalse #1 \fi}%

\newcommand{\noimage}{%
  \setlength{\fboxsep}{-\fboxrule}%
  \fbox{\phantom{\rule{100pt}{100pt}}File missing\phantom{\rule{100pt}{100pt}}}
}
\let\includegraphicsoriginal\includegraphics
\renewcommand{\includegraphics}[2][width=\textwidth]{\IfFileExists{#2}{\includegraphicsoriginal[#1]{#2}}{\noimage}}

\emergencystretch=\maxdimen
\newcounter{descriptcount}

\newcolumntype{H}{>{\setbox0=\hbox\bgroup}c<{\egroup}@{}}

\def\eqref#1{equation~\ref{#1}}

\def\1{\bm{1}}

\DeclareMathAlphabet{\mathsfit}{\encodingdefault}{\sfdefault}{m}{sl}
\SetMathAlphabet{\mathsfit}{bold}{\encodingdefault}{\sfdefault}{bx}{n}

\usepackage{times}
\usepackage{soul}
\usepackage{url}
\usepackage[hidelinks]{hyperref}
\usepackage[utf8]{inputenc}
\usepackage[small]{caption}
\usepackage{graphicx}
\usepackage{amsmath}
\usepackage{amsthm}
\usepackage{booktabs}
\usepackage{algorithm}
\usepackage{algorithmic}
\newcommand\blfootnote[1]{%
  \begingroup
  \renewcommand\thefootnote{}\footnote{#1}%
  \addtocounter{footnote}{-1}%
  \endgroup
}

\title{Deep neural network loses attention to adversarial images}

\author{
Shashank Kotyan
\and
Danilo Vasconcellos Vargas 
\affiliations
Kyushu University\\
\emails
vargas@inf.kyushu-u.ac.jp
}

\begin{document}

\maketitle

\begin{abstract}

Adversarial algorithms have shown to be effective against neural networks for a variety of tasks.
Some adversarial algorithms perturb all the pixels in the image minimally for the image classification task in image classification. 
In contrast, some algorithms perturb few pixels strongly. 
However, very little information is available regarding why these adversarial samples so diverse from each other exist. 
Recently, \cite{vargas2019understanding} showed that the existence of these adversarial samples might be due to conflicting saliency within the neural network. 
We test this hypothesis of conflicting saliency by analysing the Saliency Maps (SM) and Gradient-weighted Class Activation Maps (Grad-CAM) of original and few different types of adversarial samples. 
We also analyse how different adversarial samples distort the attention of the neural network compared to original samples. 
We show that in the case of Pixel Attack, perturbed pixels either calls the network attention to themselves or divert the attention from them.
Simultaneously, the Projected Gradient Descent Attack perturbs pixels so that intermediate layers inside the neural network lose attention for the correct class. 
We also show that both attacks affect the saliency map and activation maps differently. 
Thus, shedding light on why some defences successful against some attacks remain vulnerable against other attacks.
We hope that this analysis will improve understanding of the existence and the effect of adversarial samples and enable the community to develop more robust neural networks.
\blfootnote{Copyright © 2021 for this paper by its authors. Use permitted under Creative Commons License Attribution 4.0 International (CC BY 4.0).}

\end{abstract}

\section{Introduction}

\begin{figure}[!t]
    \centering
    \includegraphics[width=\columnwidth]{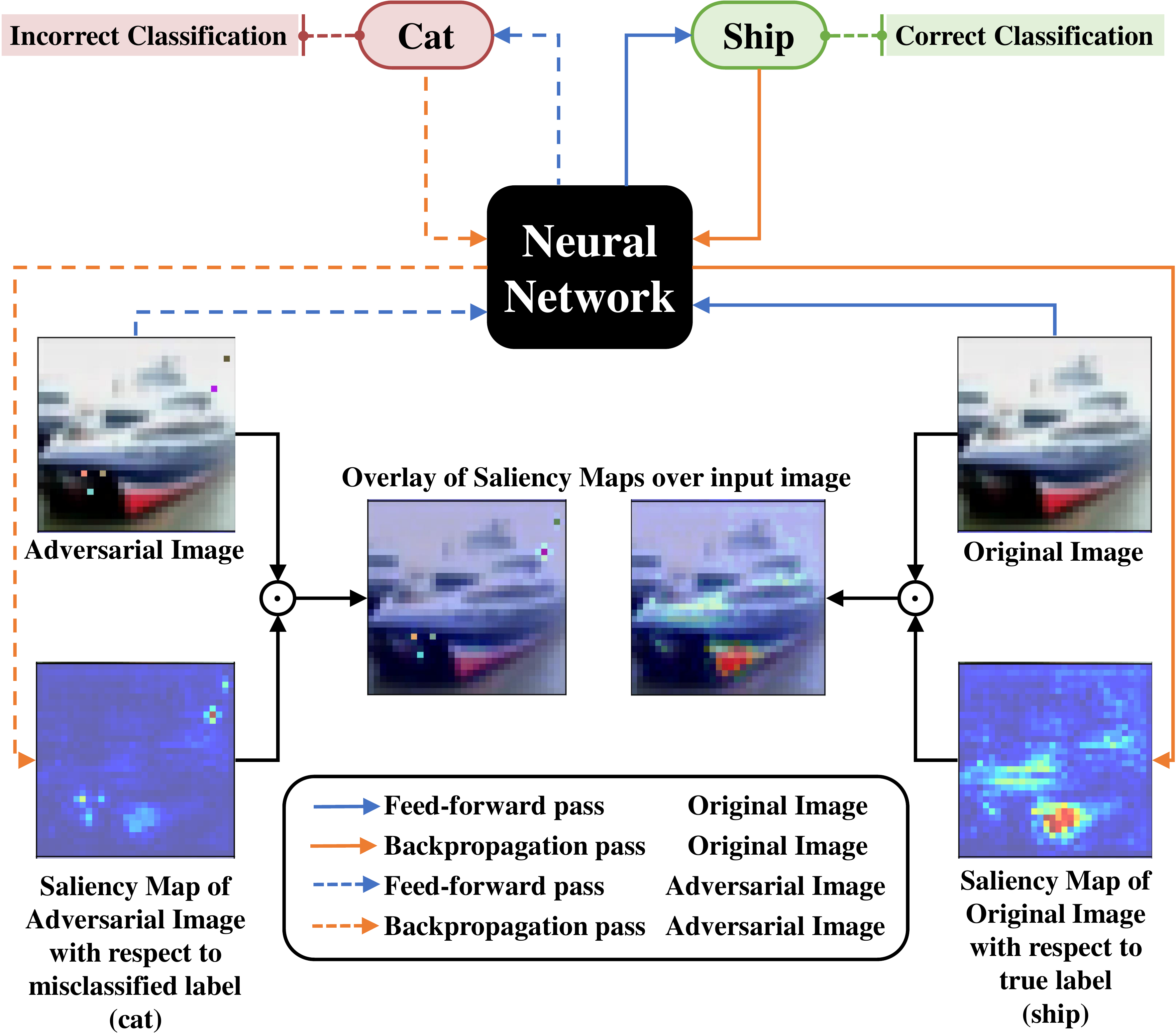}
    \caption{Illustration of the methodology to generate Saliency Maps with respect to the predicted class for original and adversarial images.}
    \label{overview} 
\end{figure}

Since adversarial samples were discovered some years ago in \cite{szegedy2014intriguing} for neural networks, the variety of adversarial samples and corresponding adversarial algorithms has grown in both number and types.
From Universal perturbations, \cite{moosavi2017universal} that can be added to almost any image to generate an adversarial sample, to the addition of crafted patches \cite{brown2017adversarial} or in fact, even the addition of one-pixel \cite{su2019one} was also shown to cause networks to be enough to misclassify. 
Some approaches rely on detecting adversarial samples to mitigate the adverse effects of adversarial algorithms, while some approaches rely on defensive algorithms.
However, most defences rely on obfuscating gradients \cite{athalye2018obfuscated} which can be broken by black-box and stronger attacks.


It was shown in \cite{vargas2019understanding} that changes in pixels of an image propagate and expand throughout the layers to either disappear or cause significant changes in the classification.
Additionally, it was also shown that perturbation in nearby pixels of successful one-pixel attack has high attack accuracy. 
This suggests that changes in a pixel may increase or decrease the influence of a receptive field (small group of nearby pixels). 
This is a direct relationship of the convolution, which is a linear operation.

In the adversarial setting, the analysis of the spatial distribution of saliency proves helpful to interpret why changing some pixels \cite{su2019one} in the network corresponds to misclassification. 
It was hypothesised in \cite{vargas2019understanding} that the existence of adversarial samples is due to conflicting saliency, which causes enough disturbance in the neural network forcing it to misclassify. 
Hence, adversarial samples are not naively fooling neural networks but diverting their attention towards another part of the image. 

\textbf{Contributions:} 
In this article, we analyse how these spatial distribution changes based on perturbed pixels.
Figure \ref{overview} shows the illustration of the methodology of generating saliency maps and the difference in saliency caused by adversarial perturbations.
We analyse the spatial distribution changes caused by perturbing pixels by evaluating Saliency Maps and Gradient-weighted Class Activation Maps. 
We use ResNet trained on CIFAR-10 for our evaluation. 
We also assess the effect for adversarial samples generated by Pixel Attack (an $L_0$ norm black box attack) \cite{su2019one}, and Projected Gradient Descent Attack (an $L_\infty$ norm white-box attack) \cite{madry2018towards}.

We evaluate the Saliency Maps and Gradient-weighted Class Activation Maps concerning the predicted class, the true class and the misclassified class of both original and adversarial images.
Our experiments reveal that both Pixel Attack and Projected Gradient Descent Attack distorts the saliency maps and activation maps differently. 
Where the Pixel Attack calls the network's attention towards the perturbed pixels, effectively changing the region of interest for the neural networks. 
The Projected Gradient Descent Attack diffuses the image's saliency and effectively destroys the class activation maps for true class for the adversarial images.

We also investigate the hypothesis that adversarial samples exist due to a conflicting image saliency raised in \cite{vargas2019understanding}.
Our experiments demonstrate that pixels define both the strength as well as the type of feature (i.e., changing pixels change the relationship between pixels of a receptive field and, therefore, the feature).
Therefore, it is possible to either destroy a feature or decrease its intensity by modifying the pixels slightly. 
It follows that the same should be valid for modifying pixels slightly all over the image (in this case, however, it is natural that the relative modification is more important).

\section{Related Works}

\subsection{Adversarial Machine Learning}

    It was exhibited in \cite{szegedy2014intriguing} that neural networks behave oddly for almost the same images.
    Afterwards, in \cite{nguyen2015deep}, the authors demonstrated that neural networks show high confidence when presented with textures and random noise.
    This led to discovering a series of vulnerabilities in neural networks, which were then exploited by adversarial attacks.
    
    Defensive distillation \cite{papernot2016distillation}, a defence was proposed, in which a smaller neural network squeezes the content learned by the original one was proposed as a defence. 
    However, it was shown not to be robust enough in \cite{carlini2017towards}.
    Adversarial training was also proposed, in which adversarial samples are used to augment the training dataset \cite{goodfellow2014explaining}, \cite{madry2018towards}. 
    Augmentation of the dataset is done so that the neural network should classify the adversarial samples, thus increasing their robustness.  
    Although adversarial training can increase the robustness slightly, the resulting neural network is still vulnerable to attacks \cite{tramer2018ensemble}.
    
    Regarding understanding the phenomenon, it is argued in \cite{goodfellow2014explaining} that neural networks' linearity is one of the main reasons.
    Another investigation proposes the conflicting saliency added by adversarial samples as the reason for misclassification \cite{vargas2019understanding}.
    A geometric perspective is analysed in \cite{moosavi2018robustness}, where it is shown that adversarial samples lie in shared subspace, along which the decision boundary of a classifier is positively curved. 
    Further, in \cite{fawzi2018empirical}, a relationship between sensitivity to additive perturbations of the inputs and the curvature of the decision boundary of deep networks is shown.
    Another aspect of robustness is discussed in \cite{madry2018towards}, where authors suggest that the capacity of the neural networks' architecture is relevant to the robustness.

\subsection{Visualisation and understanding of neural networks}

    We can visualise the attention of the neural networks to understand which part of the image neural network focuses.
    It is well known that some parts of the image affect the output more than others; in fact, the most influential areas can be seen by plotting the saliency maps of a given sample. 
    
    In \cite{itti1998model} saliency maps are defined as maps to represent conspicuity at every location in the visual field by a scalar quantity. 
    These saliency maps also help guide the selection of attended locations based on the spatial distribution of saliency. 
    Therefore, saliency maps contribute to finding the pixels of the image, which contribute more than the other pixels based on the spatial distribution of saliency to categorise the image in neural networks.

    There are two variations of the backpropagation used in saliency maps in the literature.
    One is Rectified/ Deconv Backpropagation \cite{zeiler2014visualizing}, where only positive gradient information is propagated through the layers, which correspond to the increase in output. 
    An increase in output can be interpreted as an increase in output probability for a class in a classification problem.
    Another is Guided Backpropagation \cite{springenberg2014striving}, where only positive gradient information is propagated through the layers, which have positive activation for the layers.

\section{Adversarial Machine Learning and Saliency Maps}

Let us suppose that for the image classification problem, ~~~~~~~~~~~~~$x \in \mathbb{R}^{m \times n \times c}$ be the image which is to be classified. 
Here $m, n$ is the width and the height of the image, and $c$ is the number of colour channels. 
A neural network comprises several neural layers composed of a set of perceptrons (artificial neurons) linked together.
Each of these perceptrons maps a set of inputs to output values with an activation function. 

Thus, function of the neural network (formed by a chain) can be defined as:
\begin{equation}
\begin{aligned}
g(x) = f^{(k)}( \ldots f^{(2)}(f^{(1)}(x)))
\end{aligned}
\end{equation}
where $f^{(i)}$ is the function of the $i^{\text{th}}$ layer of the network, where $ i = 1,2,3, \ldots , k$ and $k$ is the last layer of the neural network.
In the image classification problem, $g(x) \in \mathbb{R}^{N}$ is the probabilities (confidence) for all the available $N$ classes.

Also, in adversarial machine learning, one type of adversarial samples $\hat{x}$ are defined as:
\begin{equation}
\begin{aligned}
& \hat{x} = x + \epsilon_{x} \\
& \{ \hat{x} \in \mathbb{R}^{m \times n \times 3} \mid {\operatorname{argmax}}[g(x)] \ne {\operatorname{argmax}}[g(\hat{x})]  \}
\end{aligned}
\end{equation}
in which $\epsilon_{x}$ is the perturbation added to the input. 
There exists, a wide variety of norm-constraints imposed upon $\epsilon_{x}$ such as $L_0$, $L_1$, $L_2$, and $L_\infty$ which allows for different adversarial attacks.
$L_0$ norm allows attacks to perturb a few pixels strongly, $L_\infty$ norm allow all pixels to change slightly, and both $L_1$ and $L_2$ allow for a mix of both strategies.

Making use of the definition of adversarial samples, adversarial machine learning thus, can be formally defined as the following optimization problem for untargeted attacks:
\begin{equation}
\begin{aligned}
& \underset{\epsilon_{x}}{\text{minimize}}
& & g(x+\epsilon_{x})_C
& \text{subject to}
& & \Vert \epsilon_{x} \Vert_p \leq th
\label{adv_eqn}
\end{aligned}
\end{equation}
Similarly optimization problem for the targeted attacks can be defined as:
\begin{equation}
\begin{aligned}
& \underset{\epsilon_{x}}{\text{maximize}}
& & g(x+\epsilon_{x})_T
& \text{subject to}
& & \Vert \epsilon_{x} \Vert_p \leq th
\label{adv_eqn}
\end{aligned}
\end{equation}
where $g()_C$ is the soft-label for the correct class, $g()_T$ is the soft-label for the target class, $p \in \{0, 1, 2, \infty \}$ is the constraint norm on $\epsilon_{x}$ and $th$ is the threshold value for the constraint norm.

Saliency Maps (pixel-attribution maps or attribution maps or sensitivity maps) \cite{simonyan2013deep} is used to assess how output changes concerning a change in input. 
It can be used to determine the region of interest in the image for neural networks.
Mathematically, the saliency map, $SM \in \mathbb{R}^{m,n}$ can be defined as:
\begin{equation}
\begin{aligned}
& SM = \frac{\partial \text{ Output}}{\partial \text{ Input}} 
\end{aligned}
\end{equation}
Hence, we can evaluate saliency map of an image with respect to predicted class of the image $C$ as: 
\begin{equation}
\begin{aligned}
& SM_C = max_c \frac{\partial g(x)_C}{\partial x} 
\end{aligned}
\end{equation}
Similarly, we can evaluate the saliency map of the adversarial image concerning the predicted class of adversarial image $\hat{C}$ as: 
\begin{equation}
\begin{aligned}
& \widehat{SM}_{\hat{C}} = max_c \frac{\partial g(\hat{x})_{\hat{C}}}{\partial \hat{x}} = max_c \frac{\partial g(x + \epsilon_{x})_{\hat{C}}}{\partial (x + \epsilon_{x})} 
\end{aligned}
\end{equation}

Gradient-weighted Class Activation Map (Grad-CAM) \cite{selvaraju2017grad} is another type of saliency map evaluation that assesses the attention of a convolution layer concerning a label.
It can be used to determine the attention region of a convolution layer in the image for neural networks.
Mathematically, Gradient-weighted Class Activation Map with respect to label $l$ for $n^{\text{th}}$ layer $AM_l \in \mathbb{R}^{m,n}$ can be defined as:
\begin{equation}
\begin{aligned}
& AM_l = f^{(n)}(x) \boldsymbol{\cdot} mean_c \frac{\partial g(x)_l}{\partial f^{(n)}(x)} 
\end{aligned}
\end{equation}
Based on this, we can evaluate activation maps of the image concerning predicted class $C$ as:
\begin{equation}
\begin{aligned}
& AM_C = f^{(n)}(x) \boldsymbol{\cdot} mean_c \frac{\partial g(x)_C}{\partial f^{(n)}(x)} 
\end{aligned}
\end{equation}
and we can evaluate activation maps of adversarial image with respect to predicted class $\hat{C}$ 
\begin{equation}
\begin{aligned}
& \widehat{AM}_{\hat{C}} = f^{(n)}(\hat{x}) \boldsymbol{\cdot} mean_c \frac{\partial g(\hat{x})_C}{\partial f^{(n)}(\hat{x})} \\
& \quad \quad = f^{(n)}(x + \epsilon_{x}) \boldsymbol{\cdot} mean_c \frac{\partial g(x + \epsilon_{x})_C}{\partial f^{(n)}(x + \epsilon_{x})}
\end{aligned}
\end{equation}

This article assesses Saliency Maps $(SM)$ and Gradient-weighted Class Activations Maps $(AM)$ concerning correctly predicted classes (or the true class) $(C)$ and misclassified/incorrectly predicted class (or the adversarial class) $(\hat{C})$ of both adversarial and original images. 
Analysing Saliency Maps and Activations Maps for other unrelated classes are left as future work. 
We also focus on the Activations Maps of the last convolution layer in the neural network and leave the activation maps of other layers as future work.

\section{Experimental results and analysis}

\textbf{Experimental settings:}
We use ResNet \cite{he2016deep} which is trained on CIFAR-10 dataset \cite{krizhevsky2009learning}.
For adversarial attacks, we evaluate Pixel Attack ($L_0$ norm black-box attack) \cite{su2019one} and Projected Gradient Descent Attack ($L_\infty$ norm white-box attack) \cite{madry2018towards} using the Adversarial Robustness Toolbox library \cite{art2018}.
For evaluating saliency maps, we replace the traditional ReLU backpropagation with Guided ReLU backpropagation \cite{springenberg2014striving}. 

\subsection{Saliency Maps}

\begin{figure}[!t]
    \centering
    \textbf{Pixel Attack}\\
    \vspace{0.1cm}
    \hrule
    \vspace{0.1cm}\small{\hspace{0.5cm} $x$ \hspace{1.2cm} $SM_{C}$ \hspace{1.cm} $\hat{x}$\hspace{1.3cm} $\widehat{SM}_{\hat{C}}$}\\
    \includegraphics[width=0.2\columnwidth]{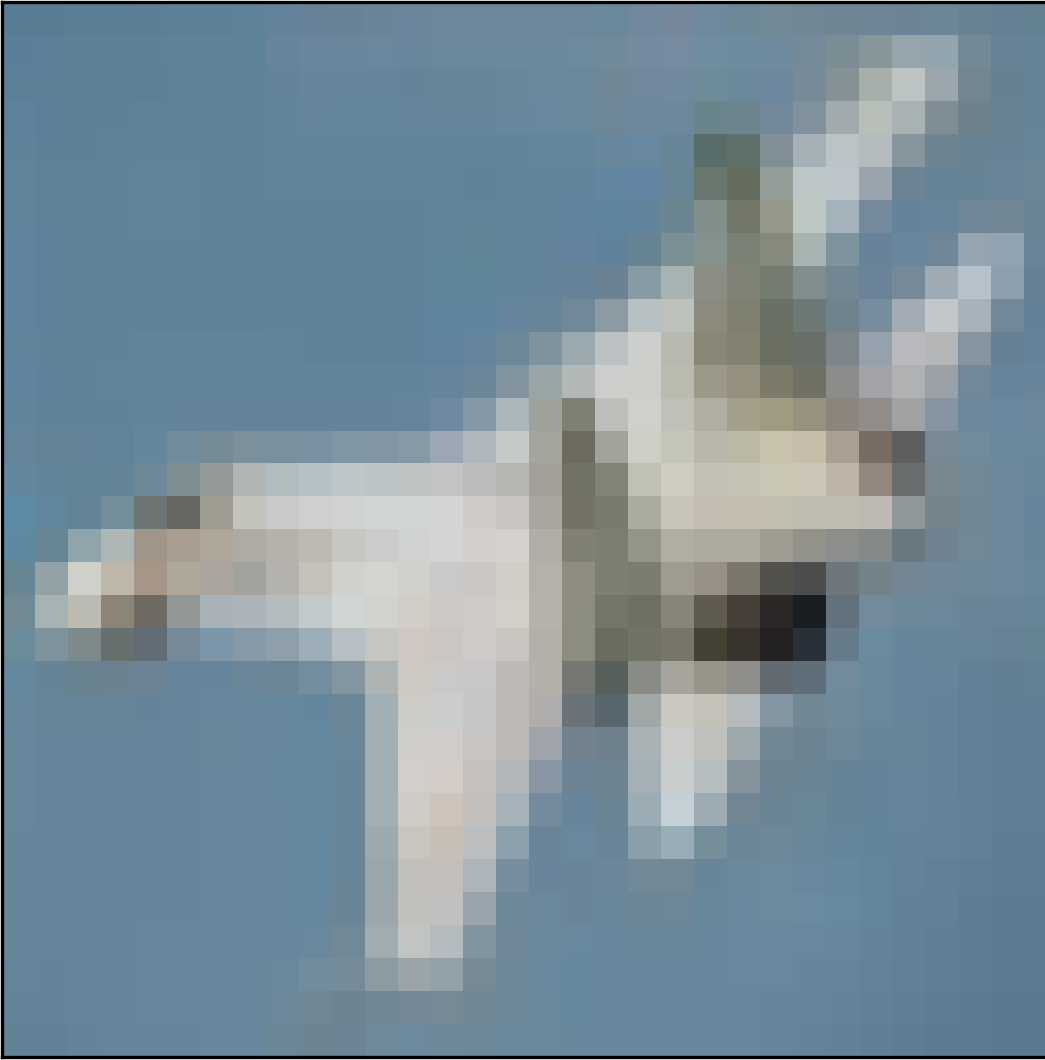} 
    \includegraphics[width=0.2\columnwidth]{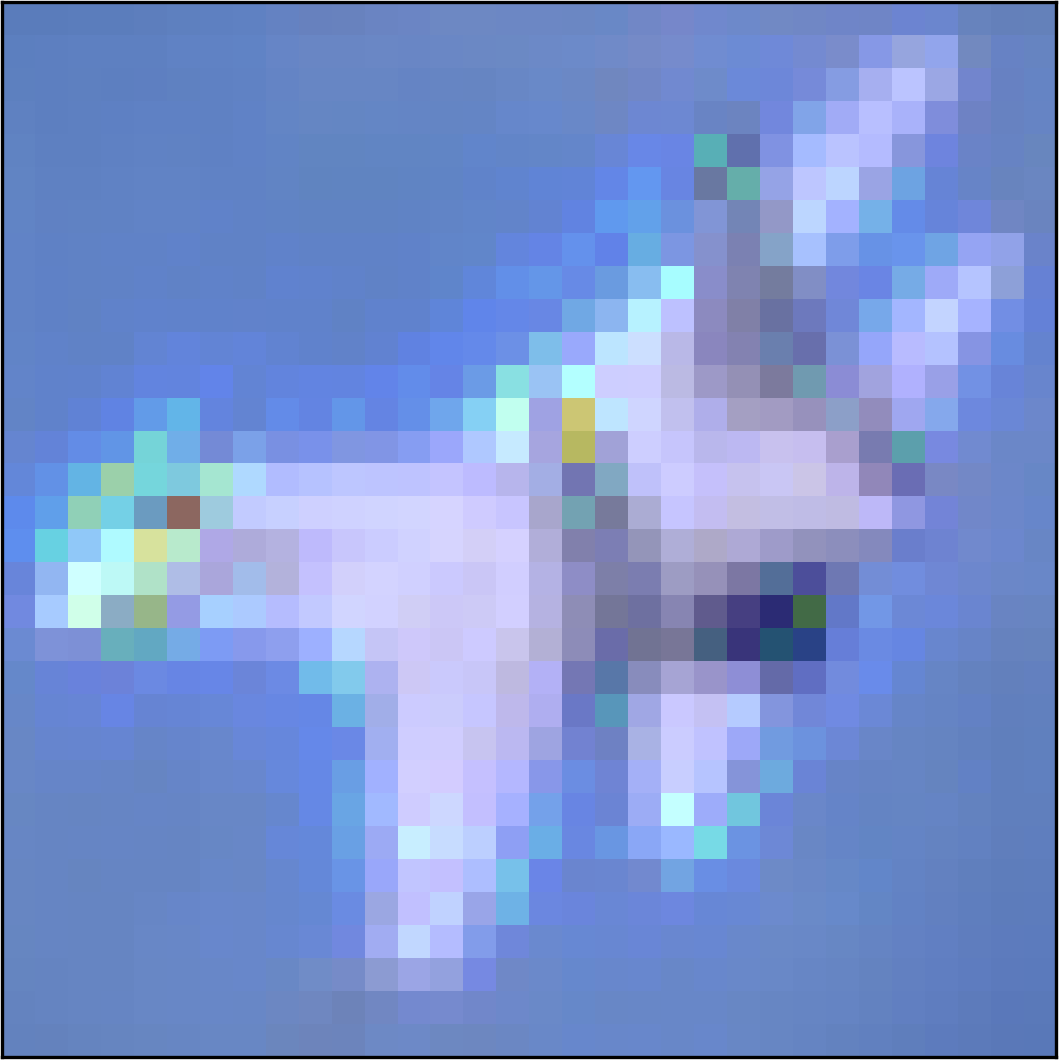} 
    \includegraphics[width=0.2\columnwidth]{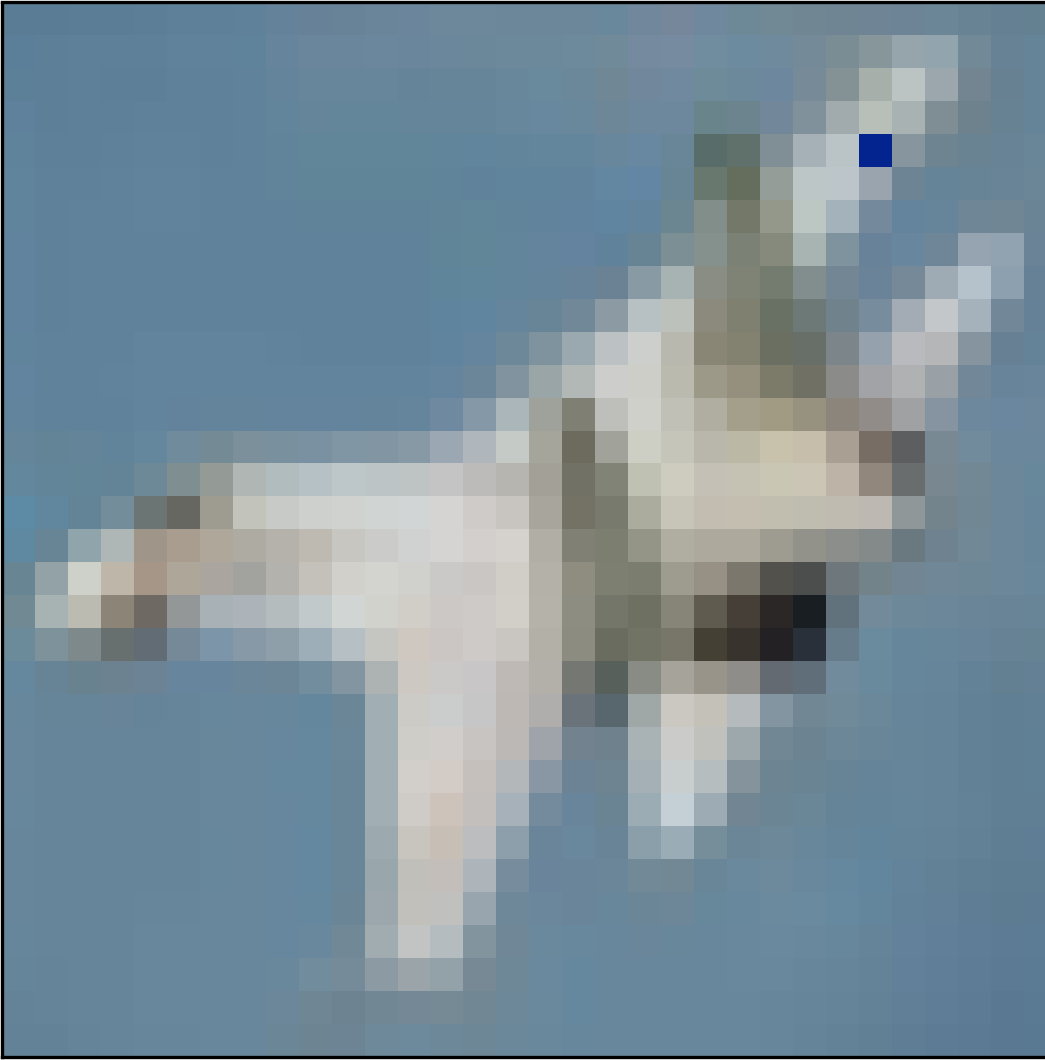} 
    \includegraphics[width=0.2\columnwidth]{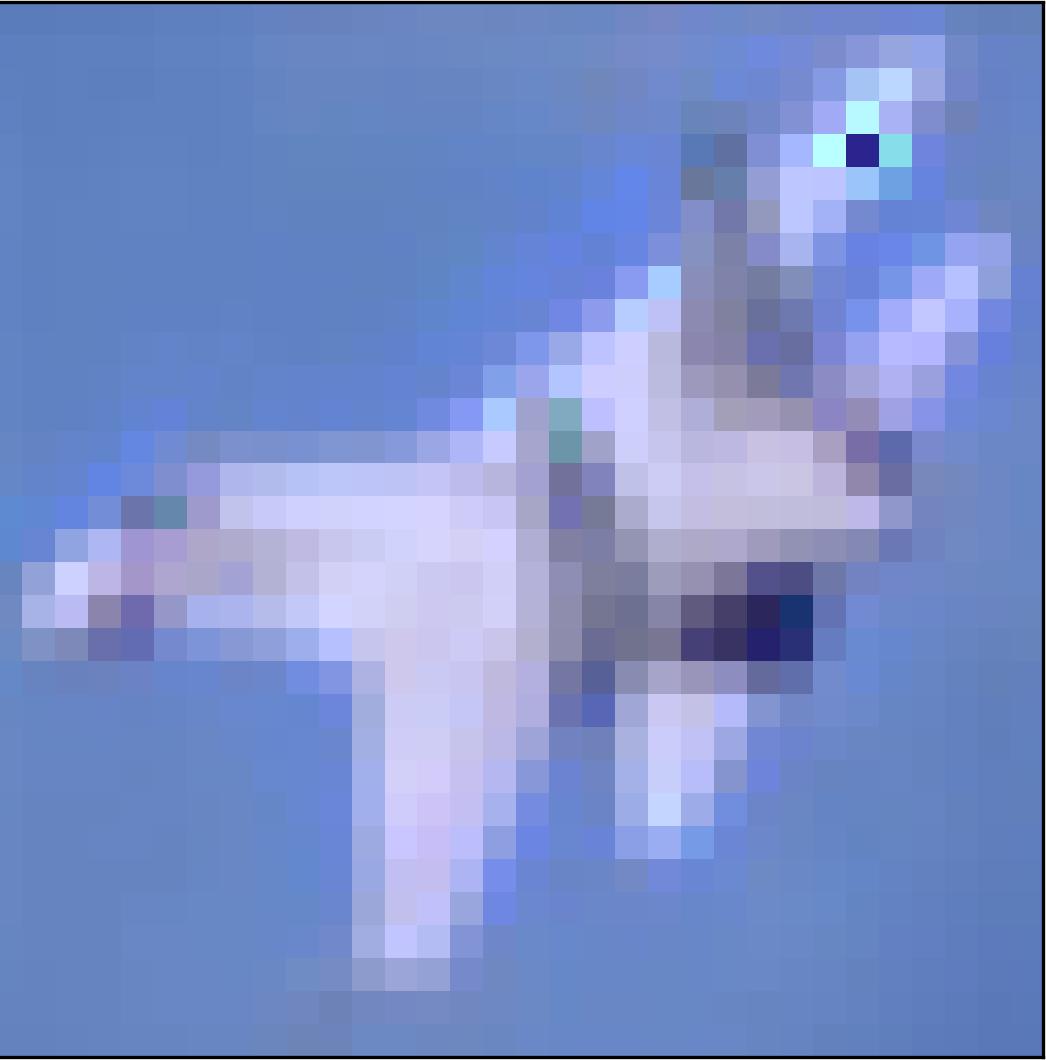} \\
    \vspace{0.1cm}\small{\hspace{0.5cm} $x$ \hspace{1.2cm} $SM_{C}$ \hspace{1.cm} $\hat{x}$\hspace{1.3cm} $\widehat{SM}_{\hat{C}}$}\\
    \includegraphics[width=0.2\columnwidth]{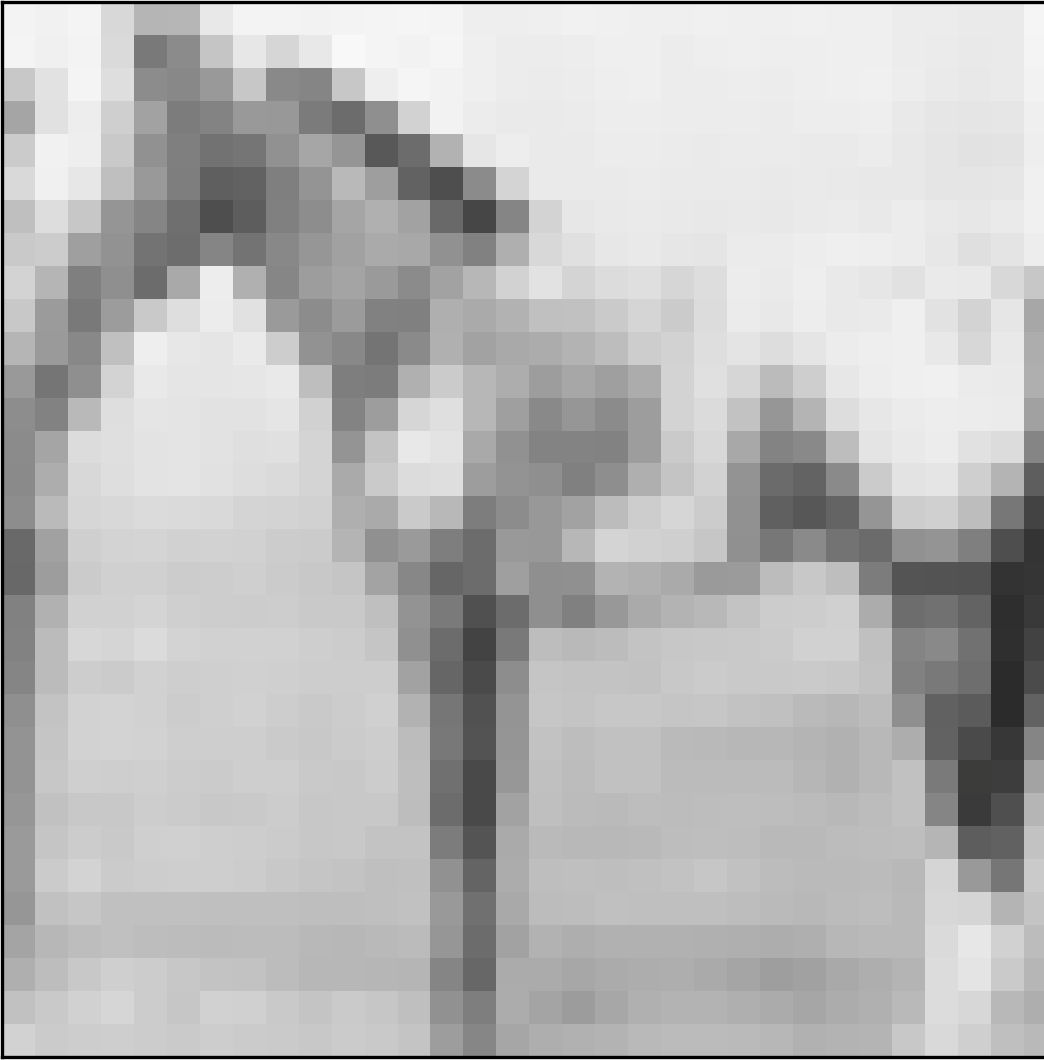} 
    \includegraphics[width=0.2\columnwidth]{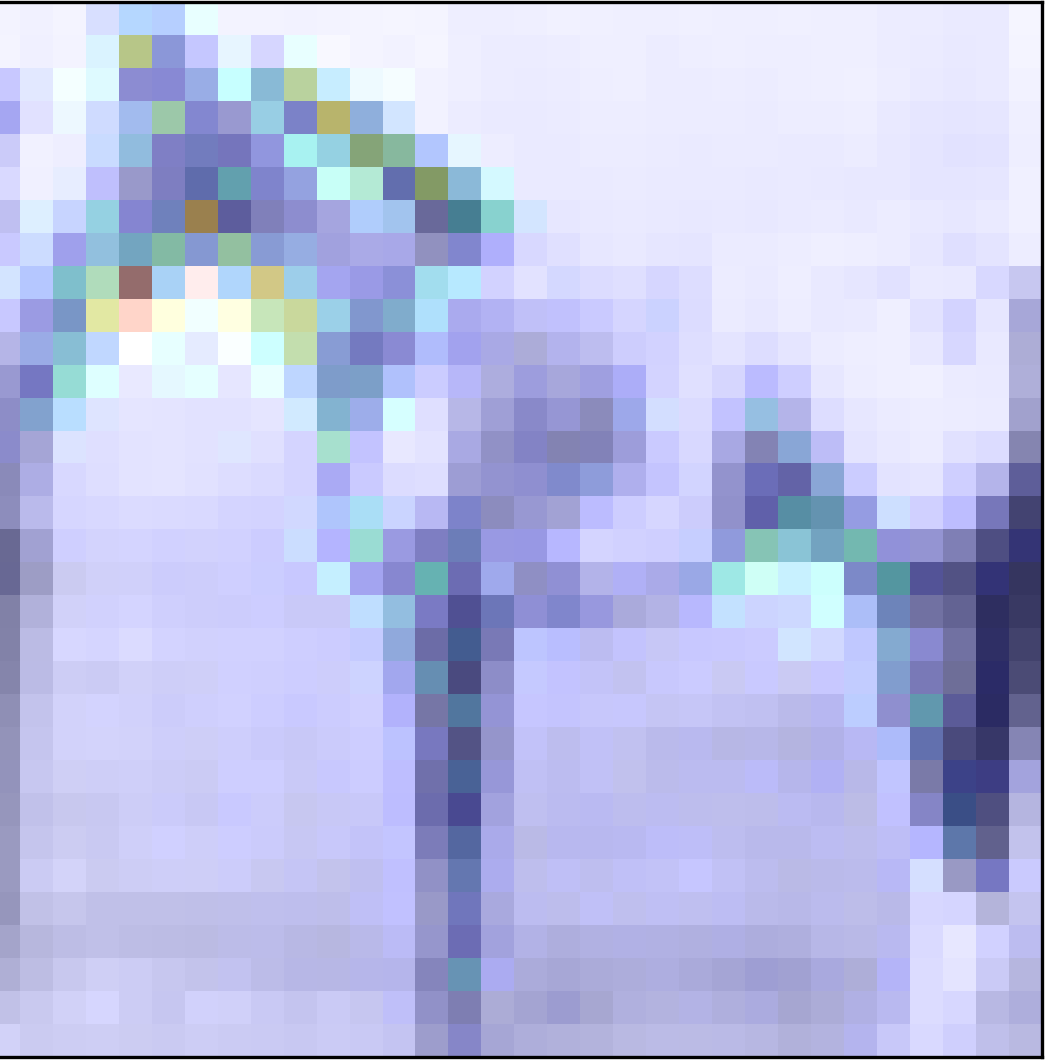} 
    \includegraphics[width=0.2\columnwidth]{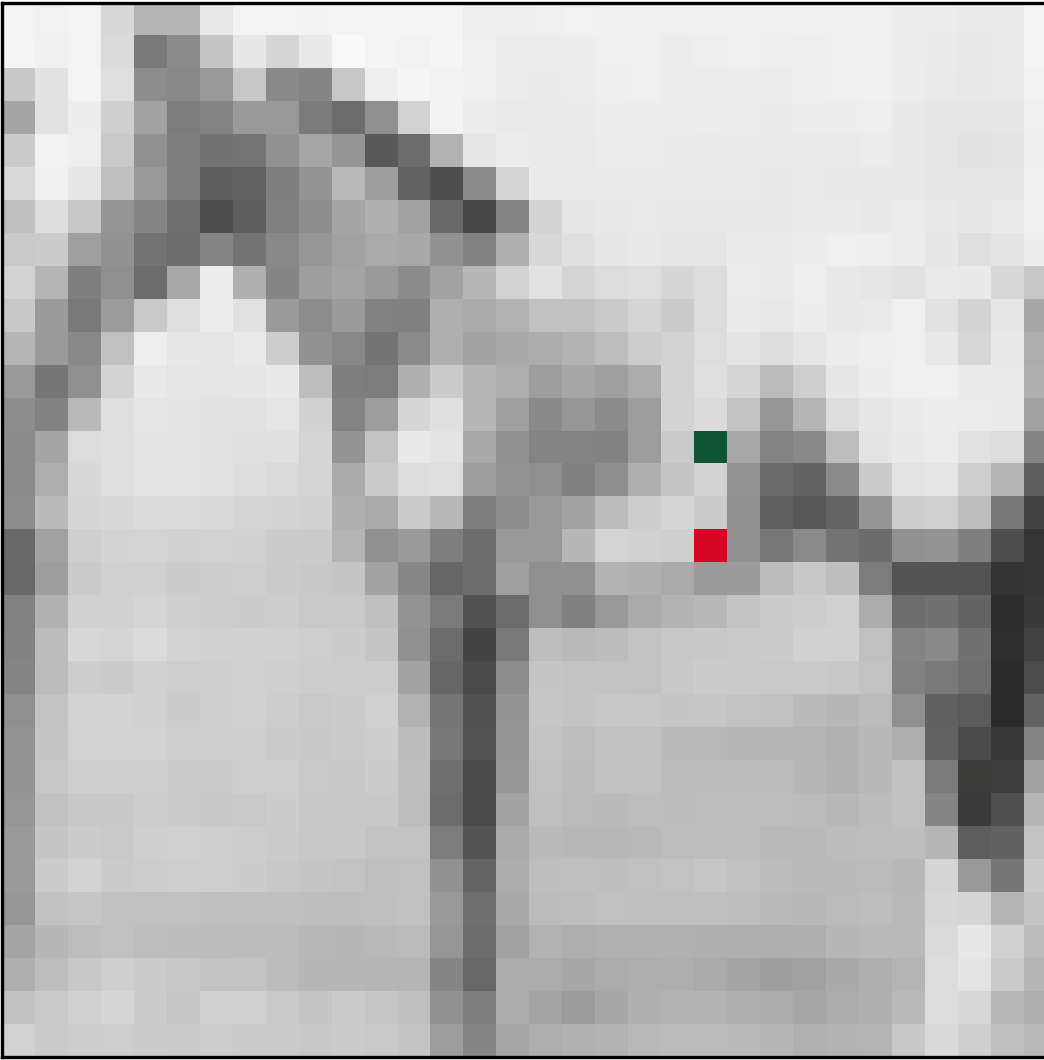} 
    \includegraphics[width=0.2\columnwidth]{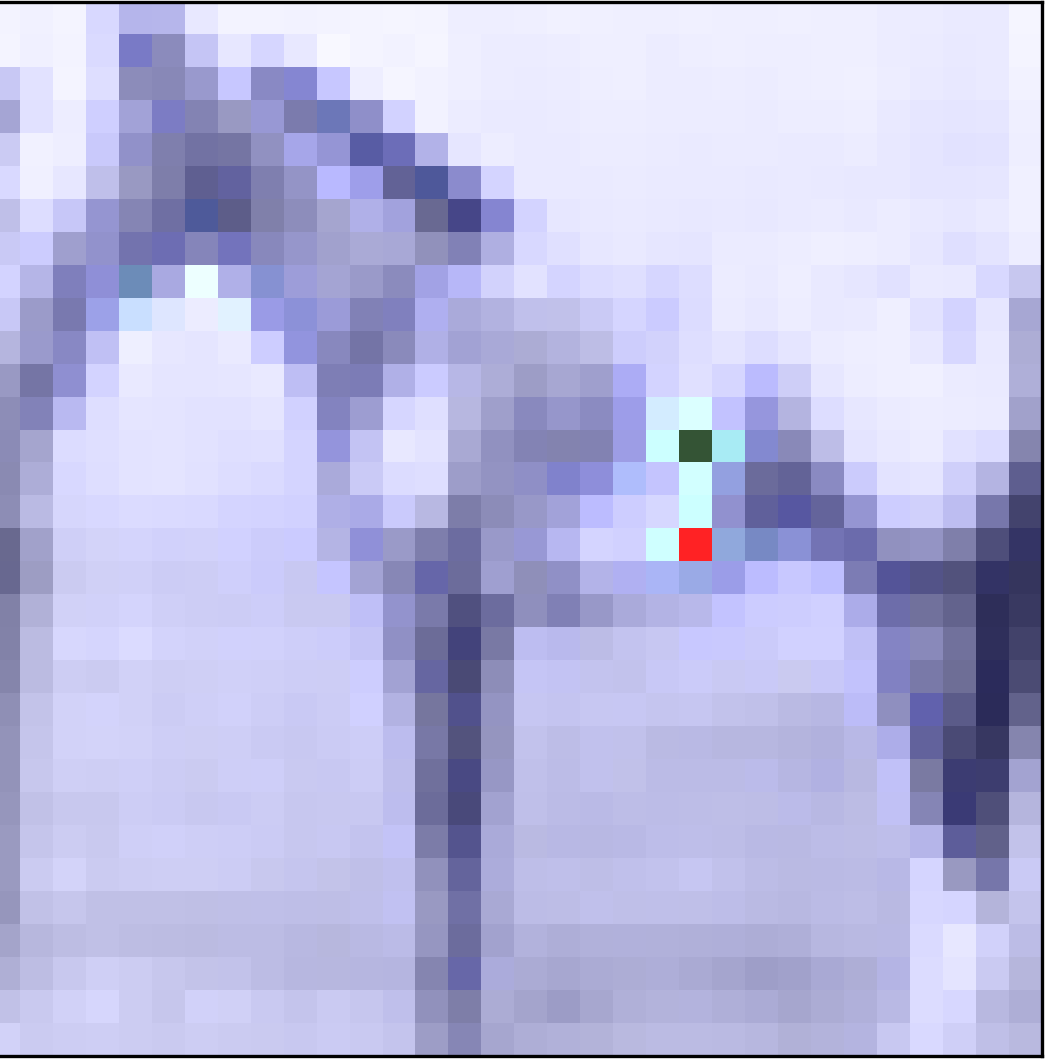} \\
    \vspace{0.1cm}\small{\hspace{0.5cm} $x$ \hspace{1.2cm} $SM_{C}$ \hspace{1.cm} $\hat{x}$\hspace{1.3cm} $\widehat{SM}_{\hat{C}}$}\\
    \includegraphics[width=0.2\columnwidth]{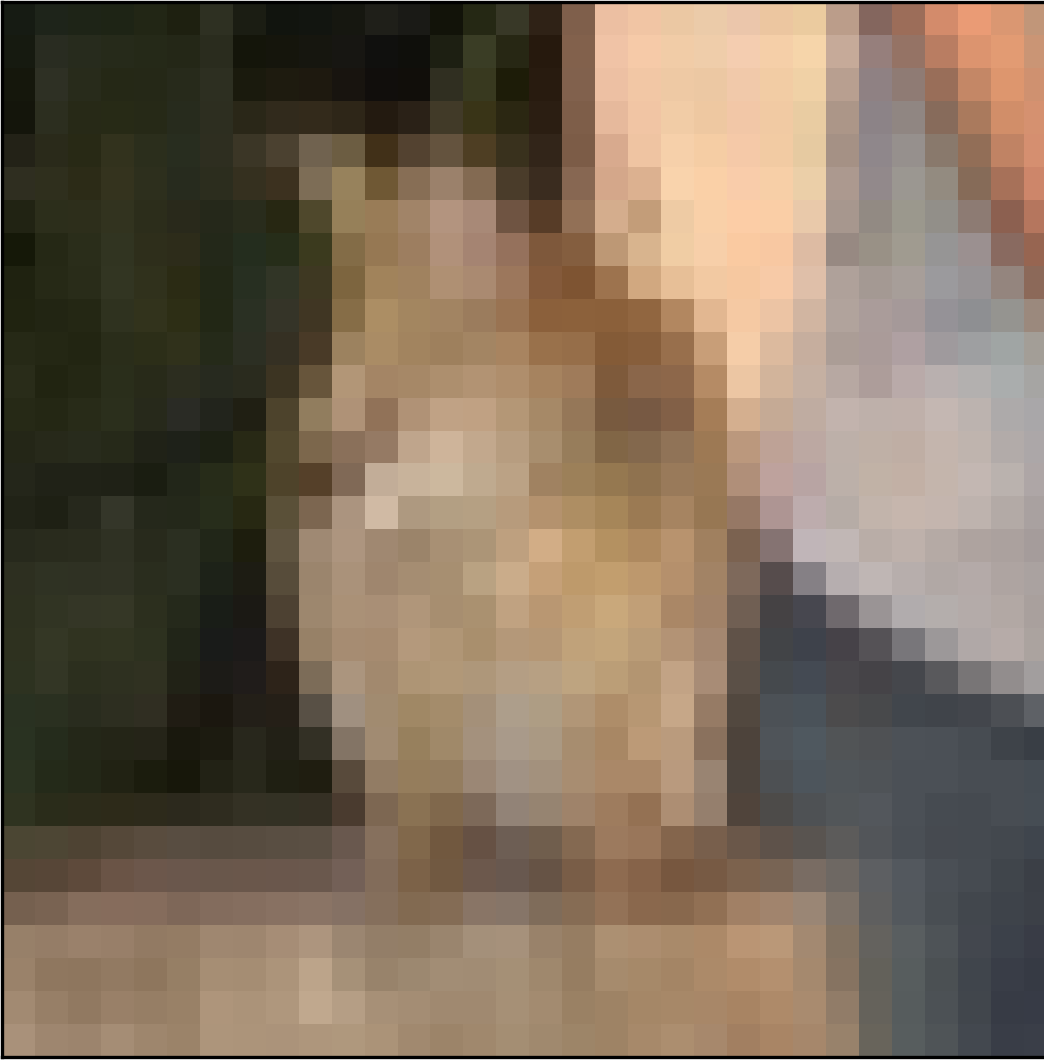} 
    \includegraphics[width=0.2\columnwidth]{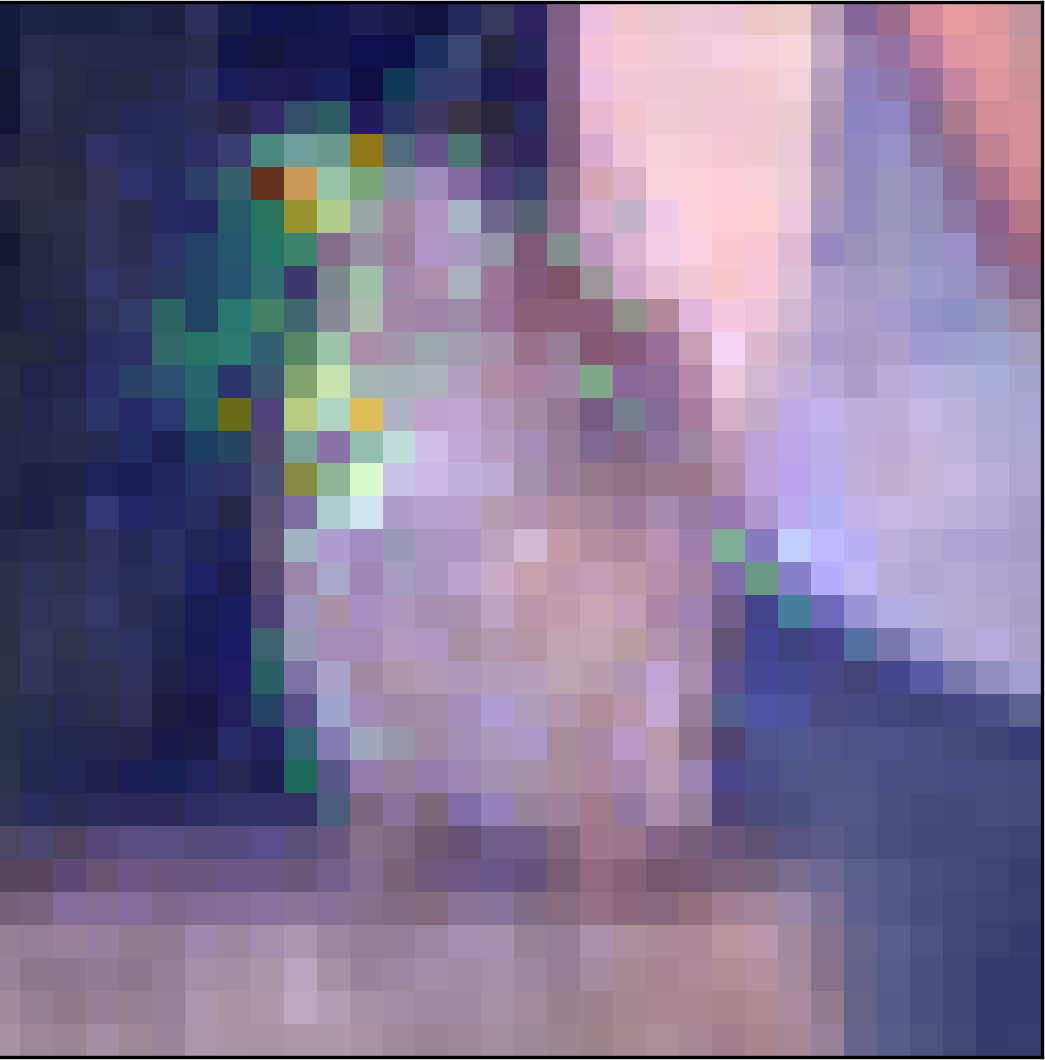} 
    \includegraphics[width=0.2\columnwidth]{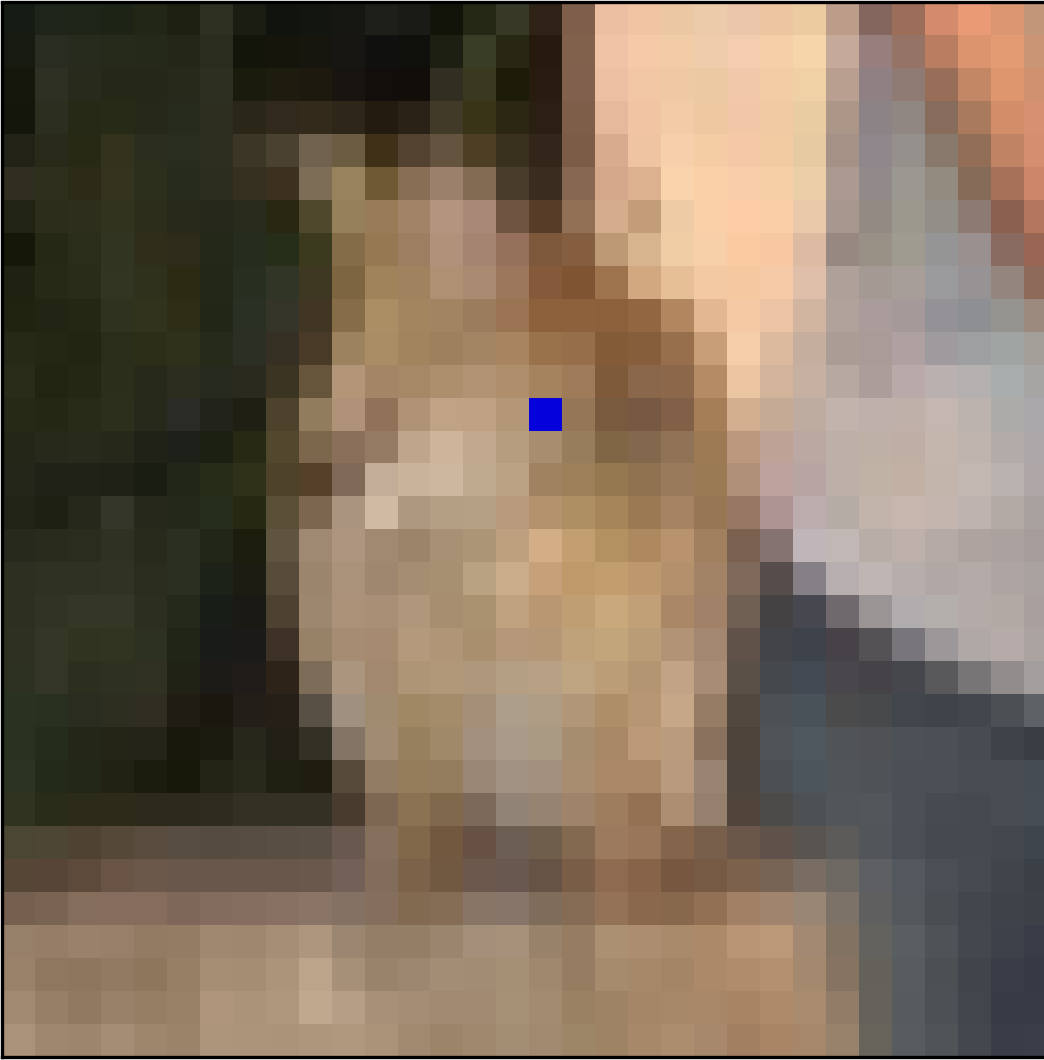} 
    \includegraphics[width=0.2\columnwidth]{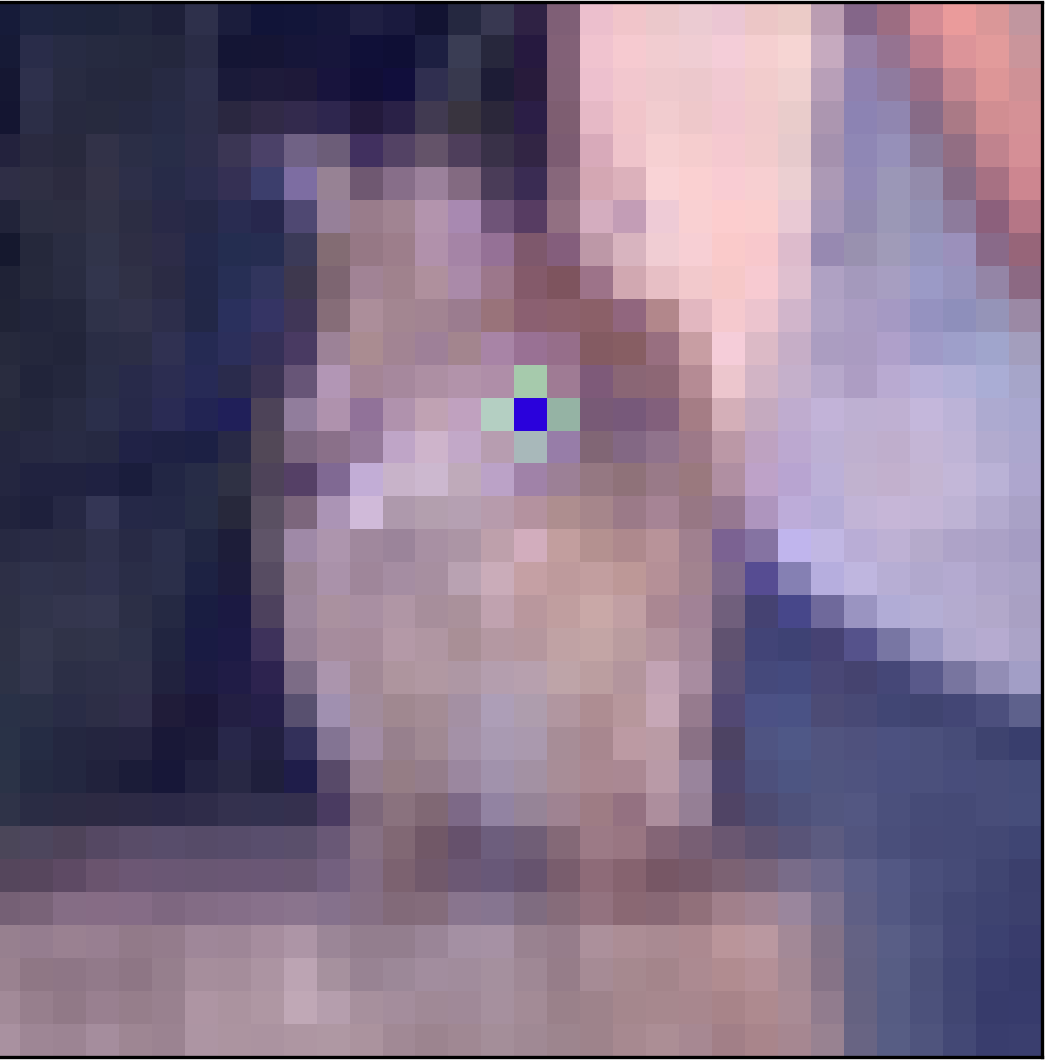} \\
    \vspace{0.1cm}
    \textbf{Projected Gradient Descent Attack}\\
    \vspace{0.1cm}
    \hrule
    \vspace{0.1cm}\small{\hspace{0.5cm} $x$ \hspace{1.2cm} $SM_{C}$ \hspace{1.cm} $\hat{x}$\hspace{1.3cm} $\widehat{SM}_{\hat{C}}$}\\
    \includegraphics[width=0.2\columnwidth]{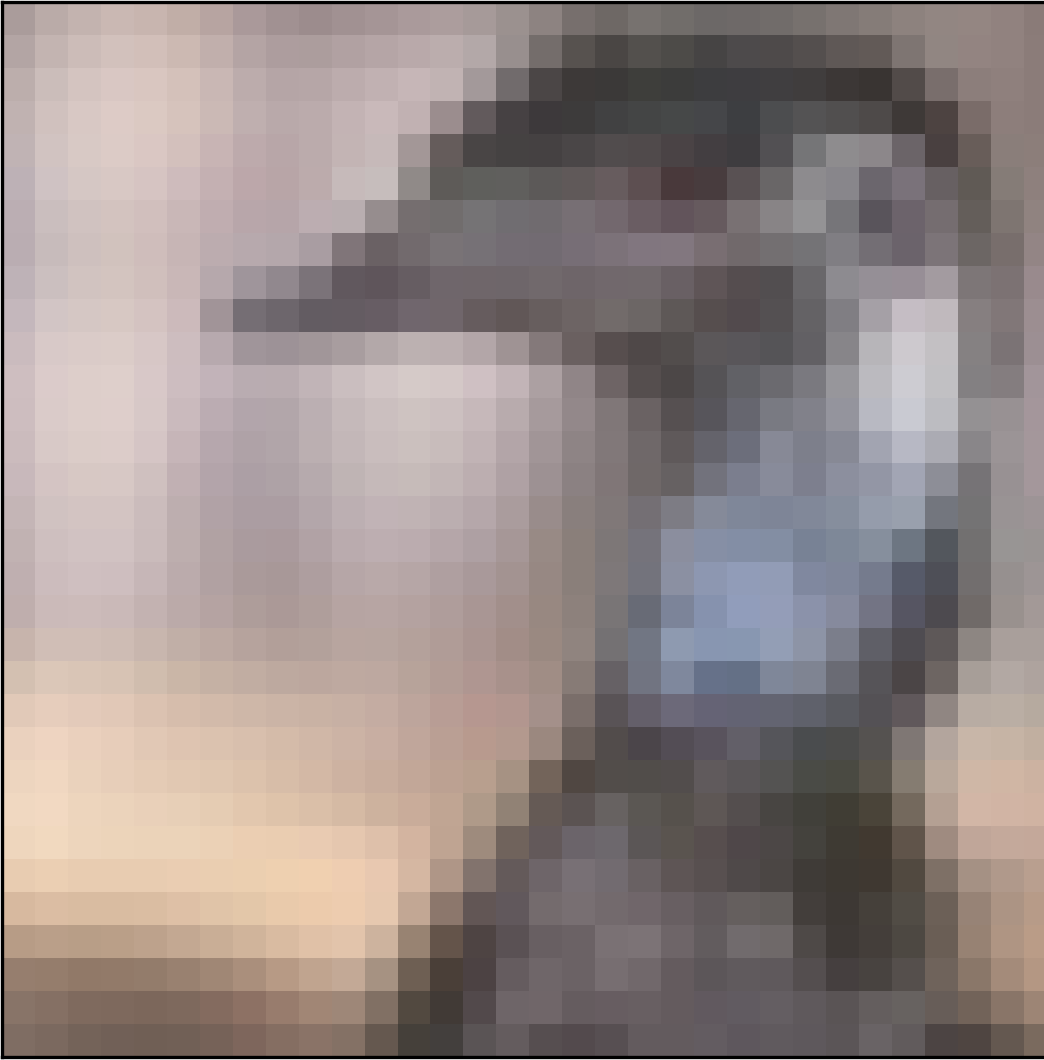} 
    \includegraphics[width=0.2\columnwidth]{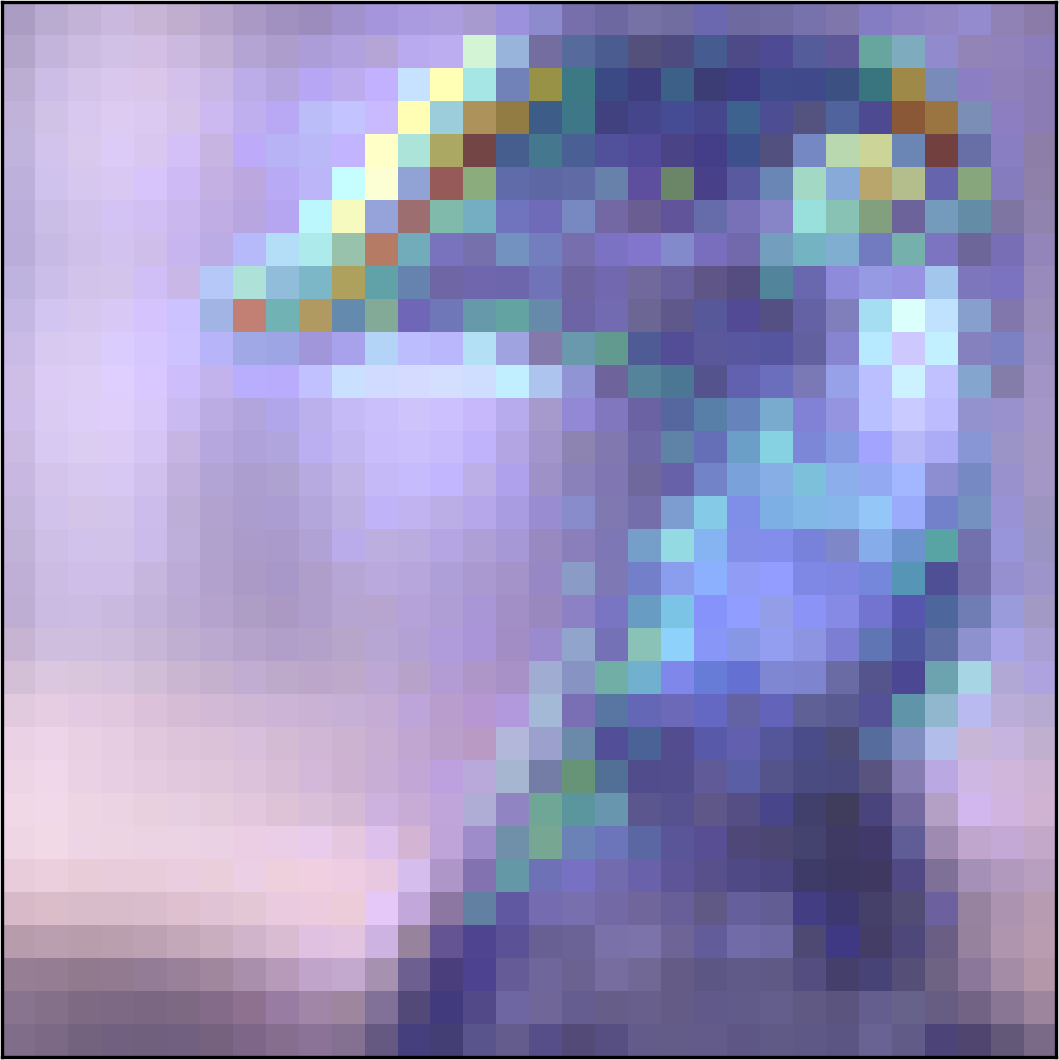} 
    \includegraphics[width=0.2\columnwidth]{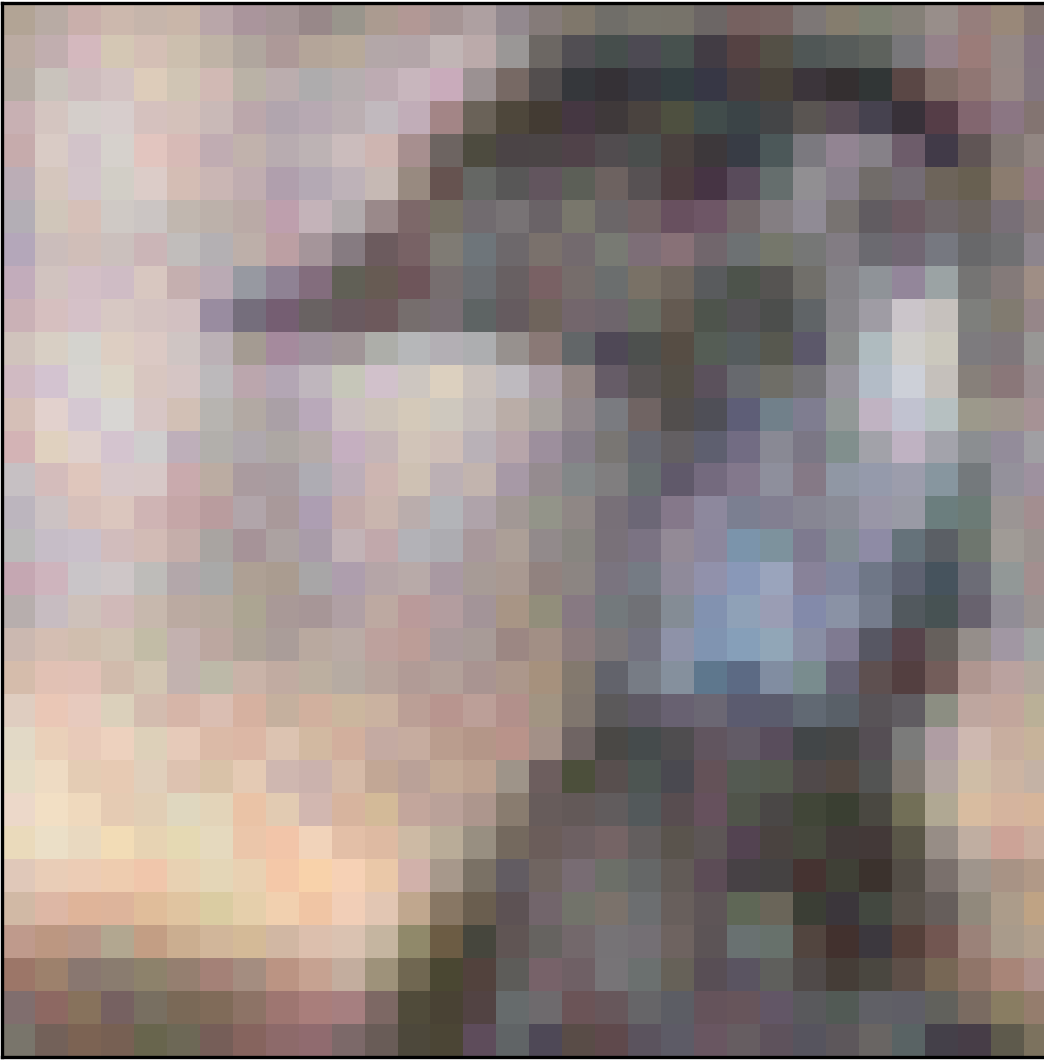} 
    \includegraphics[width=0.2\columnwidth]{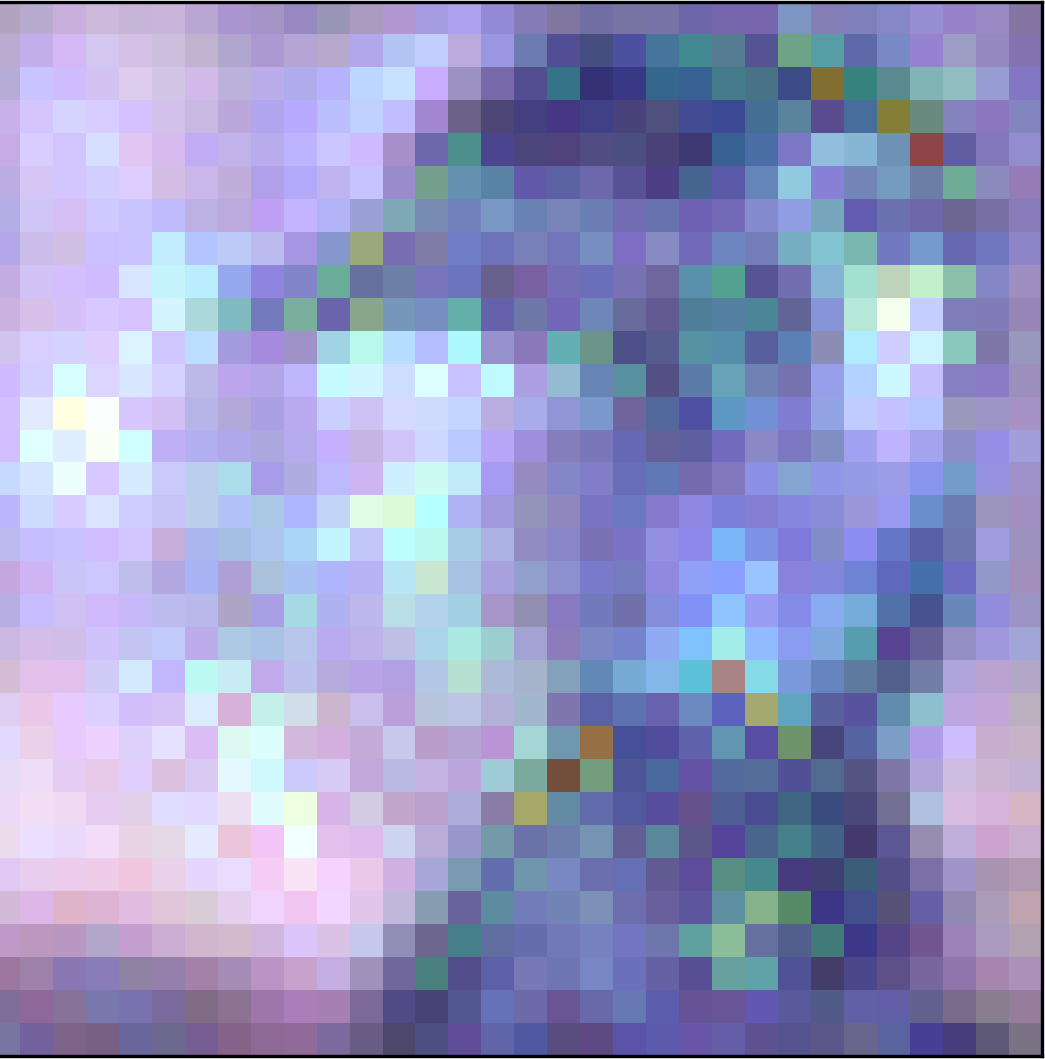} \\
    \vspace{0.1cm}\small{\hspace{0.5cm} $x$ \hspace{1.2cm} $SM_{C}$ \hspace{1.cm} $\hat{x}$\hspace{1.3cm} $\widehat{SM}_{\hat{C}}$}\\
    \includegraphics[width=0.2\columnwidth]{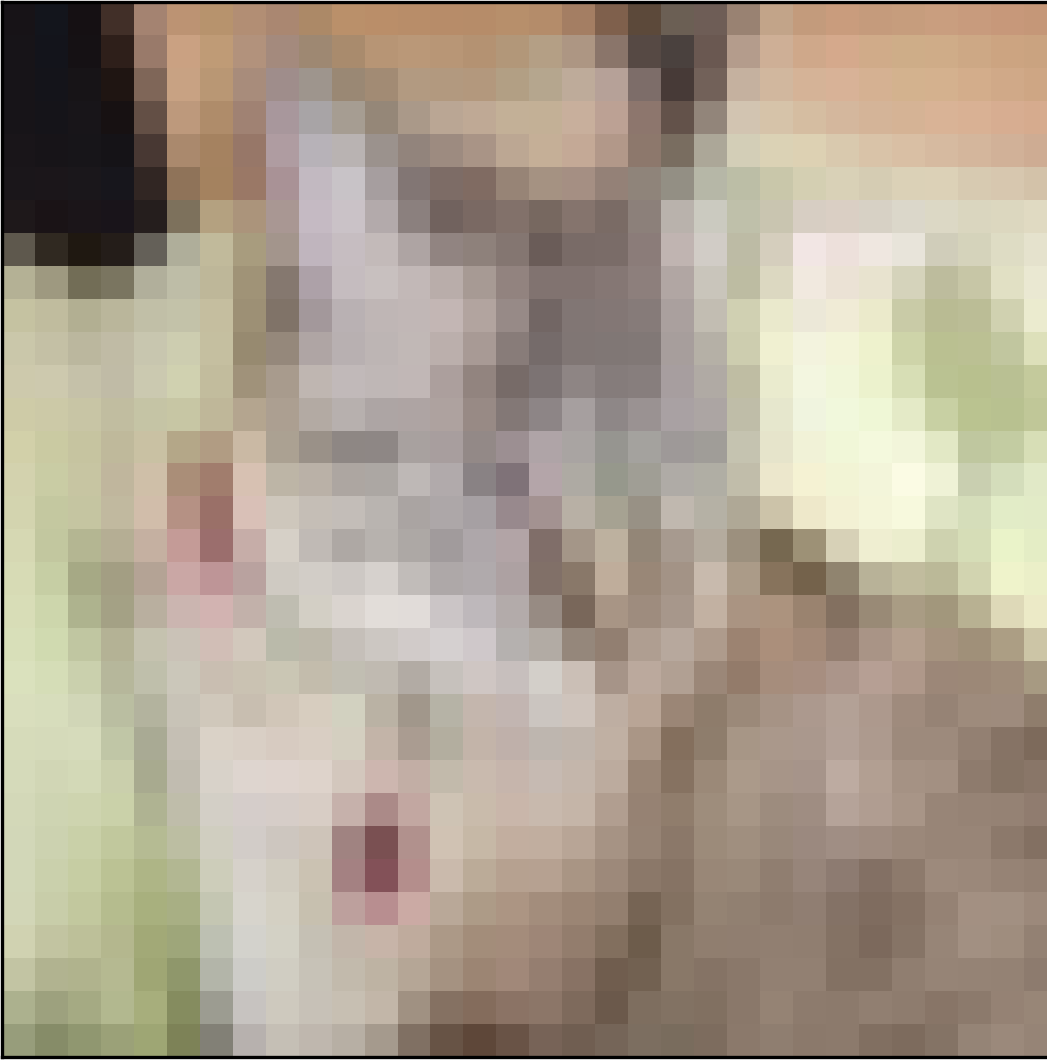} 
    \includegraphics[width=0.2\columnwidth]{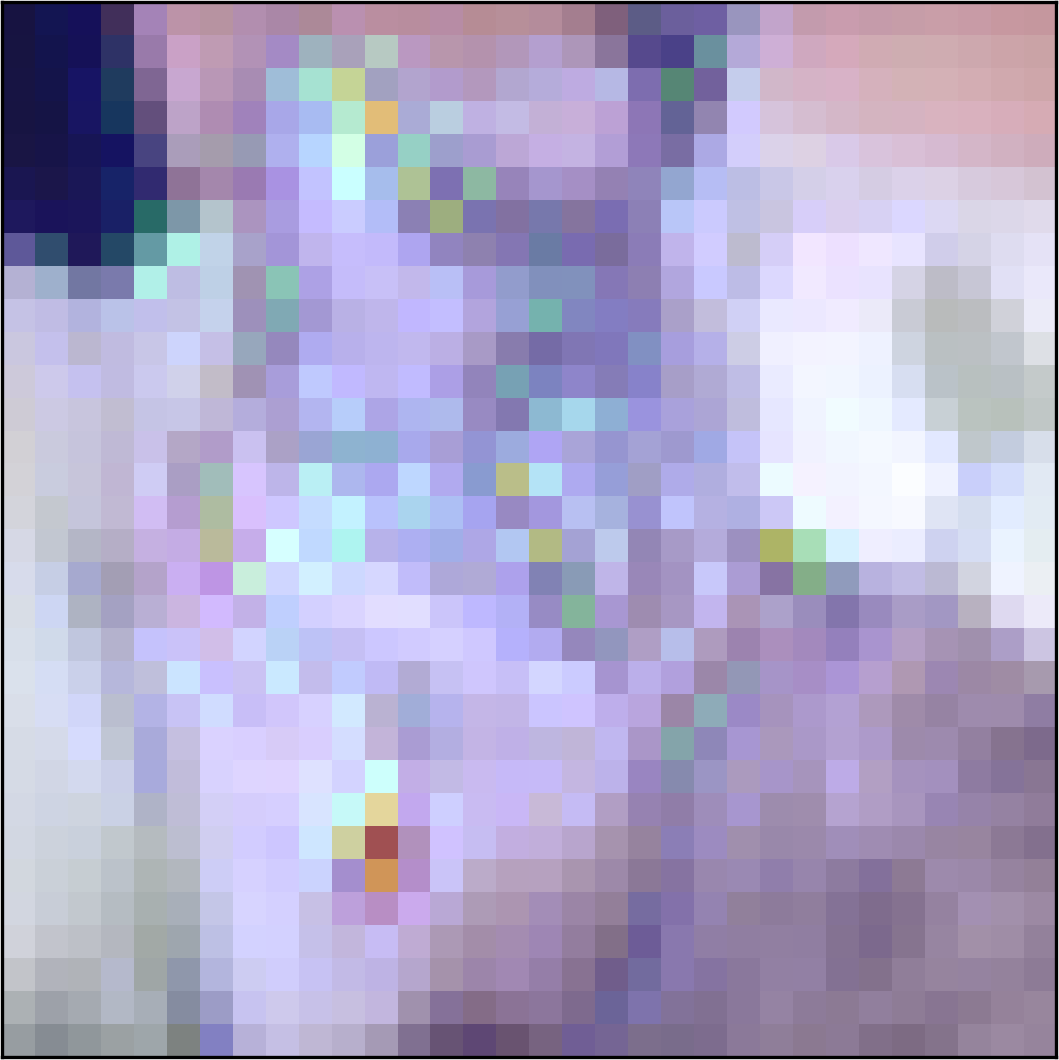} 
    \includegraphics[width=0.2\columnwidth]{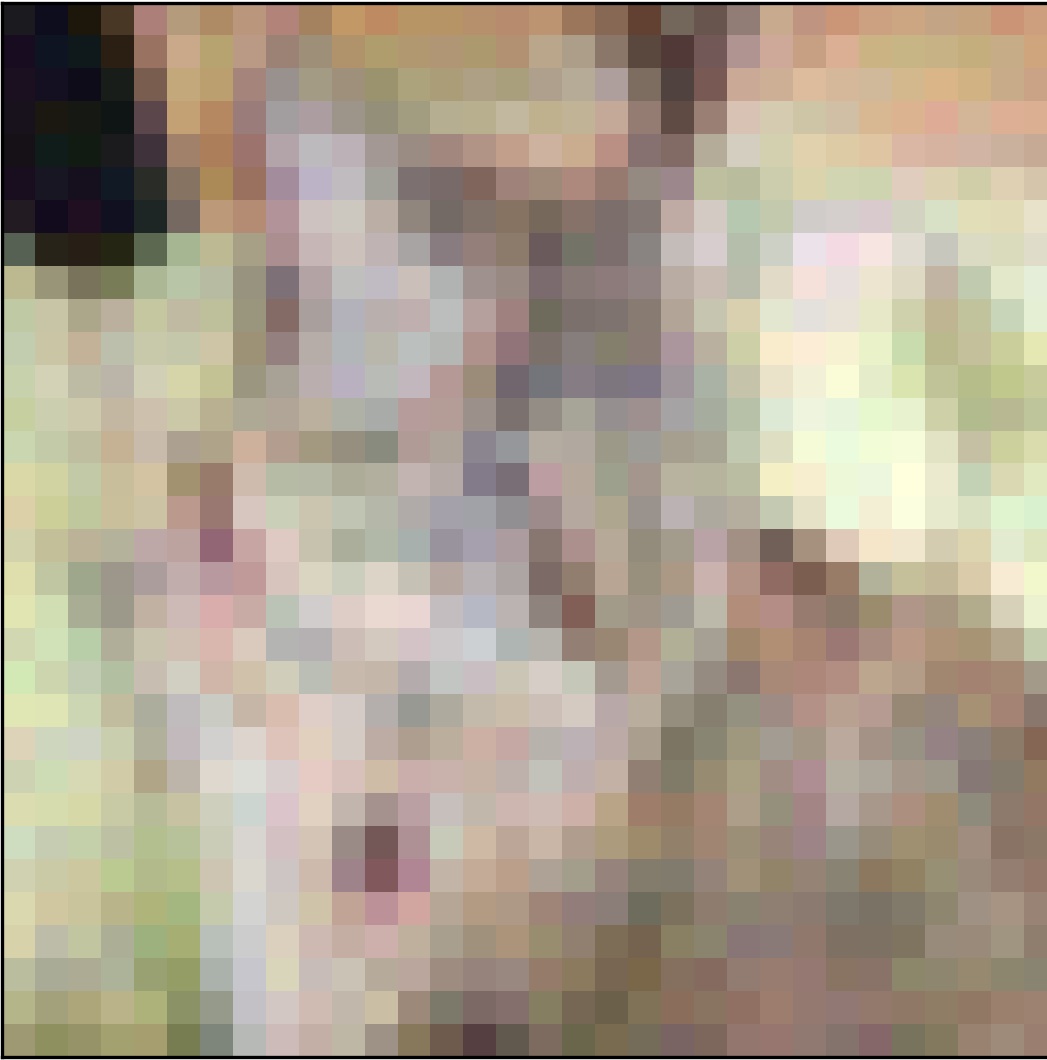} 
    \includegraphics[width=0.2\columnwidth]{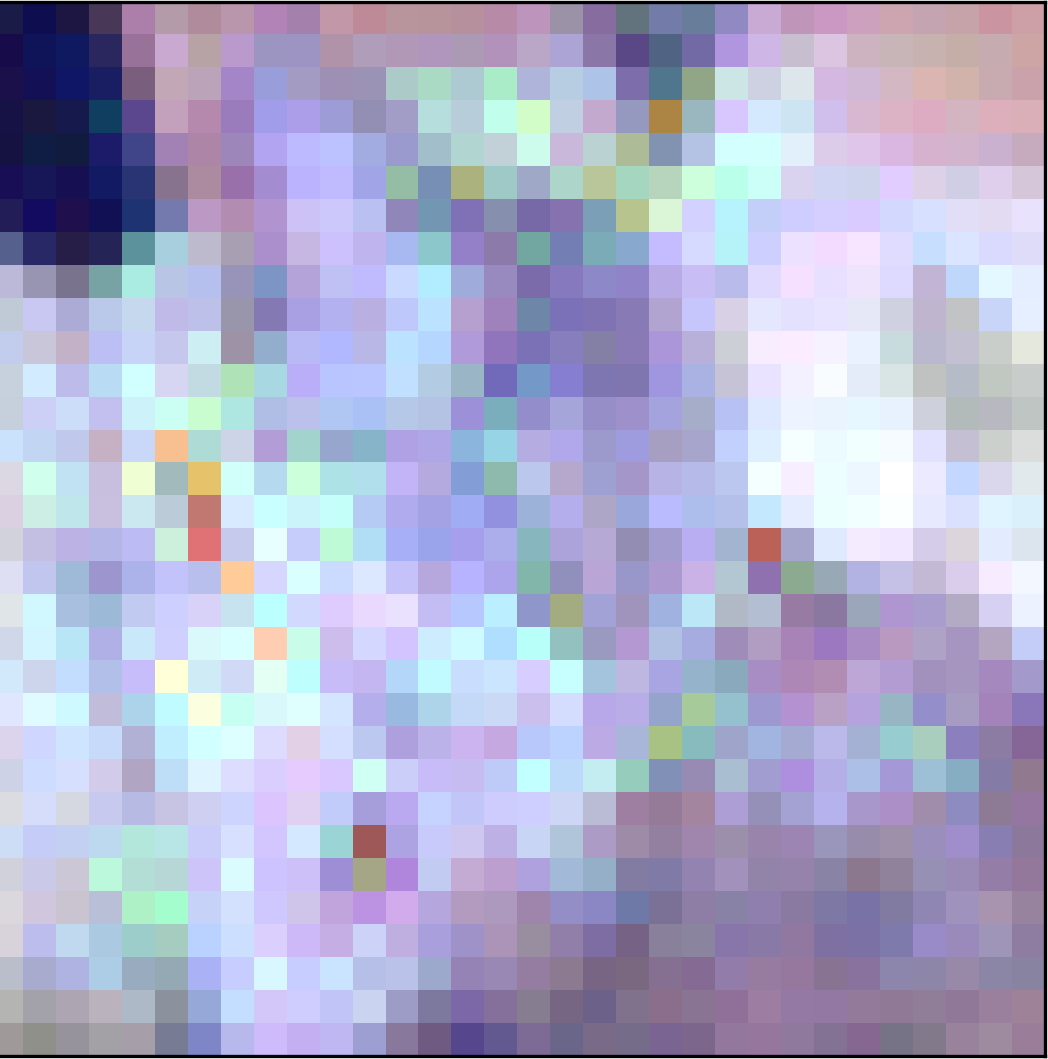} \\
    \caption{Overlayed Saliency Maps with respect to predicted class.}
    \label{saliency_pred} 
\end{figure}

\begin{figure}[!htb]
    \centering
    \textbf{Pixel Attack}\\
    \vspace{0.1cm}
    \hrule
    \vspace{0.1cm}\small{\hspace{0.5cm} $x$ \hspace{1.2cm} $SM_{C}$ \hspace{1.cm} $\hat{x}$\hspace{1.3cm} $\widehat{SM}_{C}$}\\
    \includegraphics[width=0.2\columnwidth]{images/cropped_PixelAttack-test_10_saliency_OI.png} 
    \includegraphics[width=0.2\columnwidth]{images/cropped_PixelAttack-test_10_saliency_OST.png} 
    \includegraphics[width=0.2\columnwidth]{images/cropped_PixelAttack-test_10_saliency_AI.png} 
    \includegraphics[width=0.2\columnwidth]{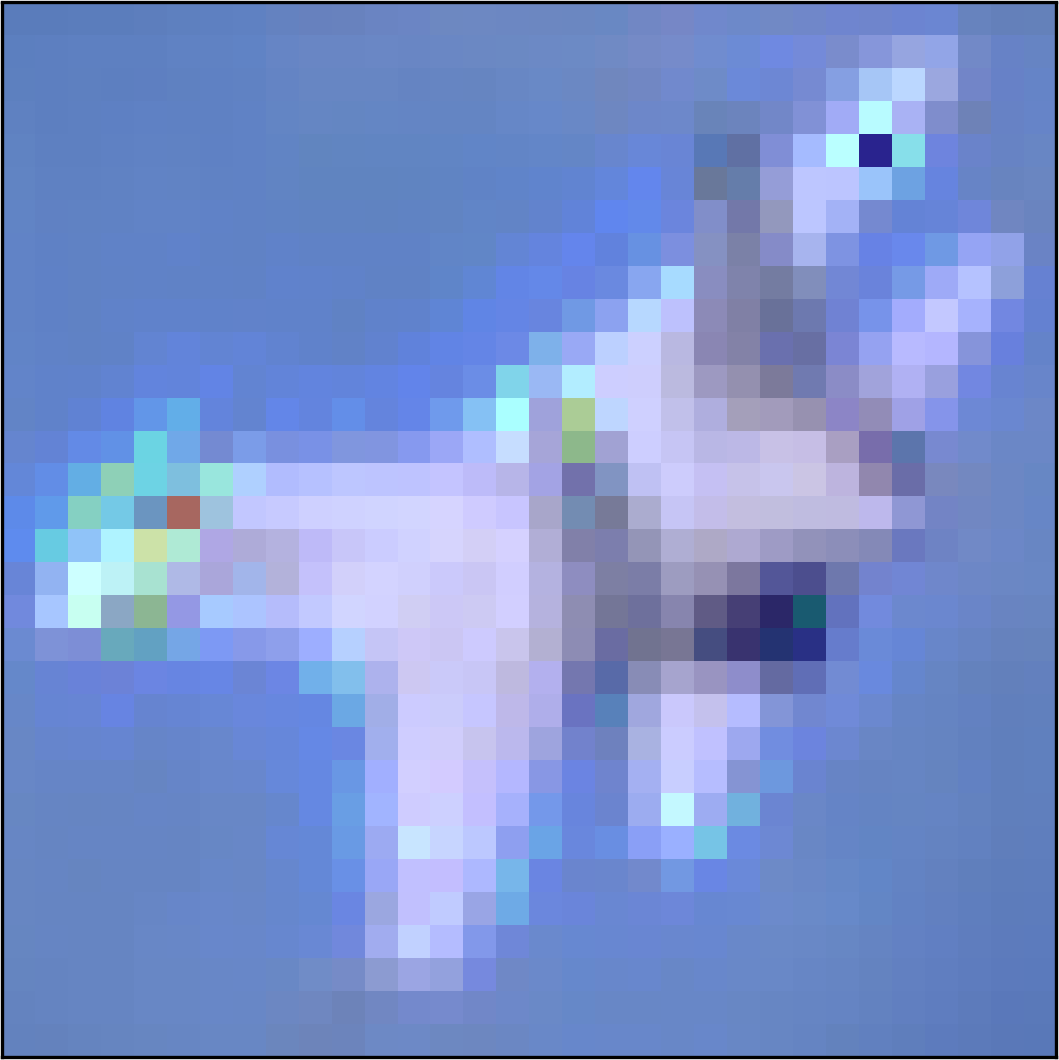} \\
    \vspace{0.1cm}\small{\hspace{0.5cm} $x$ \hspace{1.2cm} $SM_{C}$ \hspace{1.cm} $\hat{x}$\hspace{1.3cm} $\widehat{SM}_{C}$}\\
    \includegraphics[width=0.2\columnwidth]{images/cropped_PixelAttack-test_83_saliency_OI.png} 
    \includegraphics[width=0.2\columnwidth]{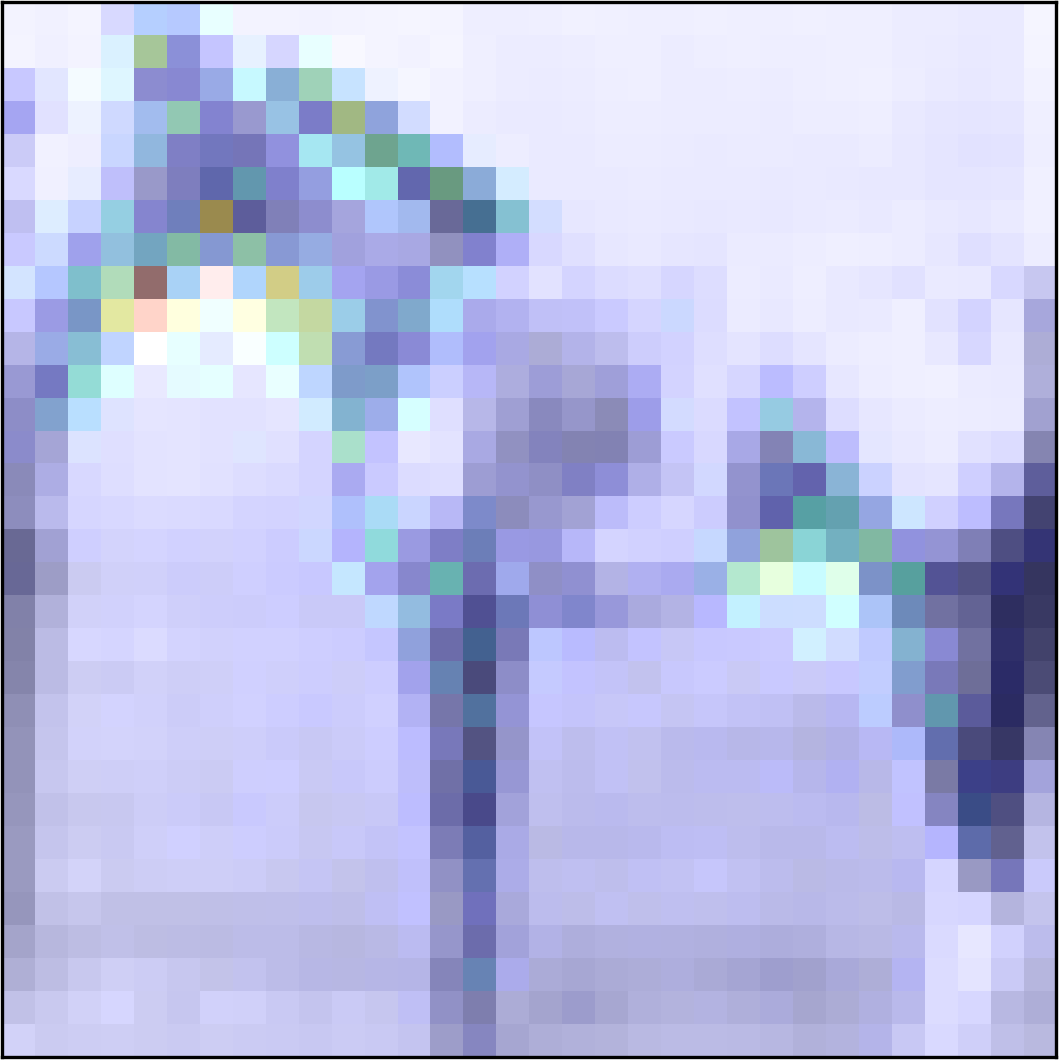} 
    \includegraphics[width=0.2\columnwidth]{images/cropped_PixelAttack-test_83_saliency_AI.png} 
    \includegraphics[width=0.2\columnwidth]{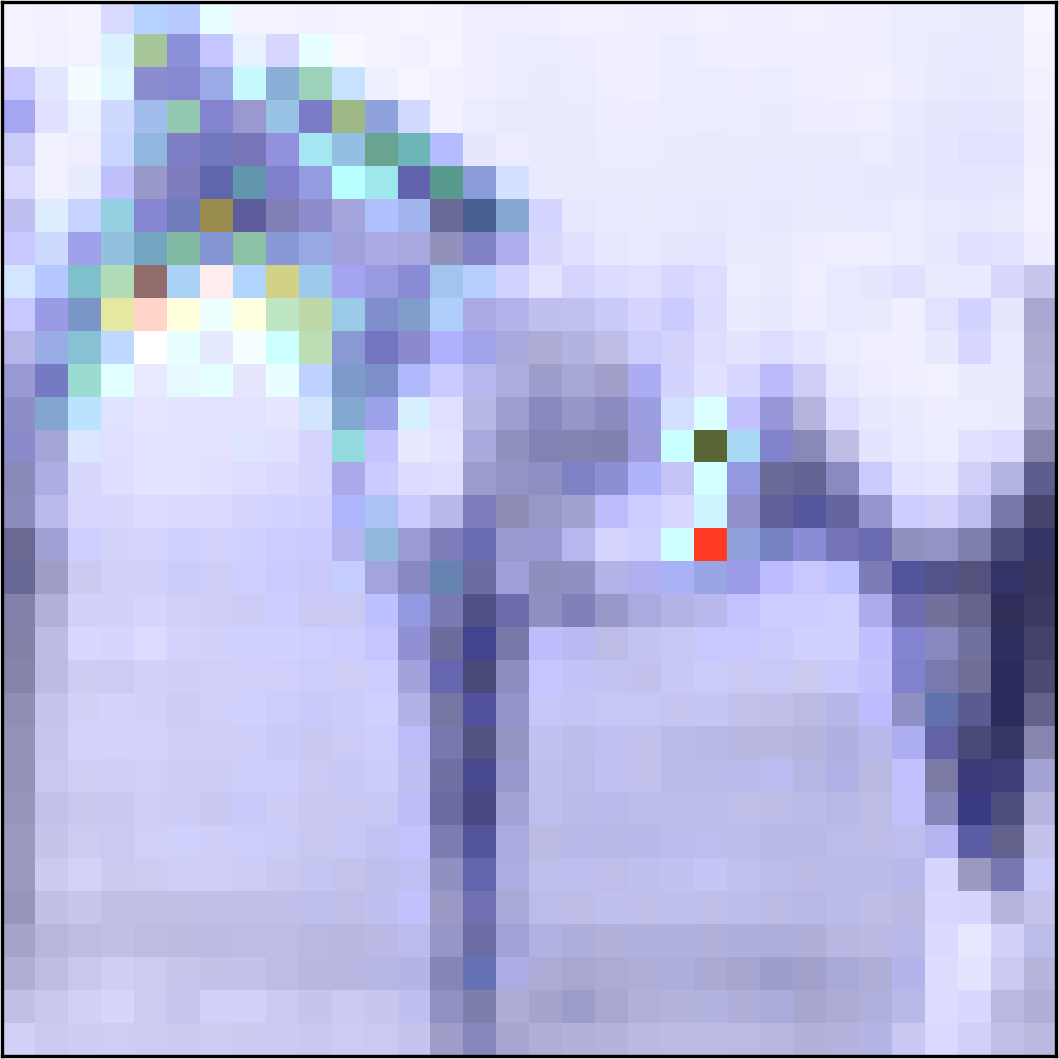} \\
    \vspace{0.1cm}\small{\hspace{0.5cm} $x$ \hspace{1.2cm} $SM_{C}$ \hspace{1.cm} $\hat{x}$\hspace{1.3cm} $\widehat{SM}_{C}$}\\
    \includegraphics[width=0.2\columnwidth]{images/cropped_PixelAttack-test_46_saliency_OI.png} 
    \includegraphics[width=0.2\columnwidth]{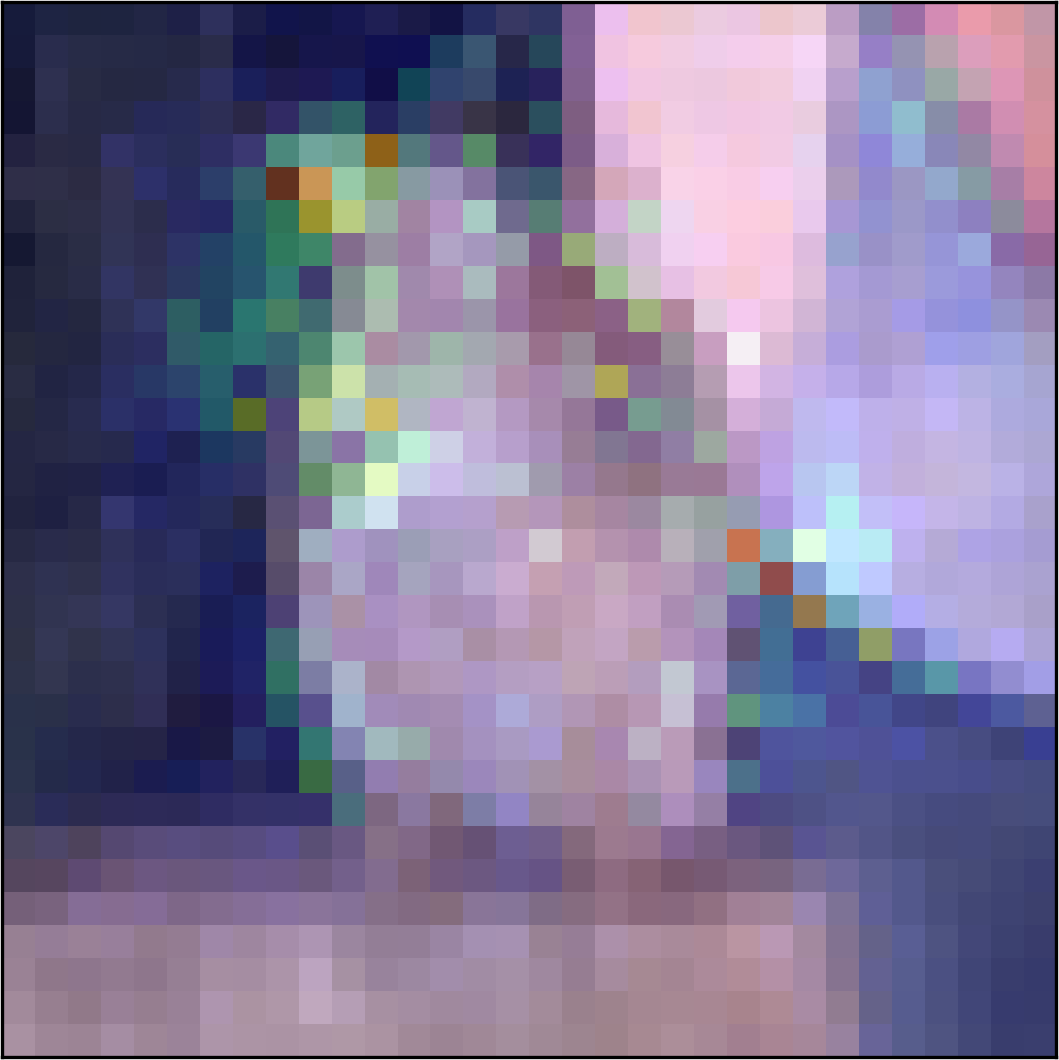} 
    \includegraphics[width=0.2\columnwidth]{images/cropped_PixelAttack-test_46_saliency_AI.png} 
    \includegraphics[width=0.2\columnwidth]{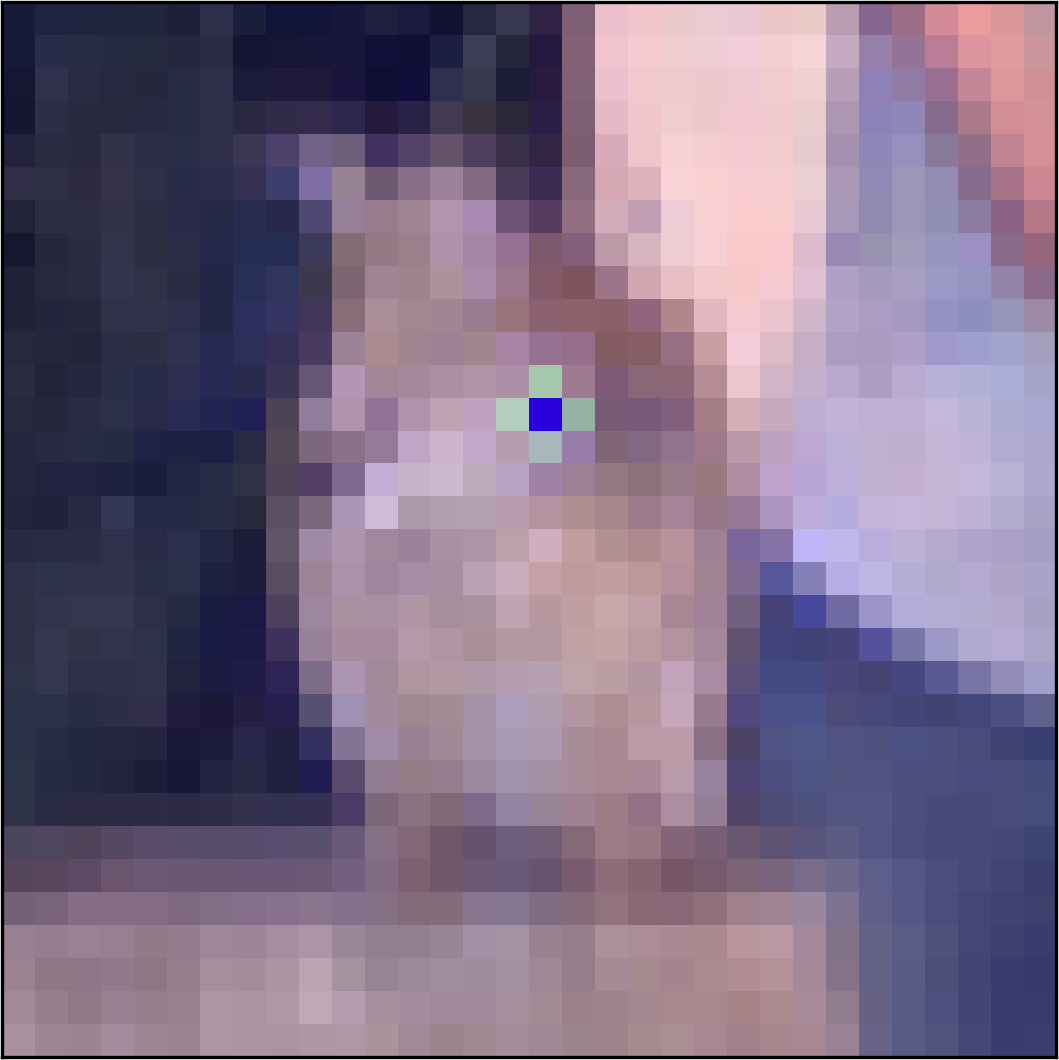} \\
    \vspace{0.1cm}
    \textbf{Projected Gradient Descent Attack}\\
    \vspace{0.1cm}
    \hrule
    \vspace{0.1cm}\small{\hspace{0.5cm} $x$ \hspace{1.2cm} $SM_{C}$ \hspace{1.cm} $\hat{x}$\hspace{1.3cm} $\widehat{SM}_{C}$}\\
    \includegraphics[width=0.2\columnwidth]{images/cropped_PGD_135_saliency_OI.png} 
    \includegraphics[width=0.2\columnwidth]{images/cropped_PGD_135_saliency_OST.png} 
    \includegraphics[width=0.2\columnwidth]{images/cropped_PGD_135_saliency_AI.png} 
    \includegraphics[width=0.2\columnwidth]{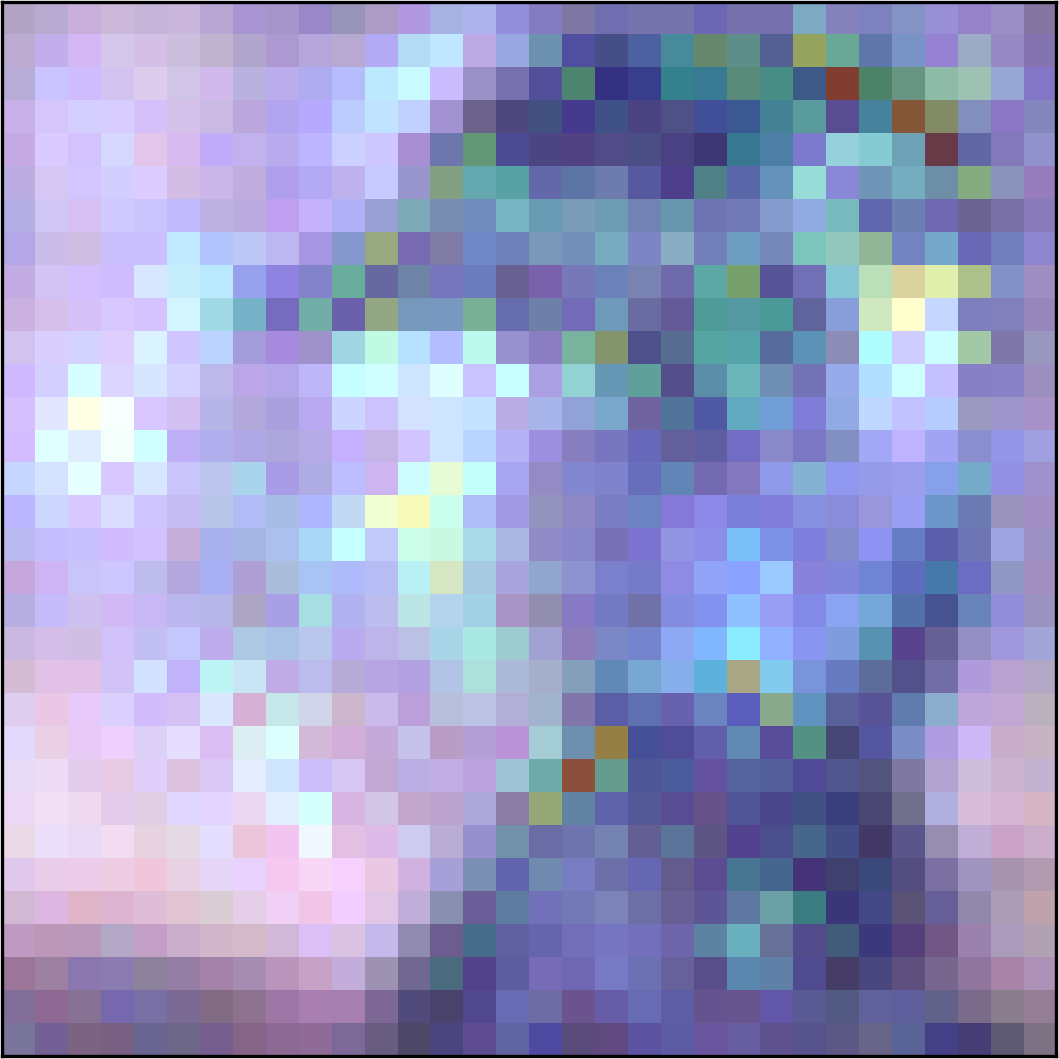} \\
    \vspace{0.1cm}\small{\hspace{0.5cm} $x$ \hspace{1.2cm} $SM_{C}$ \hspace{1.cm} $\hat{x}$\hspace{1.3cm} $\widehat{SM}_{C}$}\\
    \includegraphics[width=0.2\columnwidth]{images/cropped_PGD_8_saliency_OI.png} 
    \includegraphics[width=0.2\columnwidth]{images/cropped_PGD_8_saliency_OST.png} 
    \includegraphics[width=0.2\columnwidth]{images/cropped_PGD_8_saliency_AI.png} 
    \includegraphics[width=0.2\columnwidth]{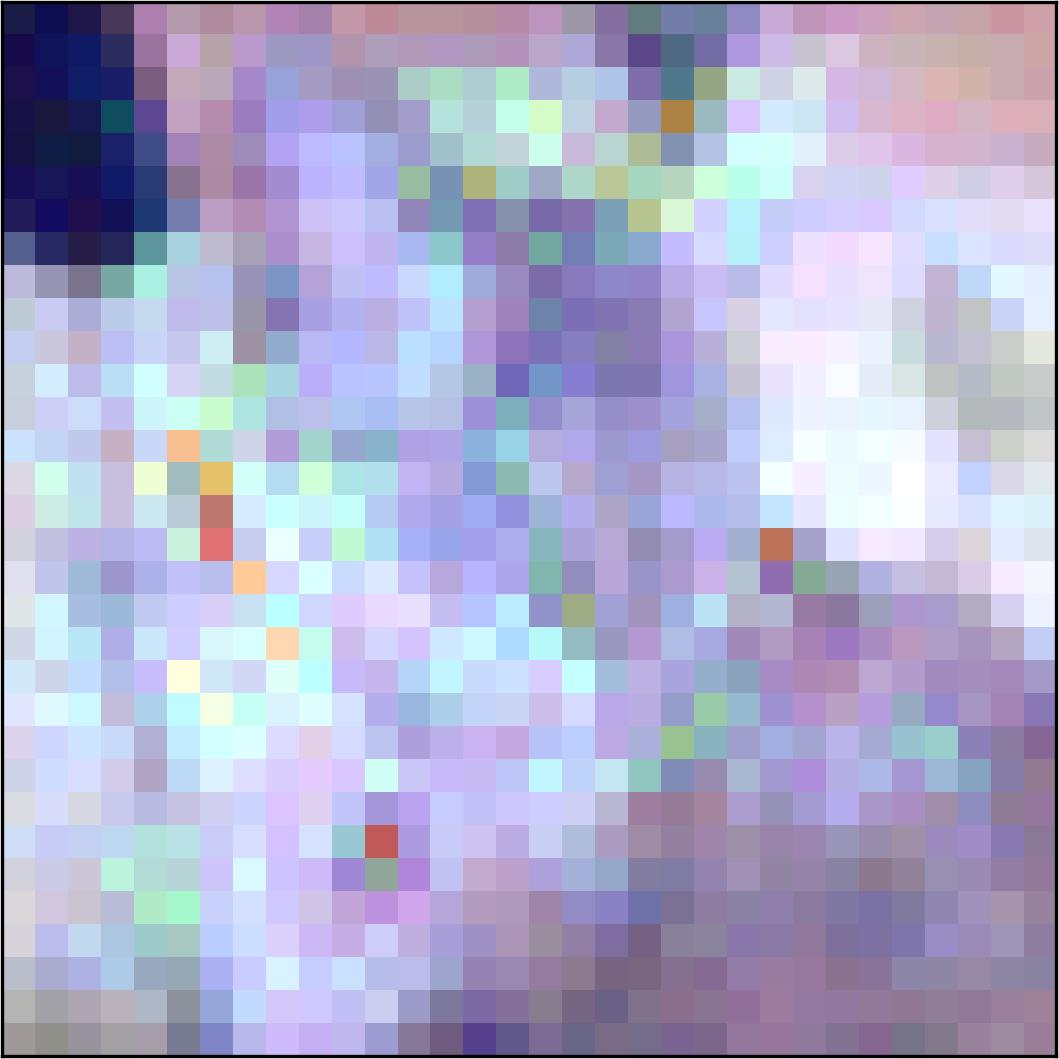} \\

    \caption{Overlayed Saliency Maps with respect to true/correct class.}
    \label{saliency_true} 
\end{figure}

\begin{figure}[!htb]
    \centering
    \textbf{Pixel Attack}\\
    \vspace{0.1cm}
    \hrule
    \vspace{0.1cm}\small{\hspace{0.5cm} $x$ \hspace{1.2cm} $SM_{\hat{C}}$ \hspace{1.cm} $\hat{x}$\hspace{1.3cm} $\widehat{SM}_{\hat{C}}$}\\
    \includegraphics[width=0.2\columnwidth]{images/cropped_PixelAttack-test_10_saliency_OI.png} 
    \includegraphics[width=0.2\columnwidth]{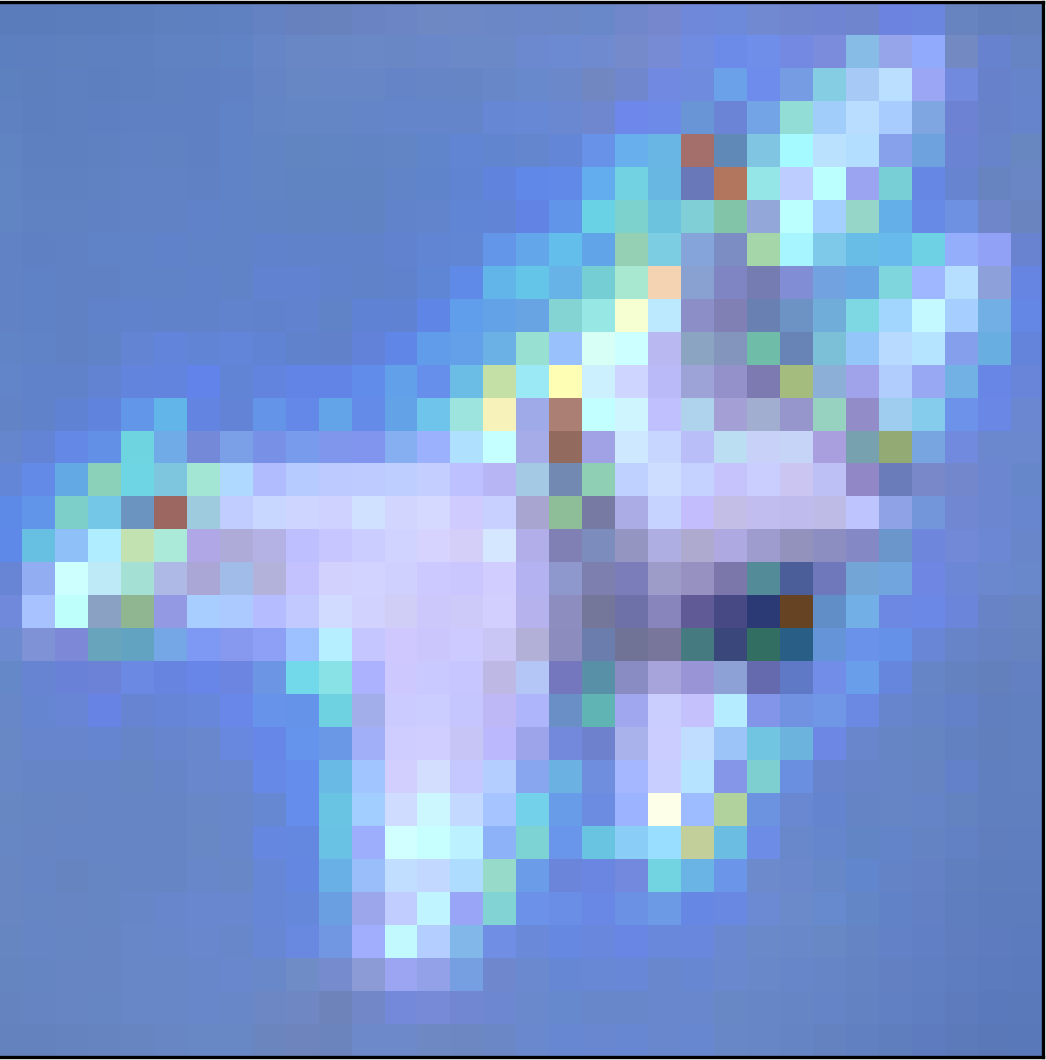} 
    \includegraphics[width=0.2\columnwidth]{images/cropped_PixelAttack-test_10_saliency_AI.png} 
    \includegraphics[width=0.2\columnwidth]{images/cropped_PixelAttack-test_10_saliency_AIM.png} \\
    \vspace{0.1cm}\small{\hspace{0.5cm} $x$ \hspace{1.2cm} $SM_{\hat{C}}$ \hspace{1.cm} $\hat{x}$\hspace{1.3cm} $\widehat{SM}_{\hat{C}}$}\\
    \includegraphics[width=0.2\columnwidth]{images/cropped_PixelAttack-test_83_saliency_OI.png} 
    \includegraphics[width=0.2\columnwidth]{images/cropped_PixelAttack-test_83_saliency_OSM.png} 
    \includegraphics[width=0.2\columnwidth]{images/cropped_PixelAttack-test_83_saliency_AI.png} 
    \includegraphics[width=0.2\columnwidth]{images/cropped_PixelAttack-test_83_saliency_AIM.png} \\
    \vspace{0.1cm}\small{\hspace{0.5cm} $x$ \hspace{1.2cm} $SM_{\hat{C}}$ \hspace{1.cm} $\hat{x}$\hspace{1.3cm} $\widehat{SM}_{\hat{C}}$}\\
    \includegraphics[width=0.2\columnwidth]{images/cropped_PixelAttack-test_46_saliency_OI.png} 
    \includegraphics[width=0.2\columnwidth]{images/cropped_PixelAttack-test_46_saliency_OSM.png} 
    \includegraphics[width=0.2\columnwidth]{images/cropped_PixelAttack-test_46_saliency_AI.png} 
    \includegraphics[width=0.2\columnwidth]{images/cropped_PixelAttack-test_46_saliency_AIM.png} \\
    \vspace{0.1cm}
    \textbf{Projected Gradient Descent Attack}\\
    \vspace{0.1cm}
    \hrule
    \vspace{0.1cm}\small{\hspace{0.5cm} $x$ \hspace{1.2cm} $SM_{\hat{C}}$ \hspace{1.cm} $\hat{x}$\hspace{1.3cm} $\widehat{SM}_{\hat{C}}$}\\
    \includegraphics[width=0.2\columnwidth]{images/cropped_PGD_135_saliency_OI.png} 
    \includegraphics[width=0.2\columnwidth]{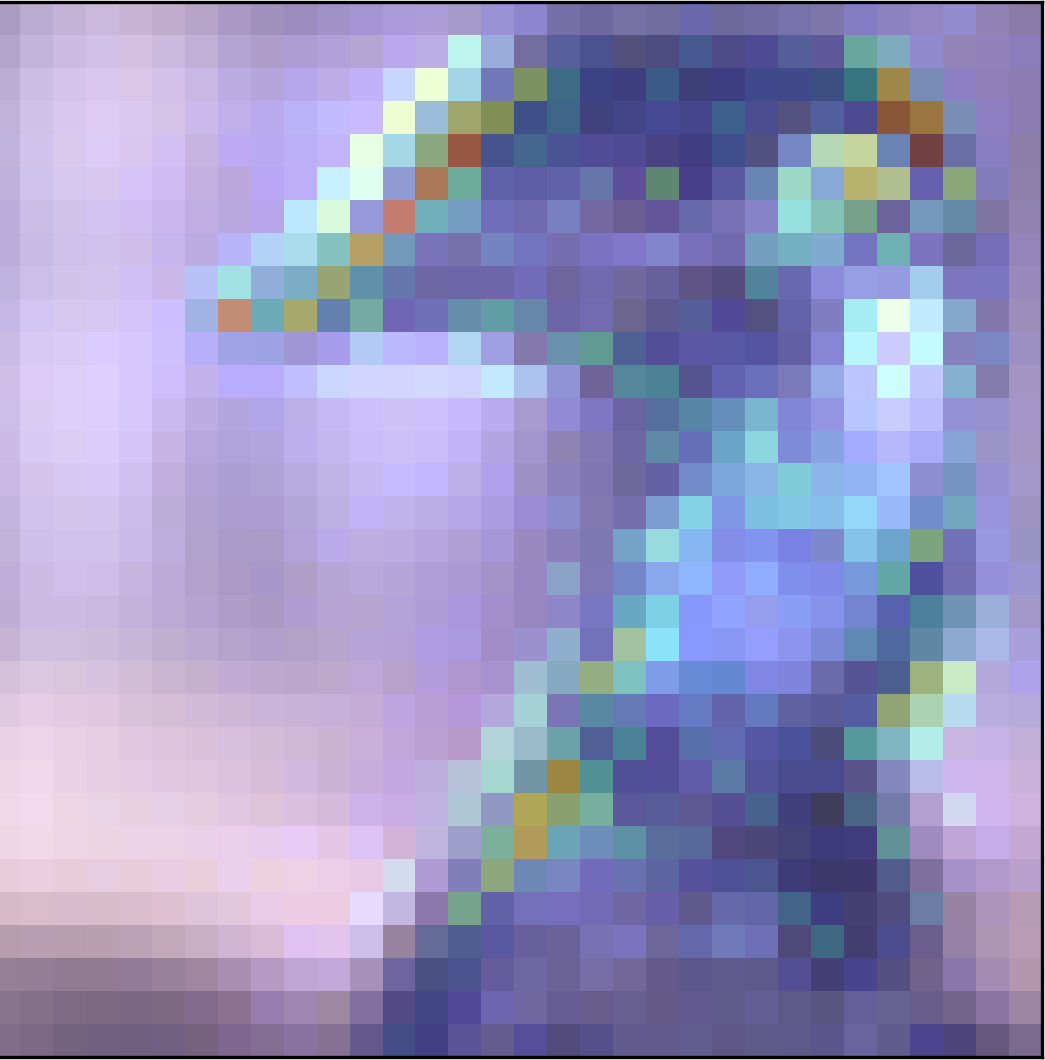} 
    \includegraphics[width=0.2\columnwidth]{images/cropped_PGD_135_saliency_AI.png} 
    \includegraphics[width=0.2\columnwidth]{images/cropped_PGD_135_saliency_AIM.png} \\
    \vspace{0.1cm}\small{\hspace{0.5cm} $x$ \hspace{1.2cm} $SM_{\hat{C}}$ \hspace{1.cm} $\hat{x}$\hspace{1.3cm} $\widehat{SM}_{\hat{C}}$}\\
    \includegraphics[width=0.2\columnwidth]{images/cropped_PGD_8_saliency_OI.png} 
    \includegraphics[width=0.2\columnwidth]{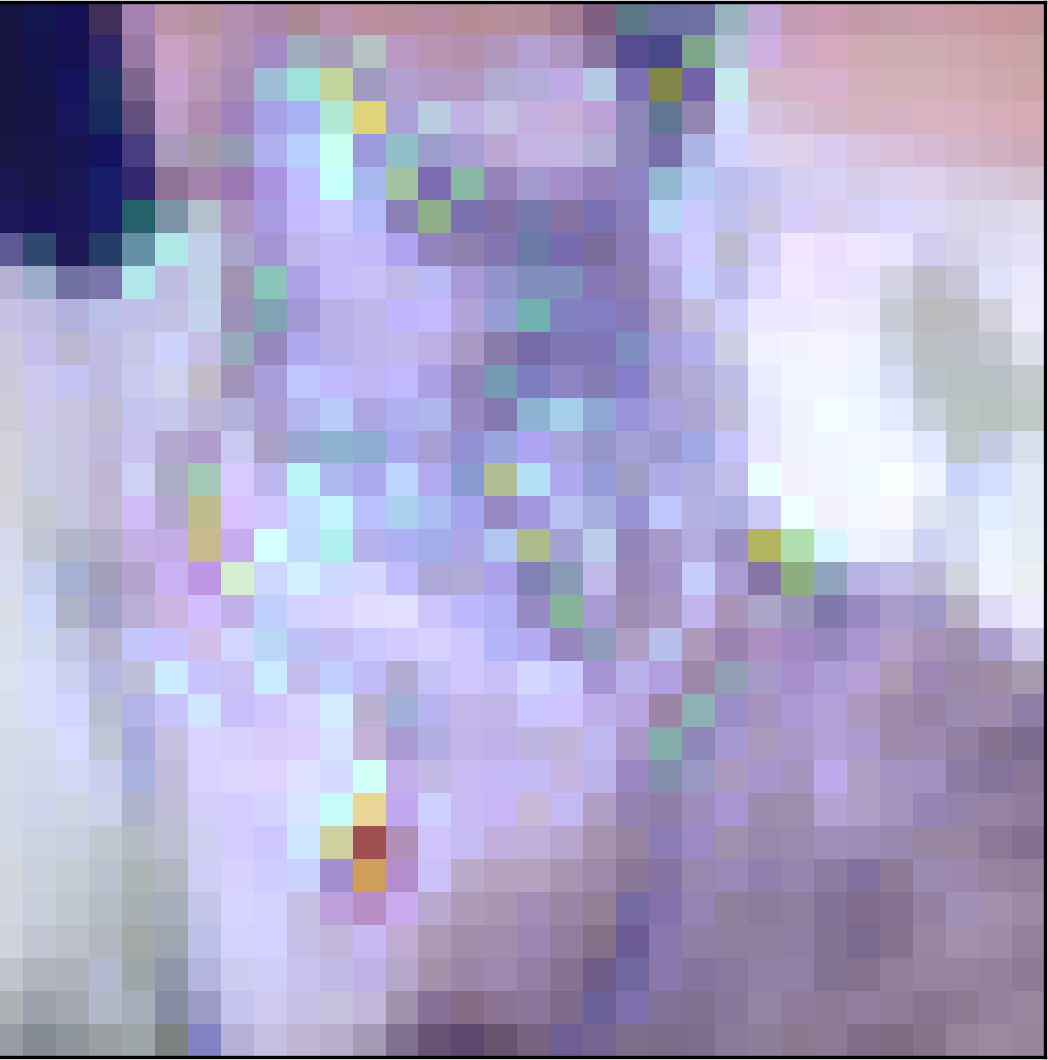} 
    \includegraphics[width=0.2\columnwidth]{images/cropped_PGD_8_saliency_AI.png} 
    \includegraphics[width=0.2\columnwidth]{images/cropped_PGD_8_saliency_AIM.png} \\

    \caption{Overlayed Saliency Maps with respect to adversarial/incorrect class.}
    \label{saliency_false} 
\end{figure}

We visualise the saliency maps of original sample $(SM)$ and saliency maps of adversarial sample $(\widehat{SM})$ with respect to predicted class of the model (Figure \ref{saliency_pred}), true/correct class $(C)$ (Figure \ref{saliency_true}), and adversarial/incorrect class $(\hat{C})$ (Figure \ref{saliency_false}) for different attacks. 
For original sample predicted class is true/correct class $(C)$ where as for adversarial sample predicted class adversarial/incorrect class $(\hat{C})$. 
Intuitively, adversarial perturbations should distort the saliency of the image as seen by the neural network to induce misclassification.
From the figures, it can be seen that both Pixel Attack and Projected Gradient Attack distorts the saliency differently.

We observe that, in the case of PGD Attack, the saliency of the image is diffused for the adversarial image created by the minimal change in pixels (Figure \ref{saliency_pred}). 
Moreover, we can visualise that PGD Attack effectively diffuses the attention around the region of interest.
Further, Figures \ref{saliency_true} and \ref{saliency_false} show that this diffusion occurs for both true and adversarial class. 
This shows that PGD Attack does not distort the region of interest determined by the neural network but only manipulating the activations inside layers such that the adversarial perturbation induces misclassification. 

We also observe some peculiar characteristics of pixel perturbations generated by Pixel Attack.
Differently from the Projected Gradient Descent Attack, we can see from the saliency maps (Figure \ref{saliency_pred}) that the attention of the neural network shifts to the perturbed region and not diffuses as in  Projected Gradient Descent Attack.
This diversion of attention in the neural network is the consequence of only introducing perturbed pixels.
Figure \ref{saliency_true} shows that the saliency map with respect to the true class changes minimally as the saliency in the adversarial image is in the same region as the original image. 
However, Figure \ref{saliency_false} shows that the saliency map changes with respect to the adversarial class and this distorted saliency dominate the attention with respect to the correct class.
This shows that perturbed pixels created by Pixel Attack call the neural network's attention towards them and diverts the network's attention towards the adversarial class. 
Thus, Pixel Attack induces misclassification by changing the region of interest for the neural network through adversarial perturbations.

Further, this effect of diversion in attention due to perturbed pixels in Pixel Attack is propagated through the network to intensify, which causes misclassification in the neural network, as shown by \cite{vargas2019understanding}. 
This further builds on the hypothesis of conflicting saliency as adversarial perturbations severely affect the saliency of the image determined by neural networks. 
This also shows that distortion in saliency proves that adversarial perturbations do not fool the neural network naively but diverts attention towards the perturbations.


\subsection{Gradient-weighted Class Activation Maps}

\begin{figure}[!t]
    \centering
    \textbf{Pixel Attack}\\
    \vspace{0.1cm}    
    \hrule
    \vspace{0.1cm}\small{\hspace{0.5cm} $x$ \hspace{1.2cm} $AM_{C}$ \hspace{1.cm} $\hat{x}$\hspace{1.3cm} $\widehat{AM}_{\hat{C}}$}\\
    \includegraphics[width=0.2\columnwidth]{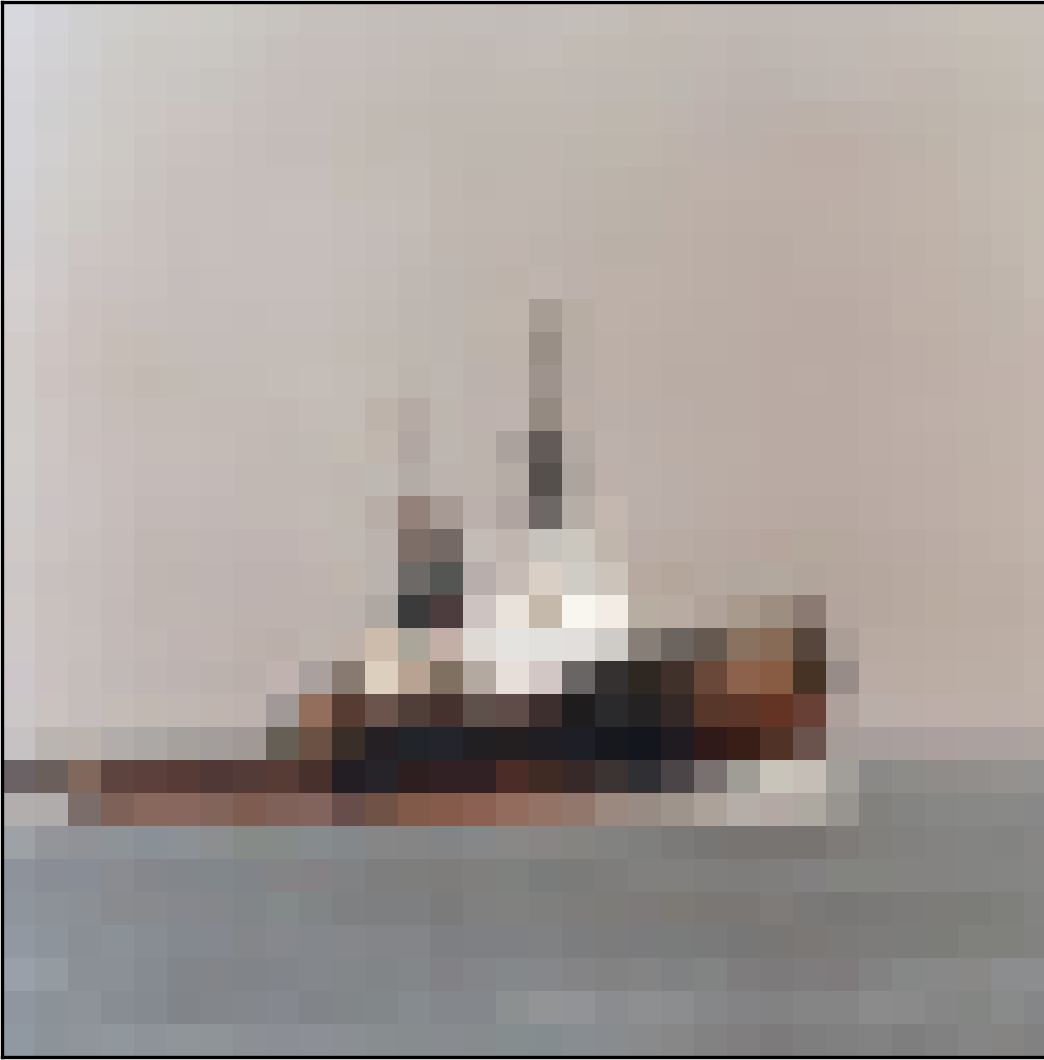} 
    \includegraphics[width=0.2\columnwidth]{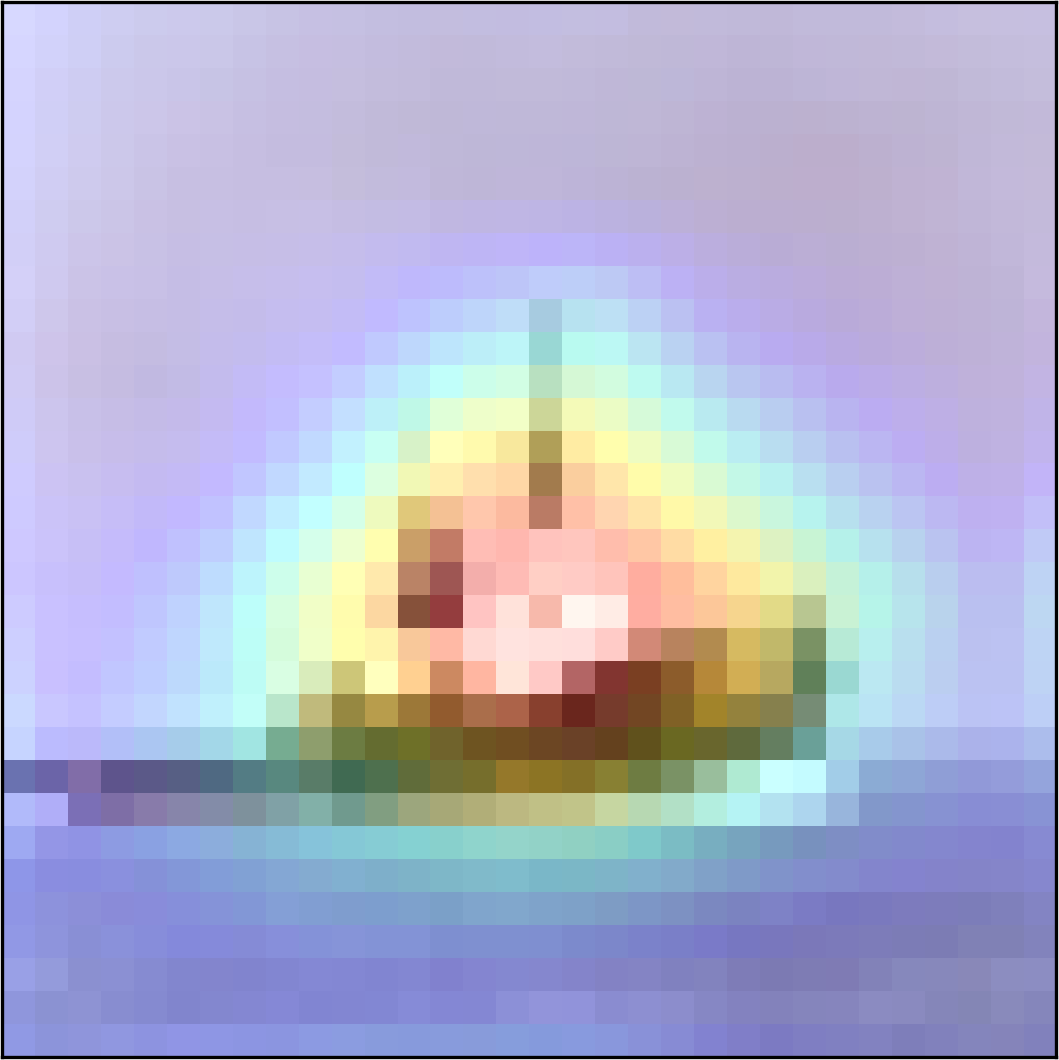} 
    \includegraphics[width=0.2\columnwidth]{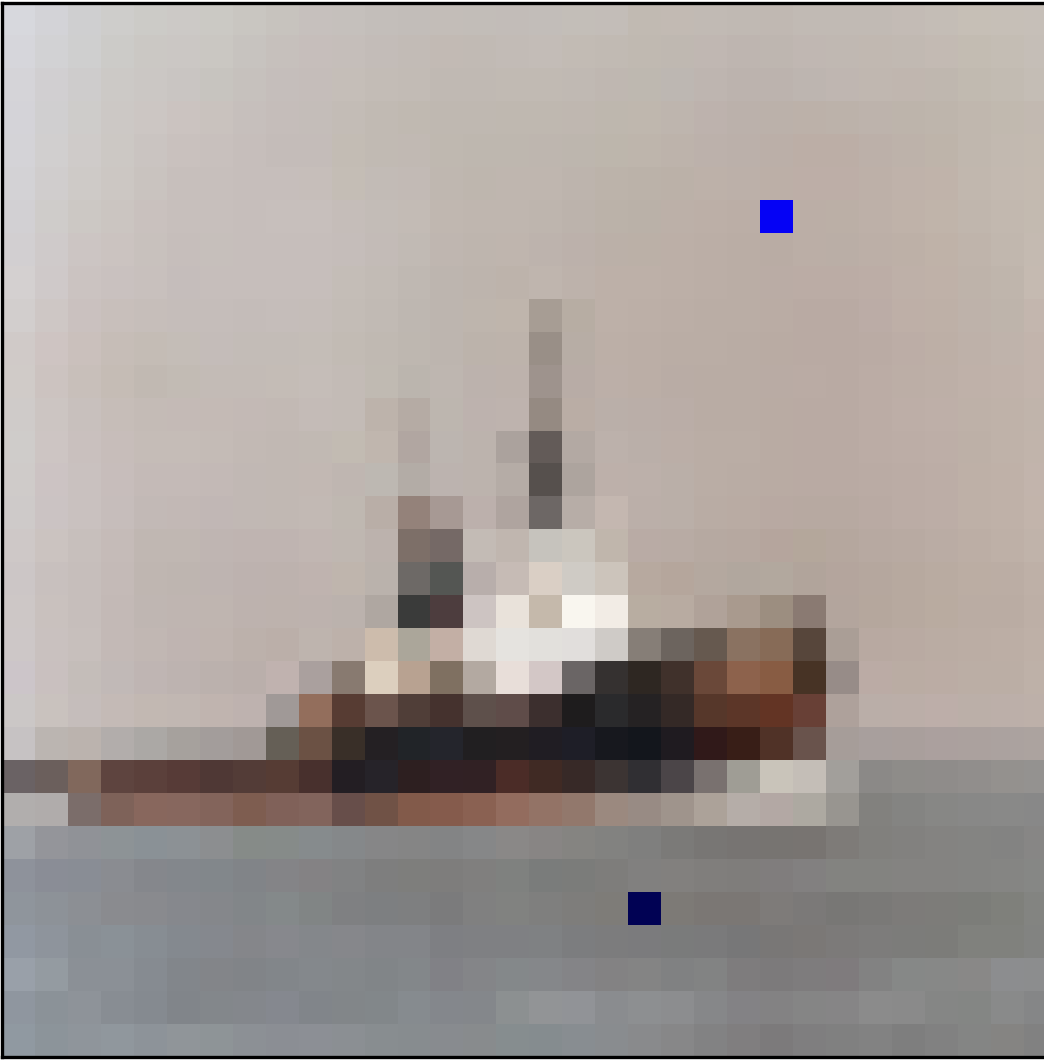} 
    \includegraphics[width=0.2\columnwidth]{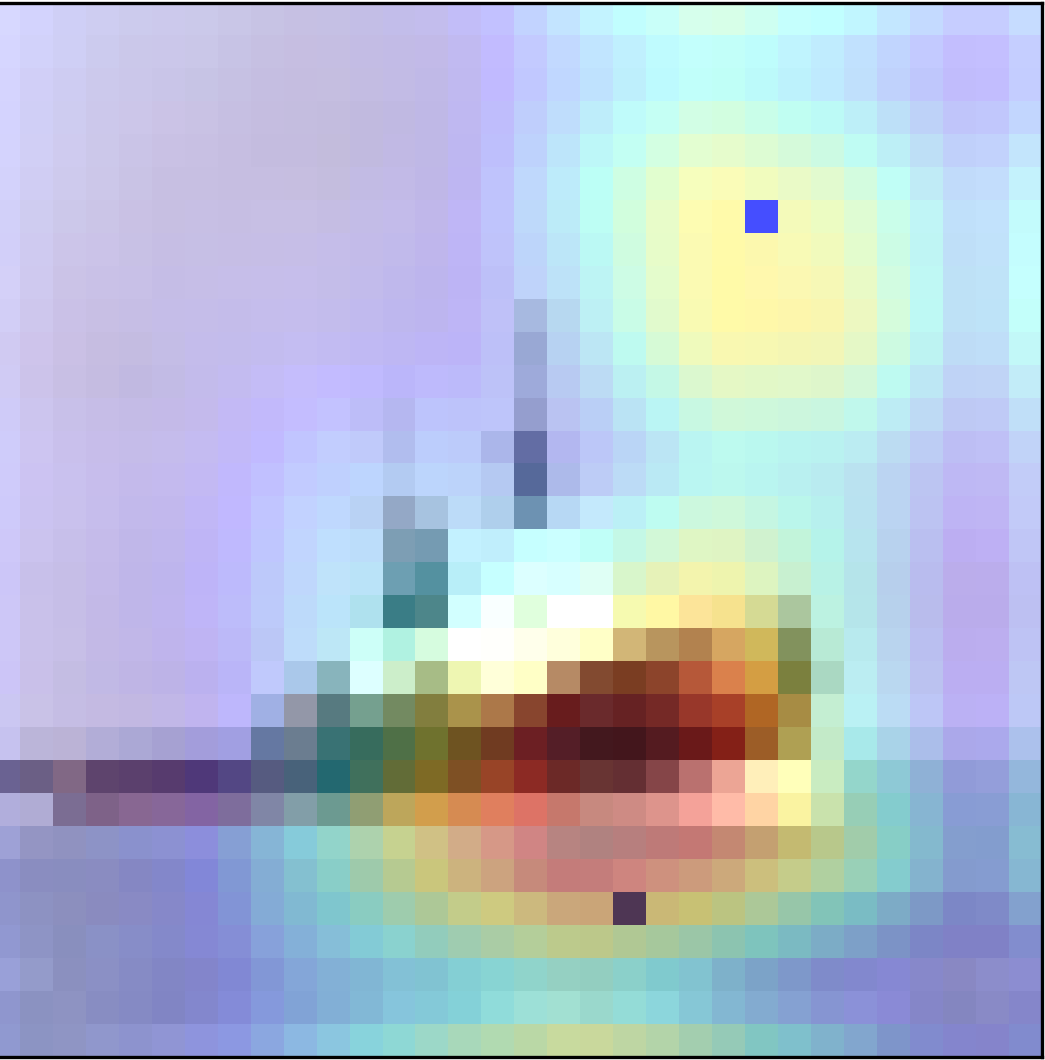} \\
    \vspace{0.1cm}\small{\hspace{0.5cm} $x$ \hspace{1.2cm} $AM_{C}$ \hspace{1.cm} $\hat{x}$\hspace{1.3cm} $\widehat{AM}_{\hat{C}}$}\\
    \includegraphics[width=0.2\columnwidth]{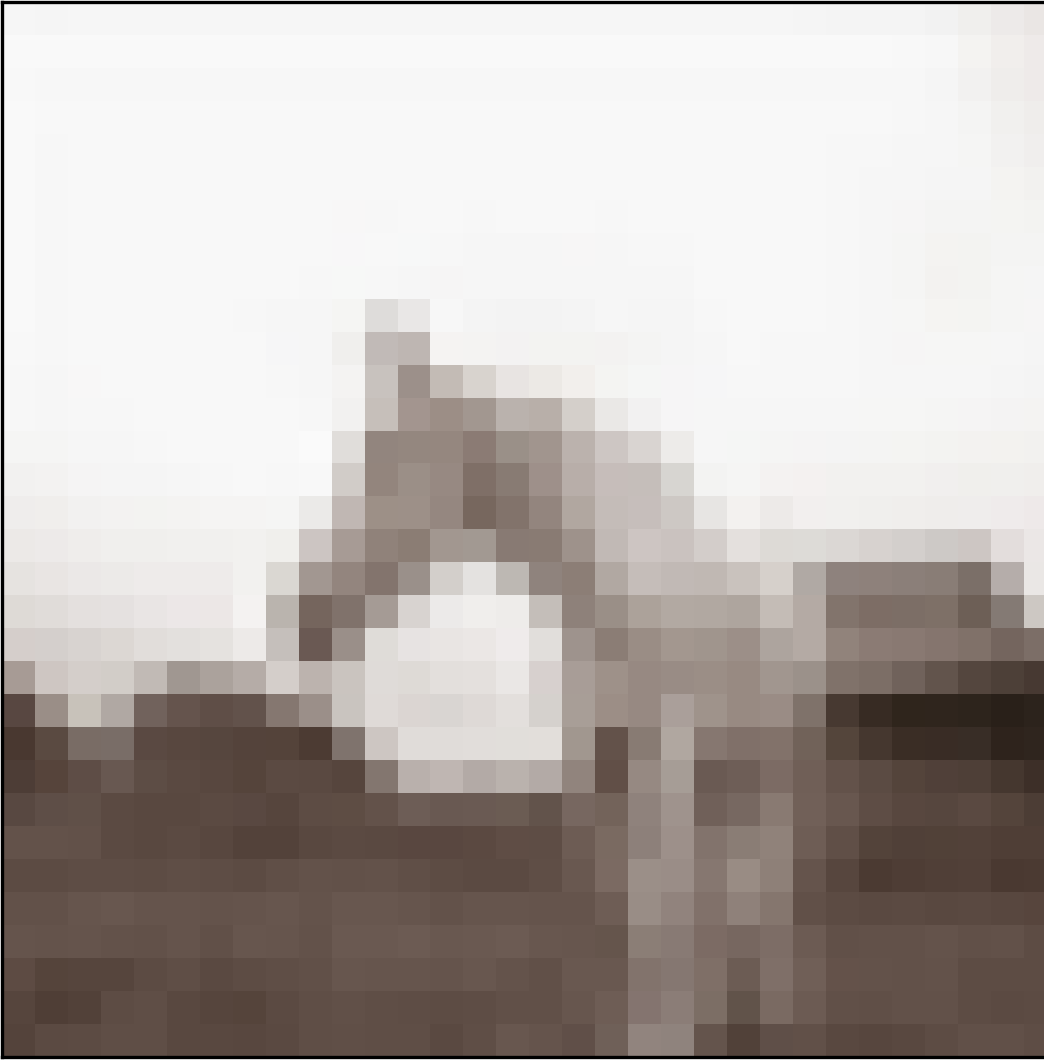} 
    \includegraphics[width=0.2\columnwidth]{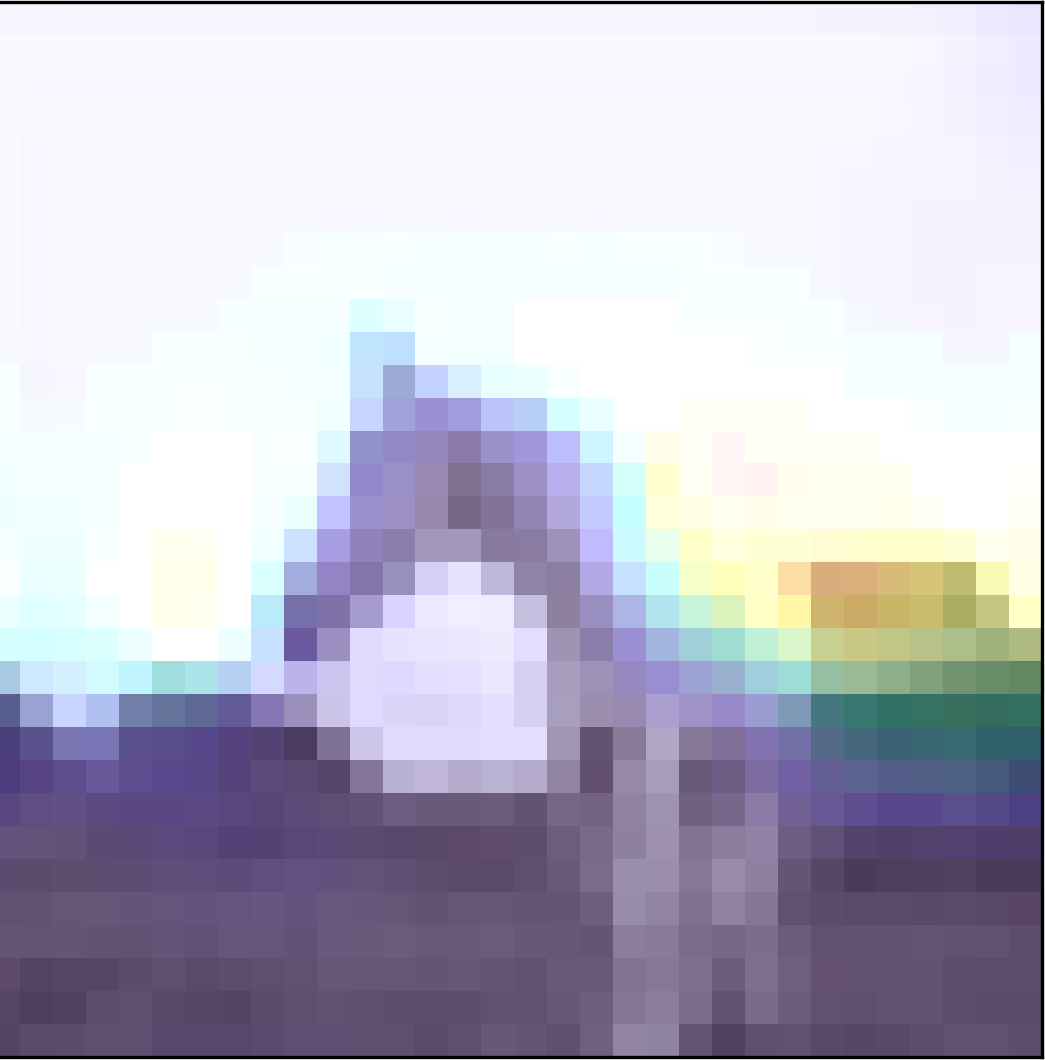} 
    \includegraphics[width=0.2\columnwidth]{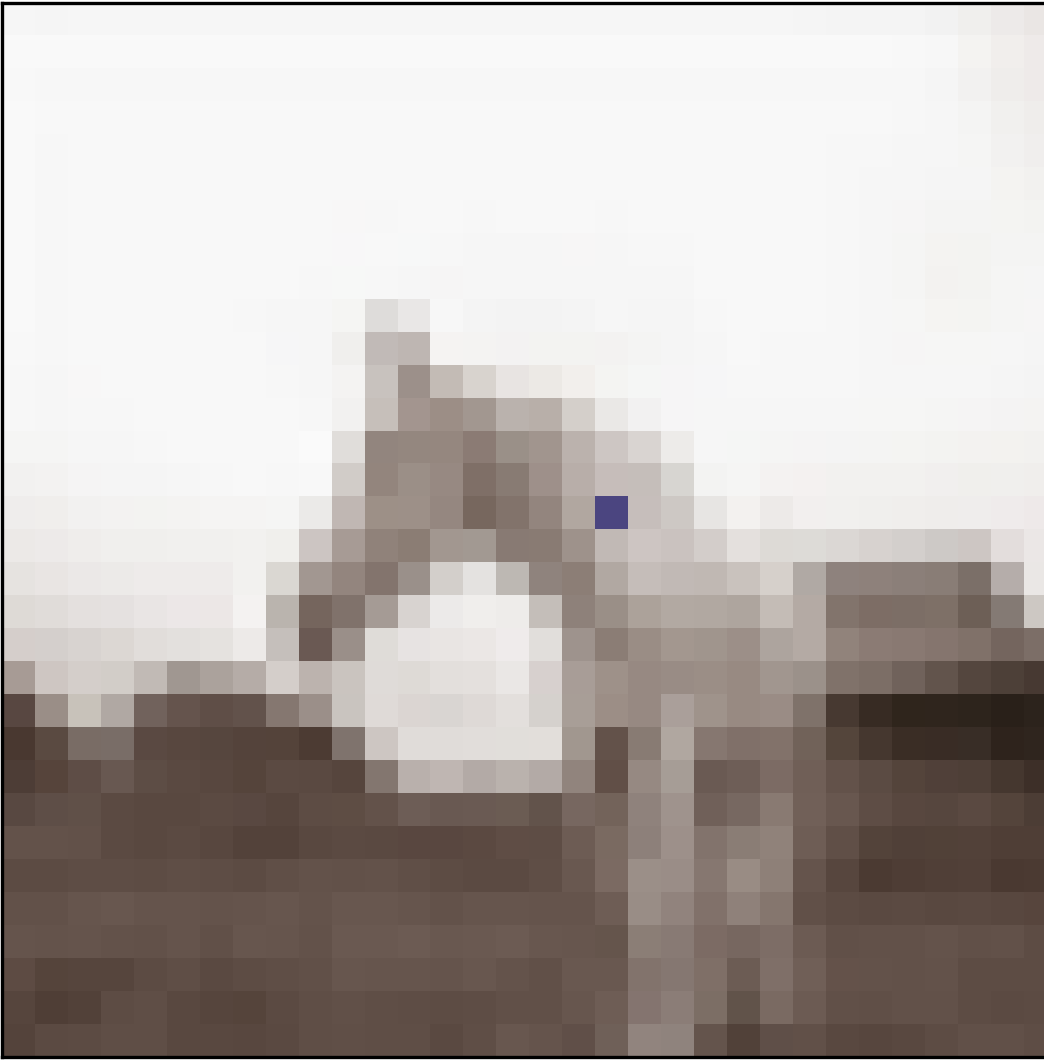} 
    \includegraphics[width=0.2\columnwidth]{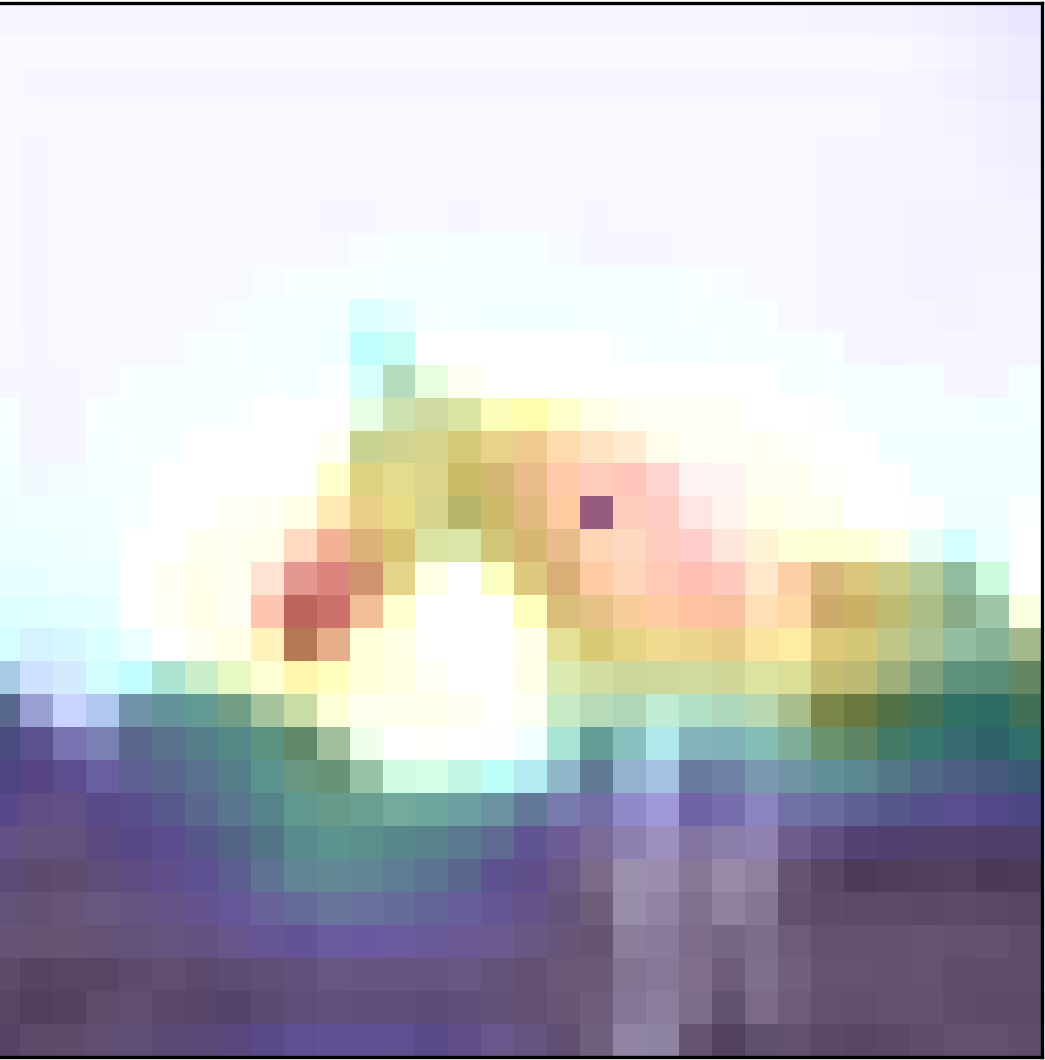} \\
    \vspace{0.1cm}\small{\hspace{0.5cm} $x$ \hspace{1.2cm} $AM_{C}$ \hspace{1.cm} $\hat{x}$\hspace{1.3cm} $\widehat{AM}_{\hat{C}}$}\\
    \includegraphics[width=0.2\columnwidth]{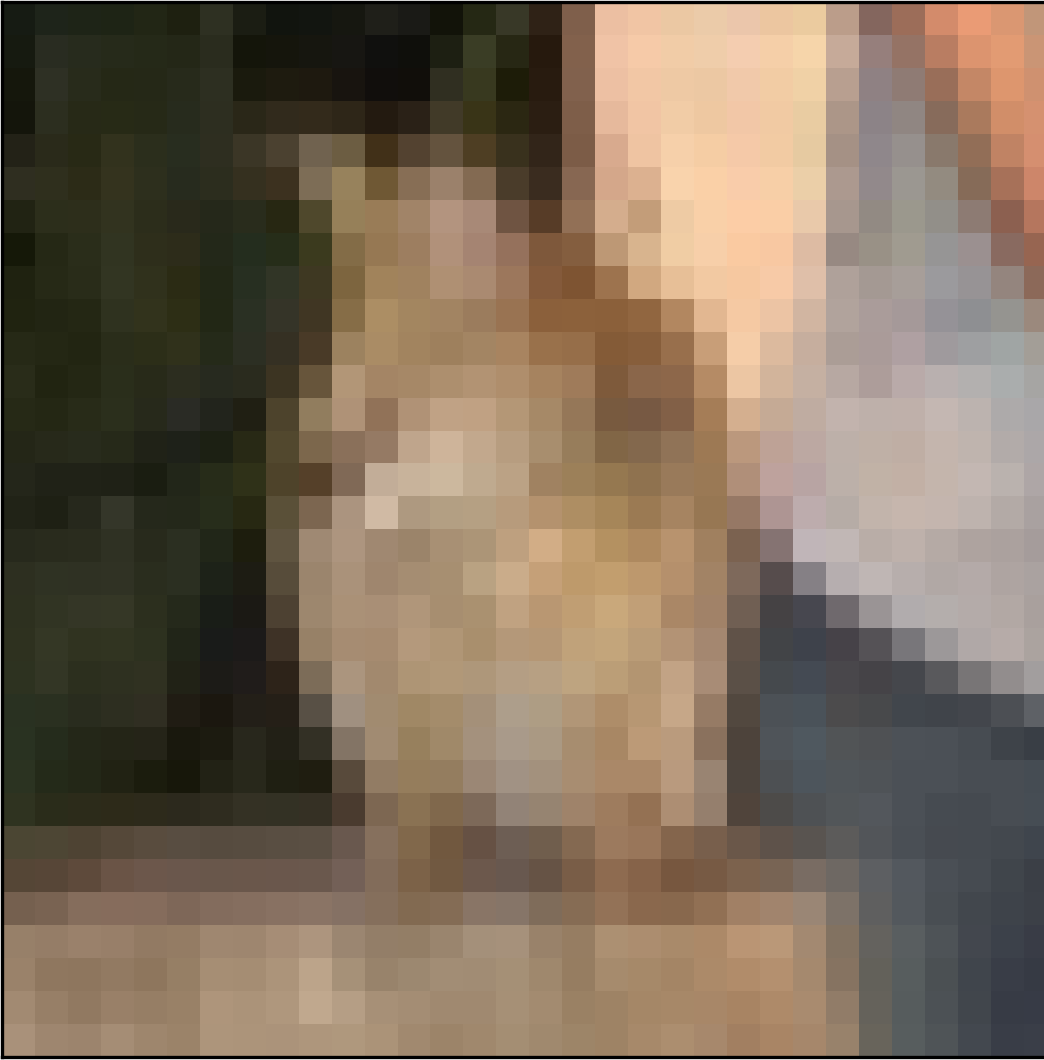} 
    \includegraphics[width=0.2\columnwidth]{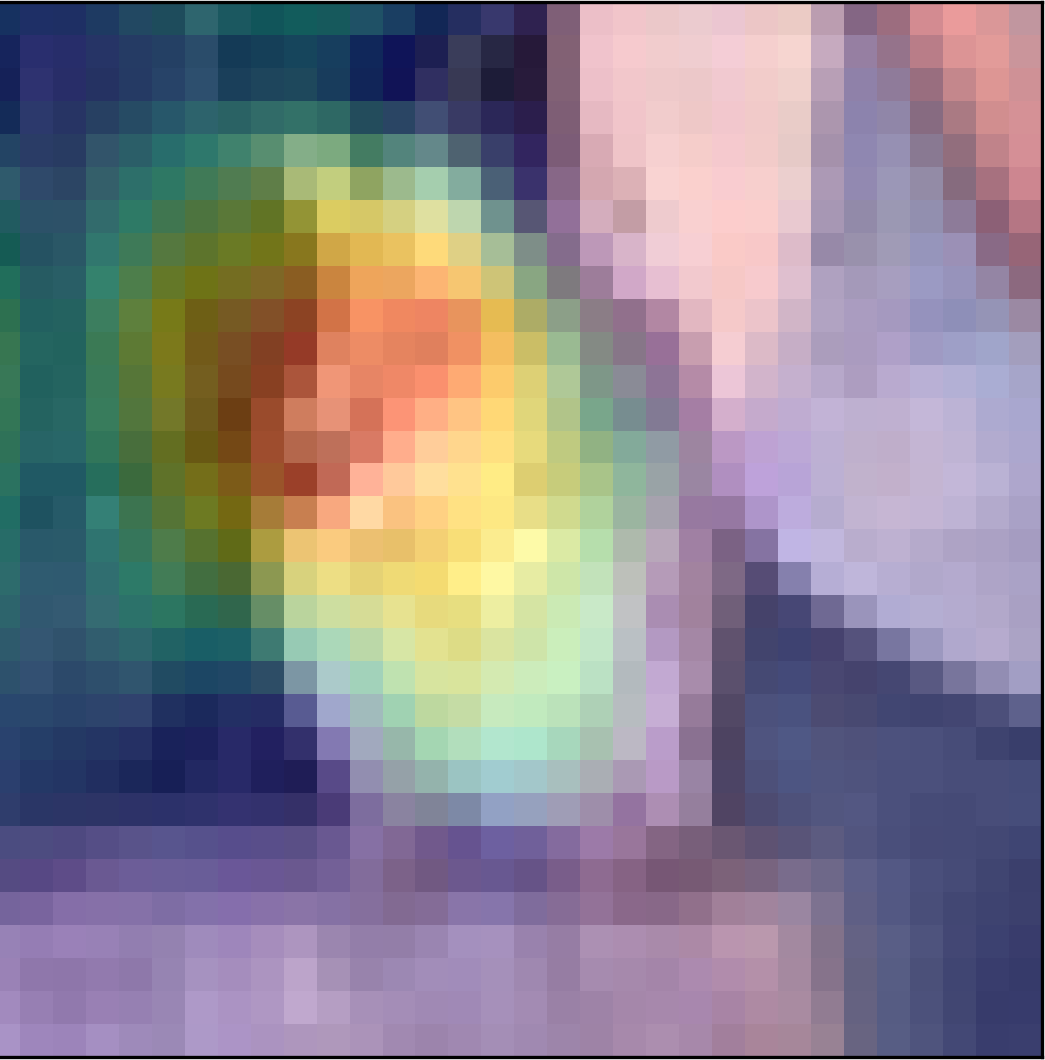} 
    \includegraphics[width=0.2\columnwidth]{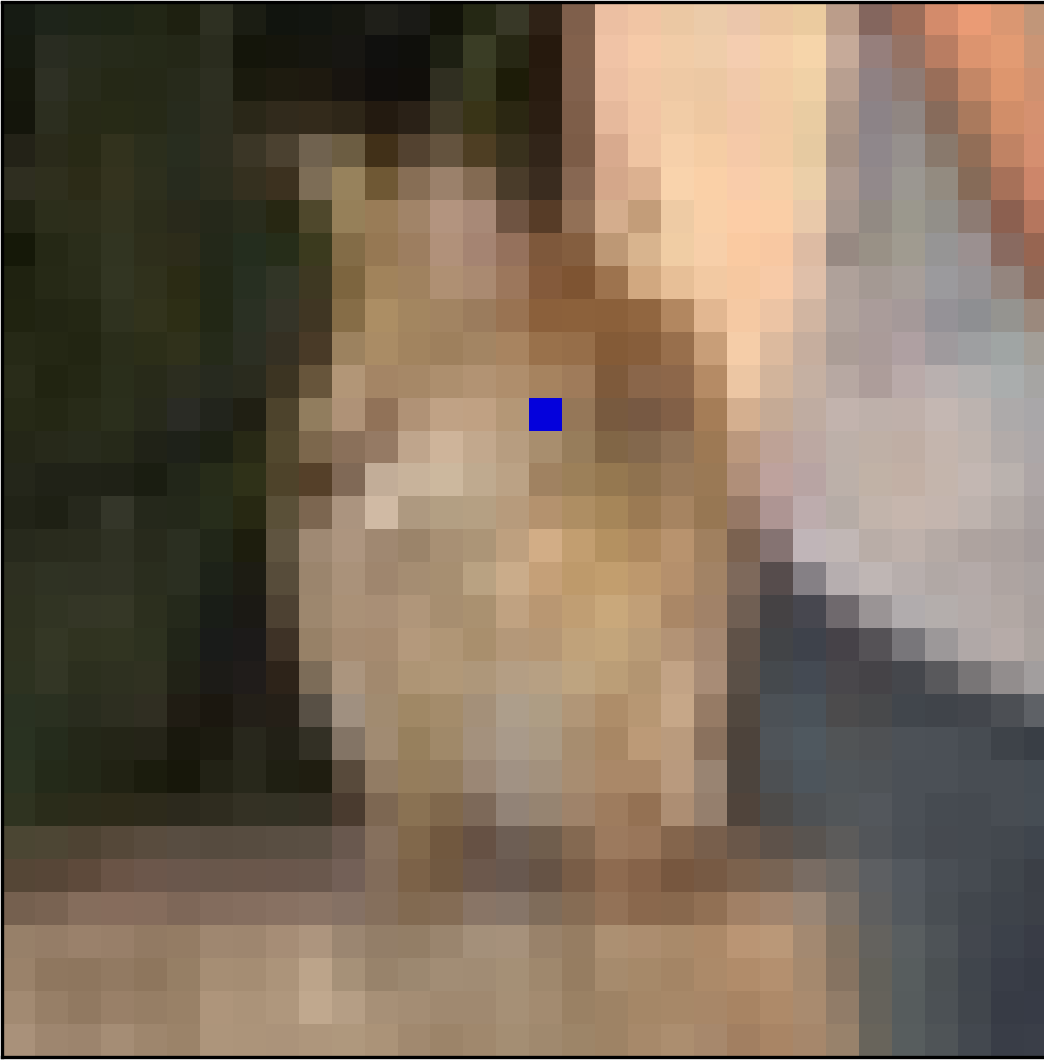} 
    \includegraphics[width=0.2\columnwidth]{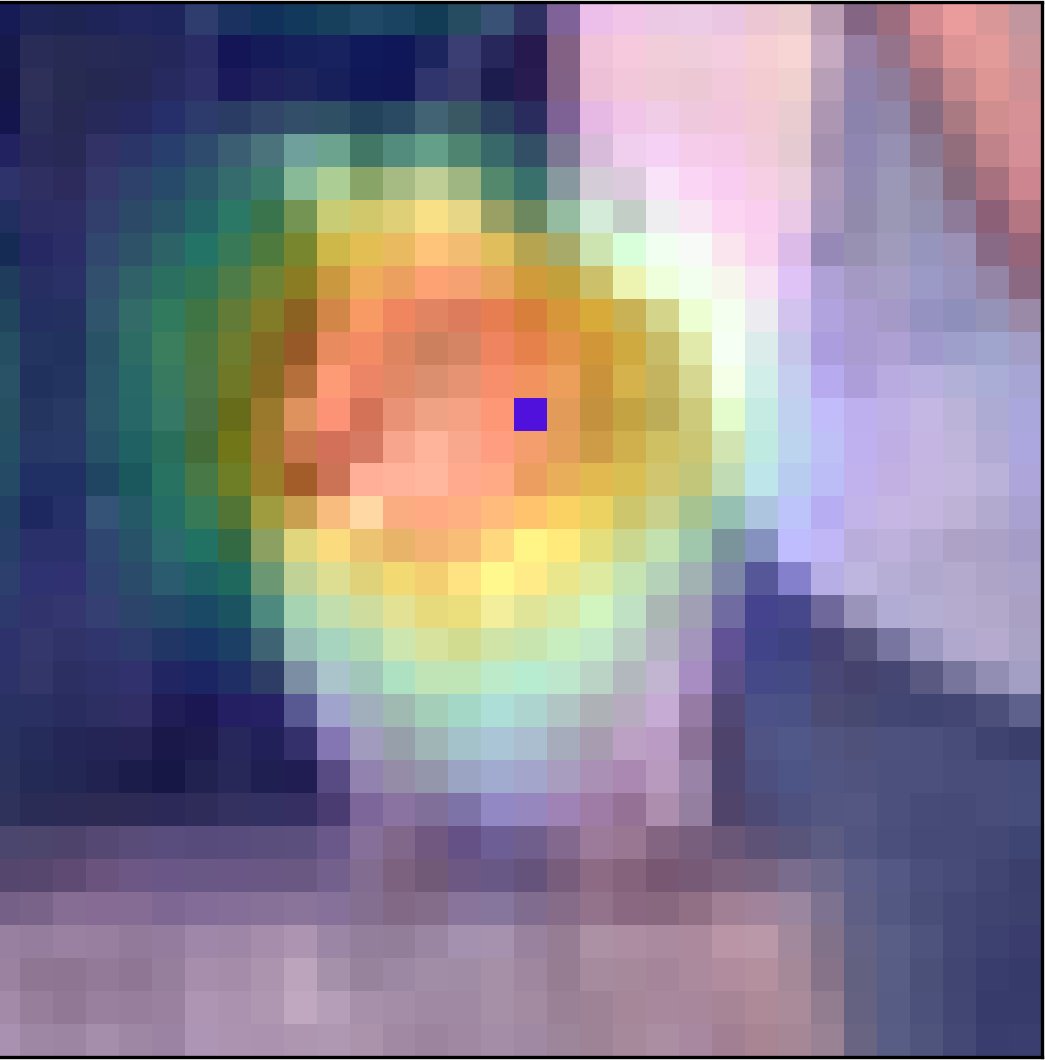} \\
    \vspace{0.1cm}
    \textbf{Projected Gradient Descent Attack}\\
    \vspace{0.1cm}
    \hrule
    \vspace{0.1cm}\small{\hspace{0.5cm} $x$ \hspace{1.2cm} $AM_{C}$ \hspace{1.cm} $\hat{x}$\hspace{1.3cm} $\widehat{AM}_{\hat{C}}$}\\
    \includegraphics[width=0.2\columnwidth]{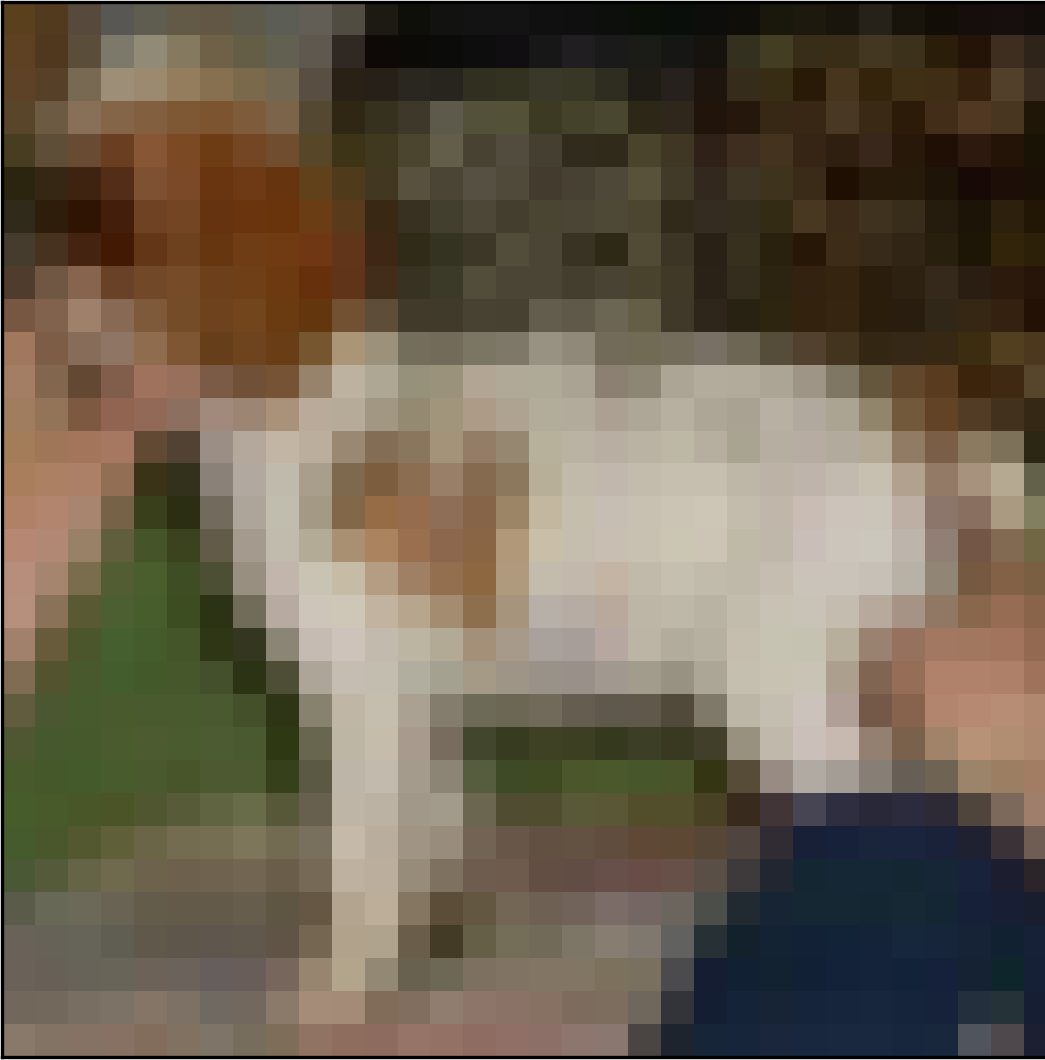} 
    \includegraphics[width=0.2\columnwidth]{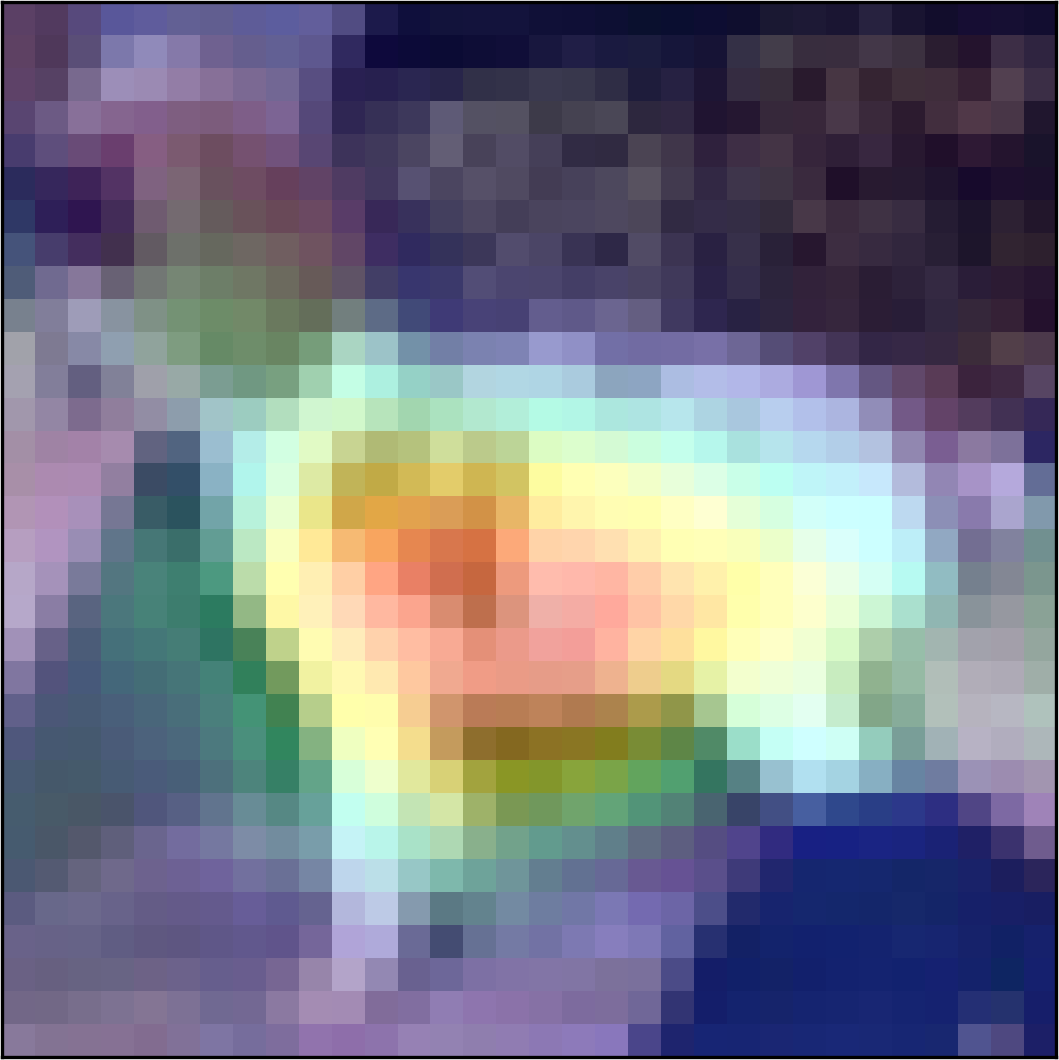} 
    \includegraphics[width=0.2\columnwidth]{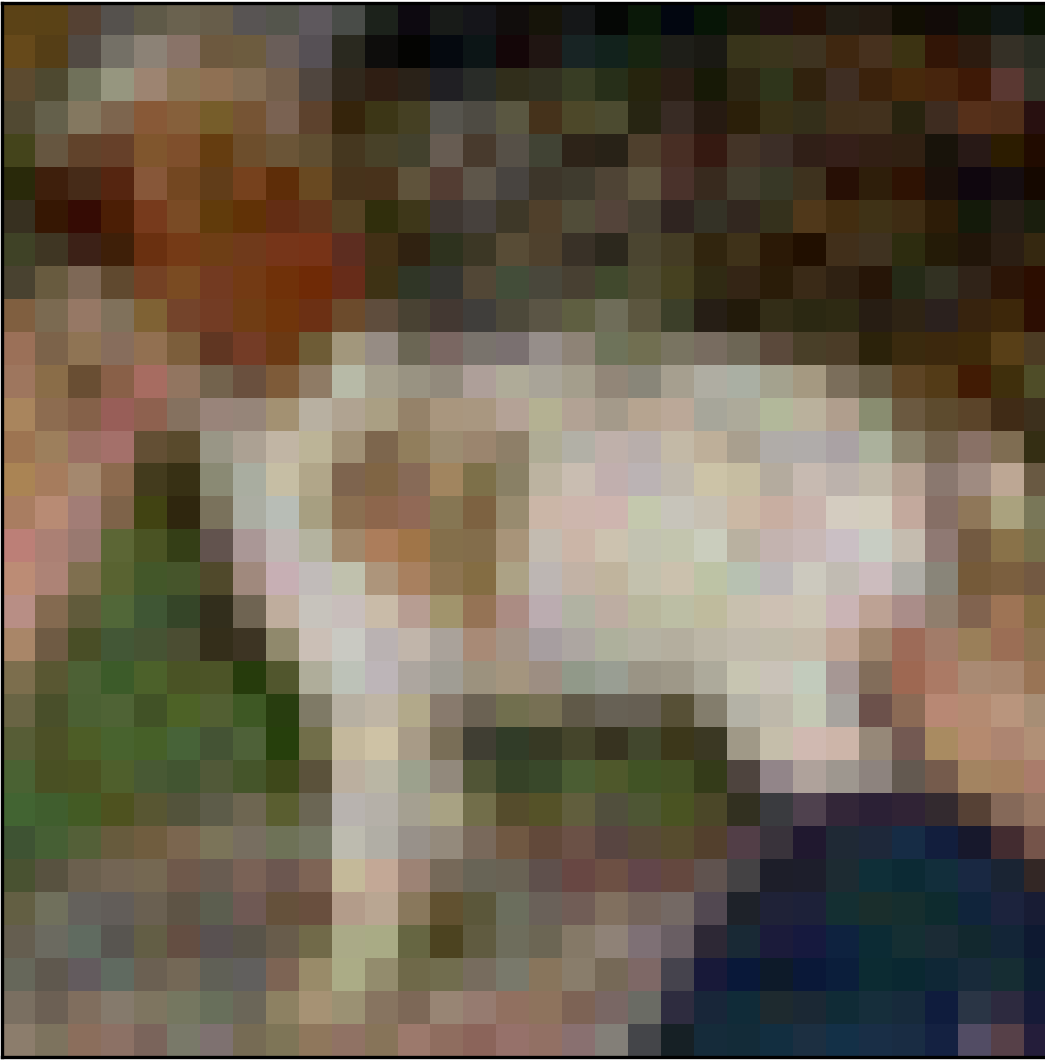} 
    \includegraphics[width=0.2\columnwidth]{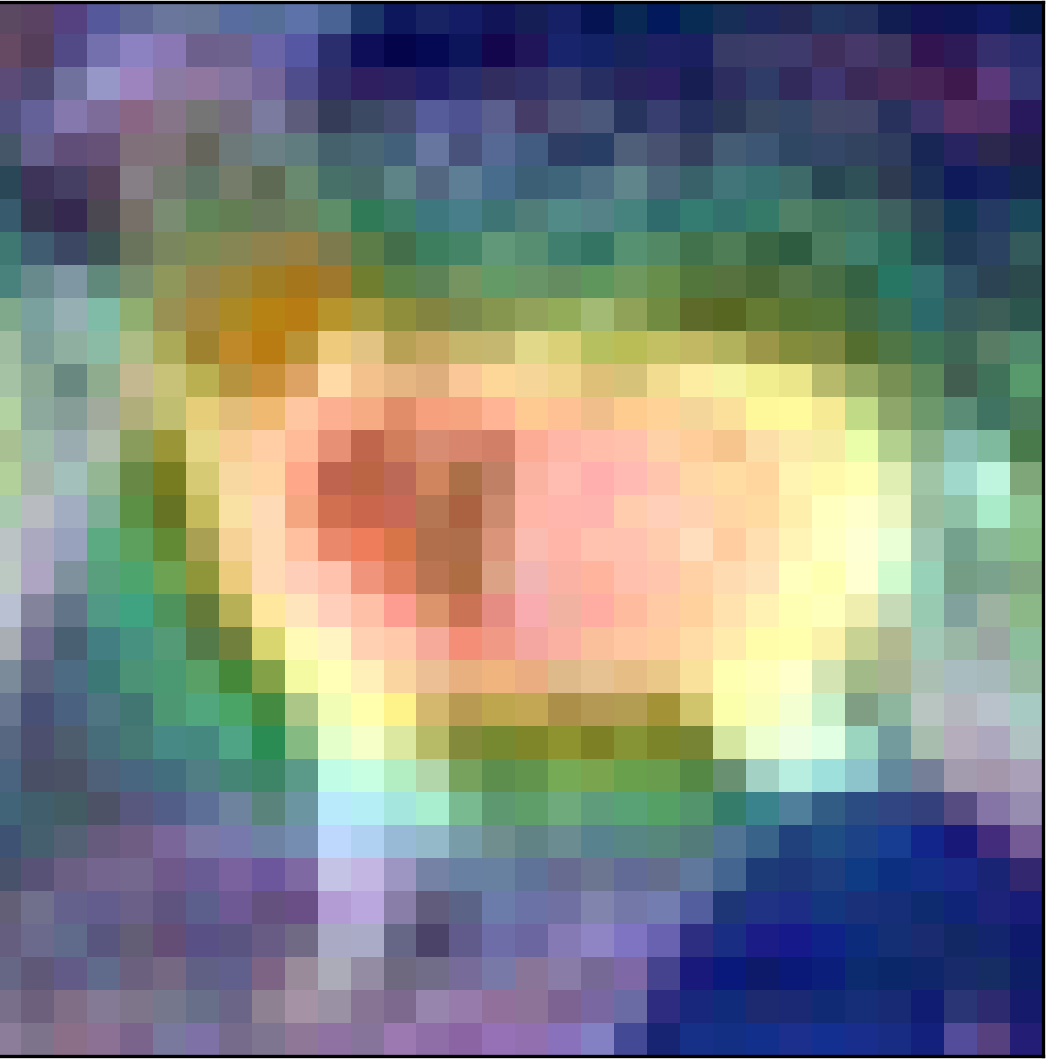} \\
    \vspace{0.1cm}\small{\hspace{0.5cm} $x$ \hspace{1.2cm} $AM_{C}$ \hspace{1.cm} $\hat{x}$\hspace{1.3cm} $\widehat{AM}_{\hat{C}}$}\\
    \includegraphics[width=0.2\columnwidth]{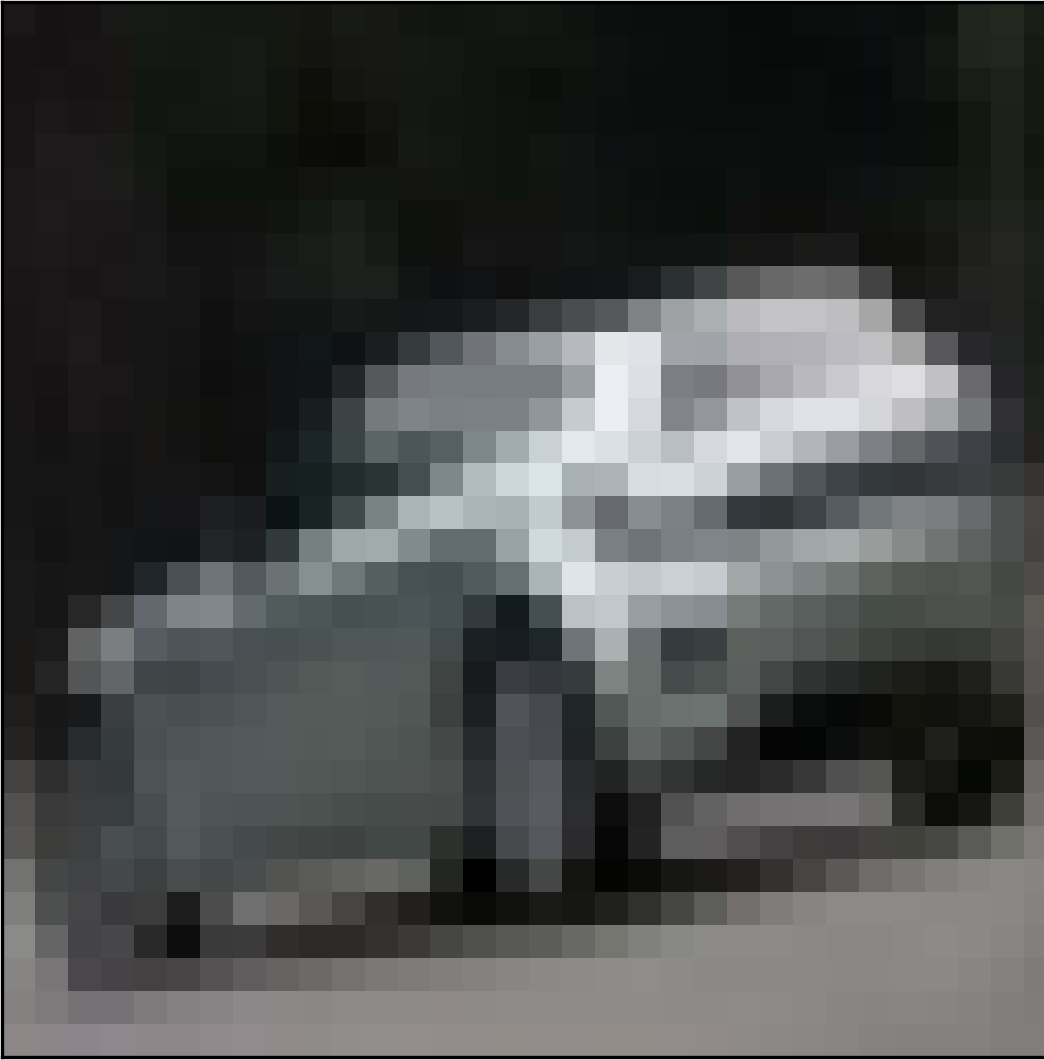} 
    \includegraphics[width=0.2\columnwidth]{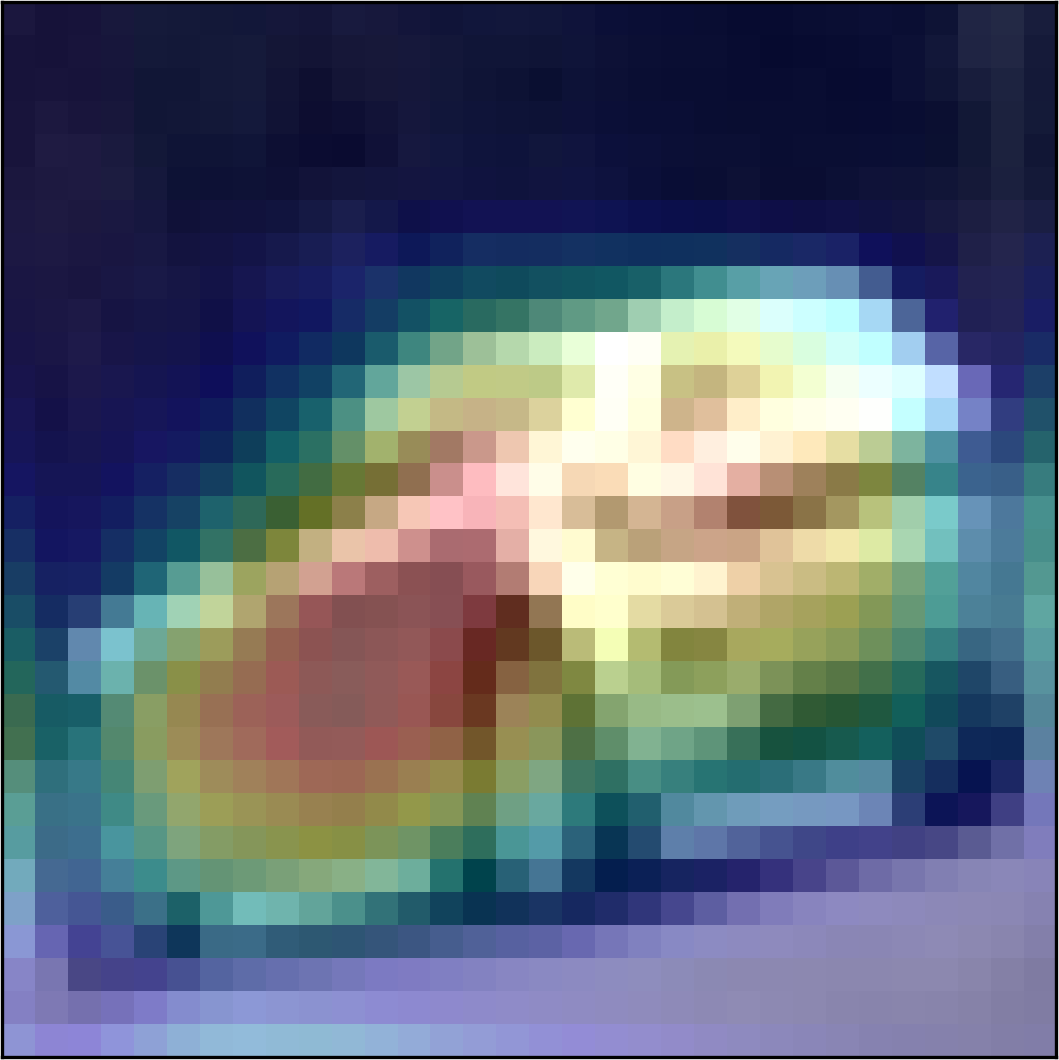} 
    \includegraphics[width=0.2\columnwidth]{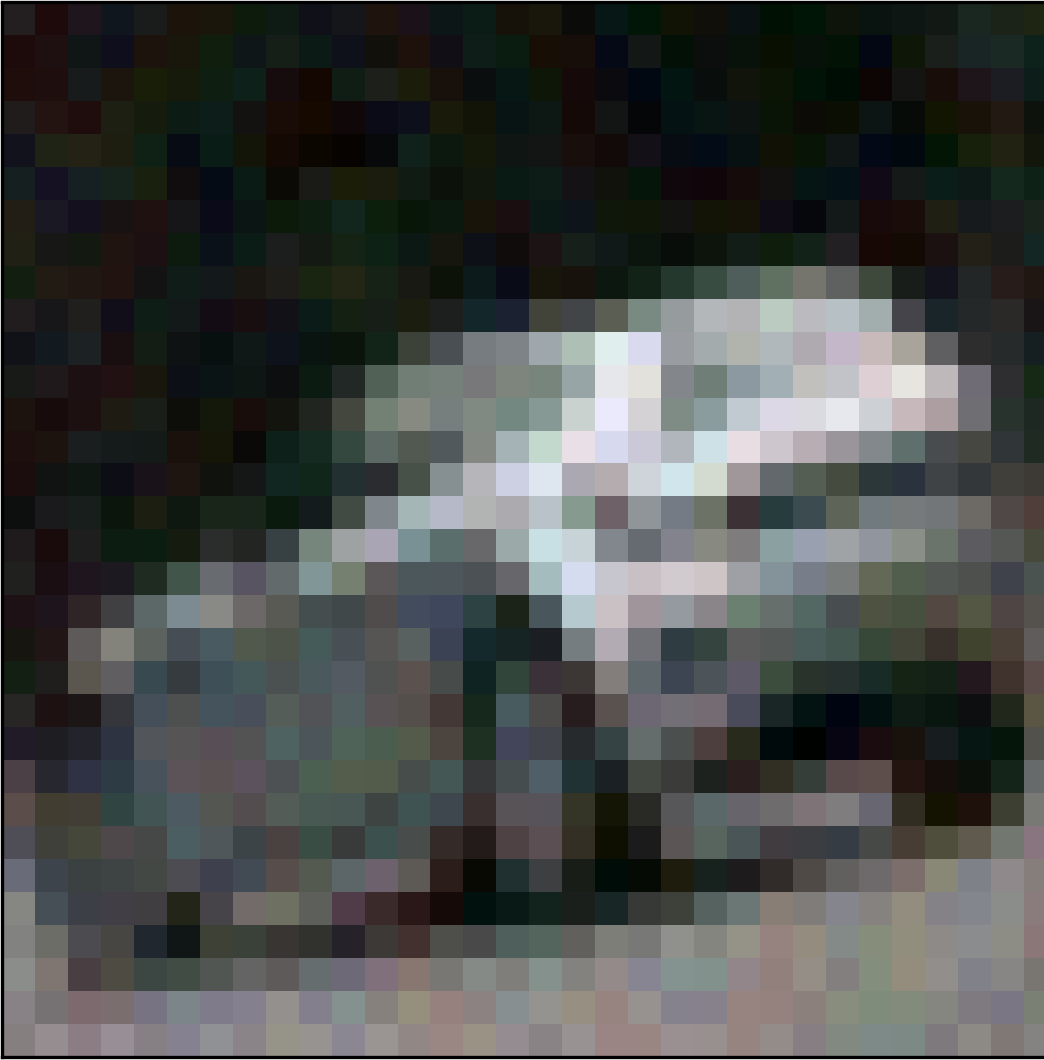} 
    \includegraphics[width=0.2\columnwidth]{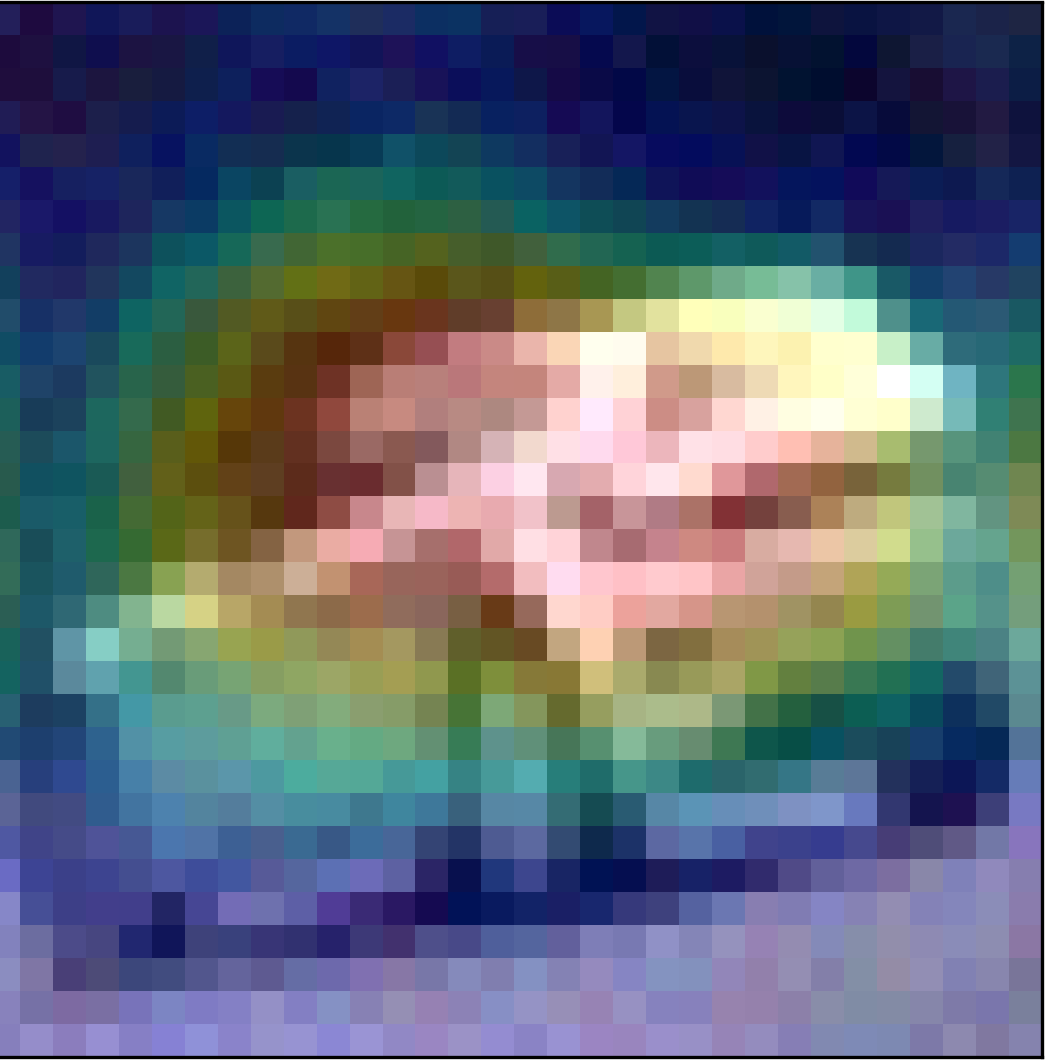} \\
    \caption{Overlayed Gradient-weighted Class Activation Maps with respect to predicted class.}
    \label{cam_pred}
\end{figure}

\begin{figure}[!htb]
    \centering
    \textbf{Pixel Attack}\\
    \vspace{0.1cm}
    \hrule
    \vspace{0.1cm}\small{\hspace{0.5cm} $x$ \hspace{1.2cm} $AM_{C}$ \hspace{1.cm} $\hat{x}$\hspace{1.3cm} $\widehat{AM}_{C}$}\\
    \includegraphics[width=0.2\columnwidth]{images/cropped_PixelAttack-test_72_cam_OI.png} 
    \includegraphics[width=0.2\columnwidth]{images/cropped_PixelAttack-test_72_cam_OST.png} 
    \includegraphics[width=0.2\columnwidth]{images/cropped_PixelAttack-test_72_cam_AI.png} 
    \includegraphics[width=0.2\columnwidth]{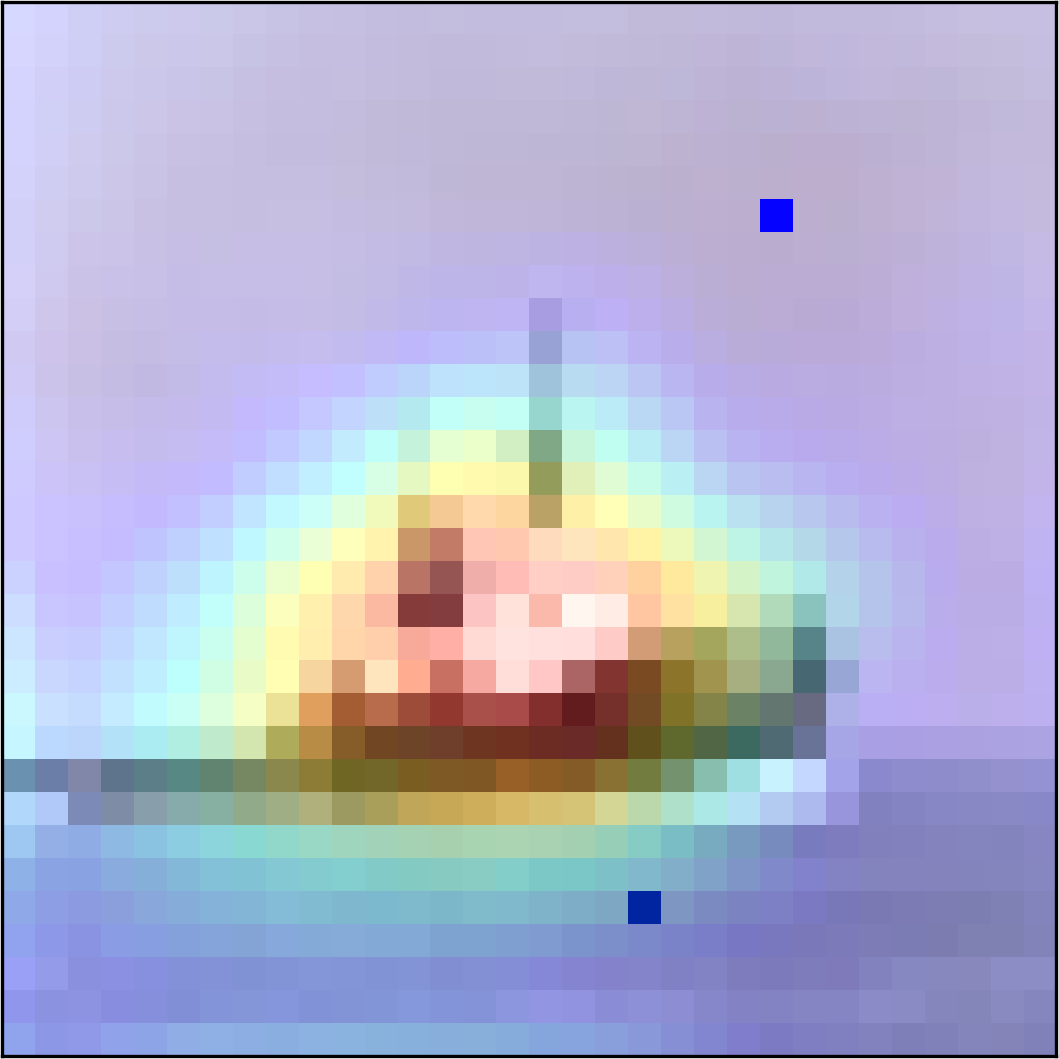} \\
    \vspace{0.1cm}\small{\hspace{0.5cm} $x$ \hspace{1.2cm} $AM_{C}$ \hspace{1.cm} $\hat{x}$\hspace{1.3cm} $\widehat{AM}_{C}$}\\
    \includegraphics[width=0.2\columnwidth]{images/cropped_PixelAttack-test_87_cam_OI.png} 
    \includegraphics[width=0.2\columnwidth]{images/cropped_PixelAttack-test_87_cam_OSM.png} 
    \includegraphics[width=0.2\columnwidth]{images/cropped_PixelAttack-test_87_cam_AI.png} 
    \includegraphics[width=0.2\columnwidth]{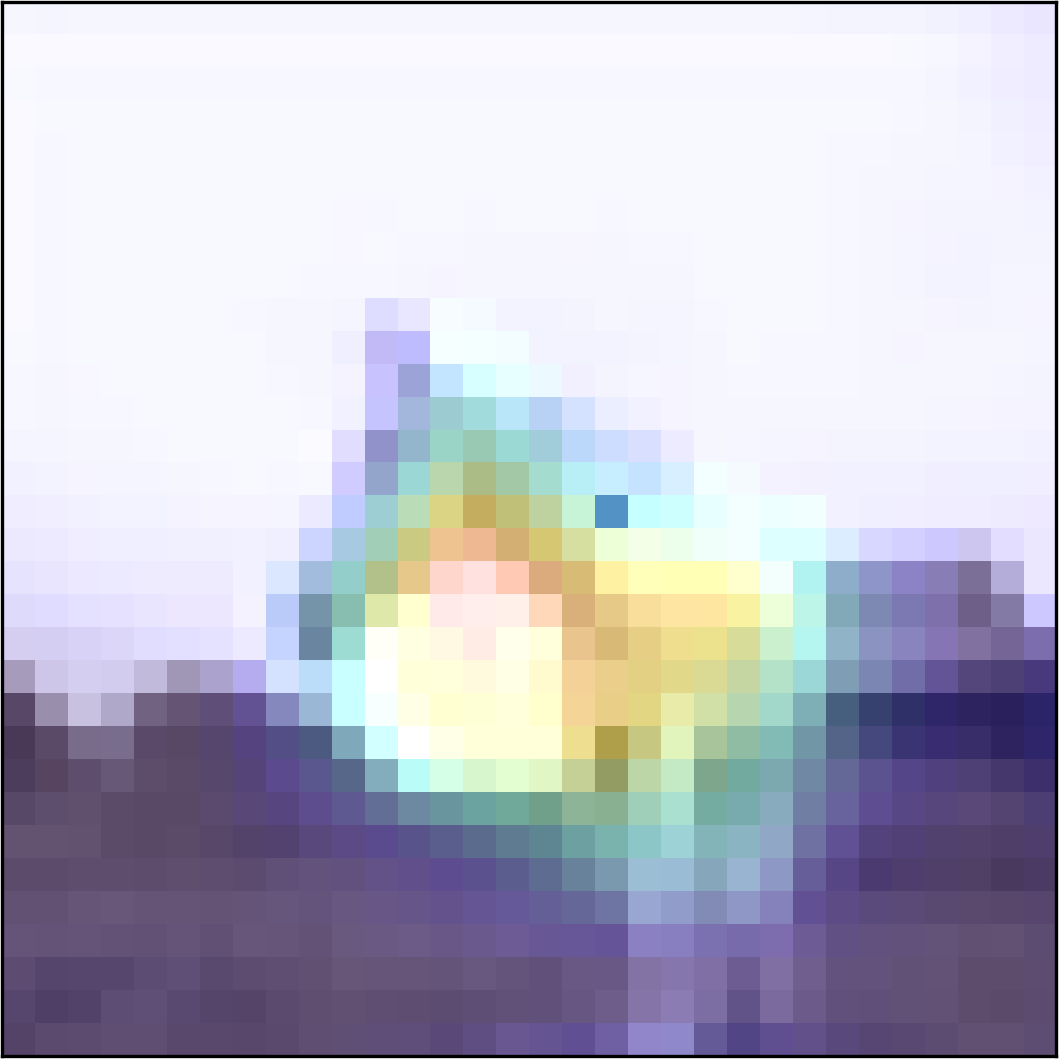} \\
    \vspace{0.1cm}\small{\hspace{0.5cm} $x$ \hspace{1.2cm} $AM_{C}$ \hspace{1.cm} $\hat{x}$\hspace{1.3cm} $\widehat{AM}_{C}$}\\
    \includegraphics[width=0.2\columnwidth]{images/cropped_PixelAttack-test_46_cam_OI.png} 
    \includegraphics[width=0.2\columnwidth]{images/cropped_PixelAttack-test_46_cam_OSM.png} 
    \includegraphics[width=0.2\columnwidth]{images/cropped_PixelAttack-test_46_cam_AI.png} 
    \includegraphics[width=0.2\columnwidth]{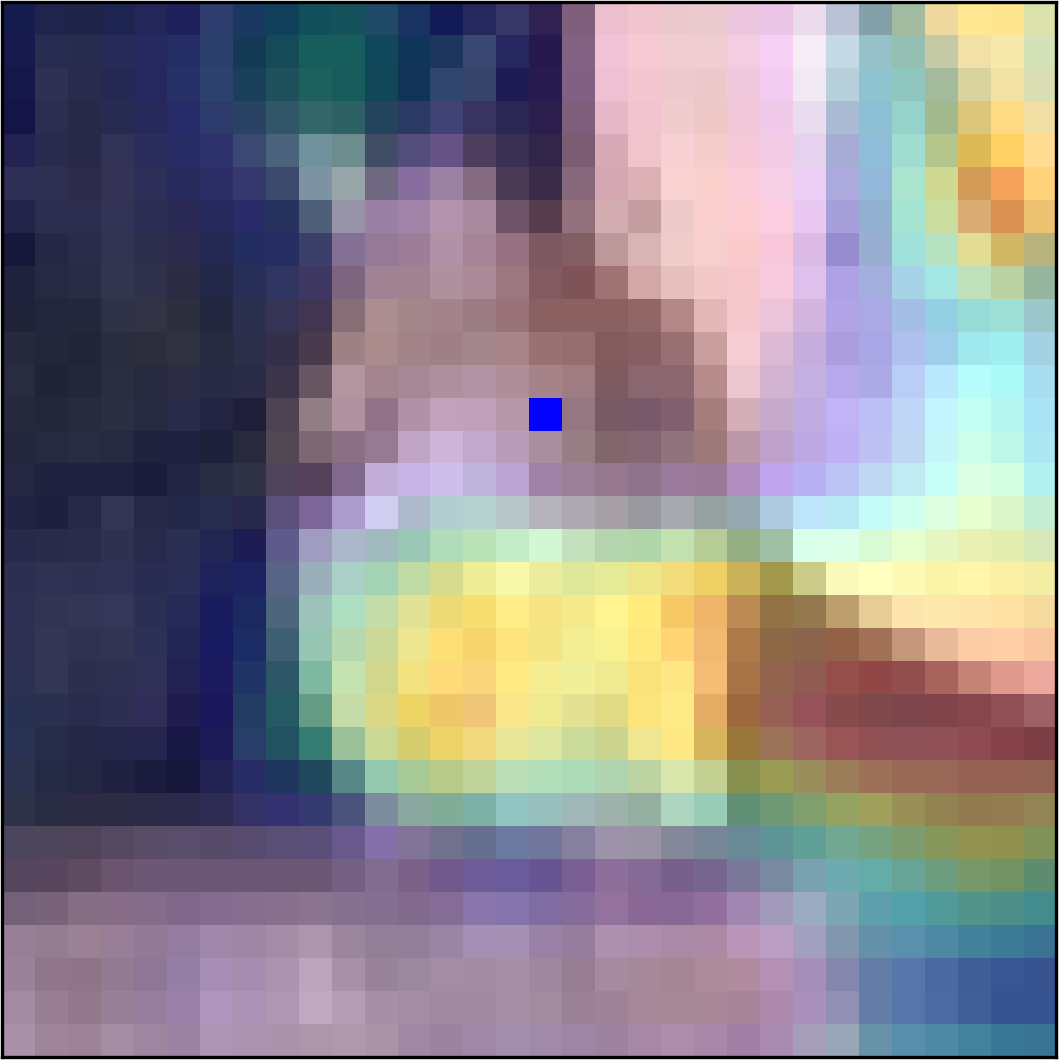} \\
    \vspace{0.1cm}
    \textbf{Projected Gradient Descent Attack}\\
    \vspace{0.1cm}
    \hrule
    \vspace{0.1cm}\small{\hspace{0.5cm} $x$ \hspace{1.2cm} $AM_{C}$ \hspace{1.cm} $\hat{x}$\hspace{1.3cm} $\widehat{AM}_{C}$}\\
    \includegraphics[width=0.2\columnwidth]{images/cropped_PGD_12_cam_OI.png} 
    \includegraphics[width=0.2\columnwidth]{images/cropped_PGD_12_cam_OST.png} 
    \includegraphics[width=0.2\columnwidth]{images/cropped_PGD_12_cam_AI.png} 
    \includegraphics[width=0.2\columnwidth]{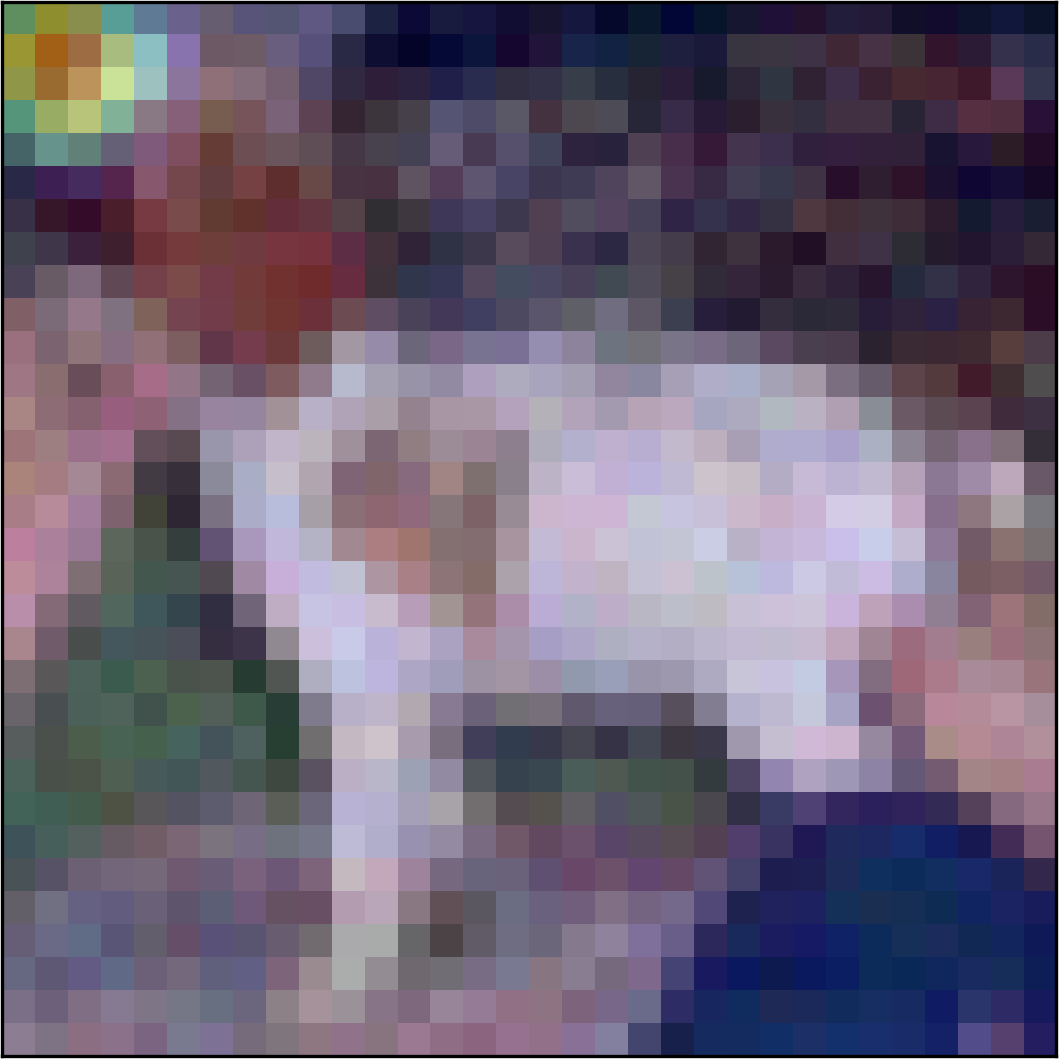} \\
    \vspace{0.1cm}\small{\hspace{0.5cm} $x$ \hspace{1.2cm} $AM_{C}$ \hspace{1.cm} $\hat{x}$\hspace{1.3cm} $\widehat{AM}_{C}$}\\
    \includegraphics[width=0.2\columnwidth]{images/cropped_PGD_161_cam_OI.png} 
    \includegraphics[width=0.2\columnwidth]{images/cropped_PGD_161_cam_OST.png} 
    \includegraphics[width=0.2\columnwidth]{images/cropped_PGD_161_cam_AI.png} 
    \includegraphics[width=0.2\columnwidth]{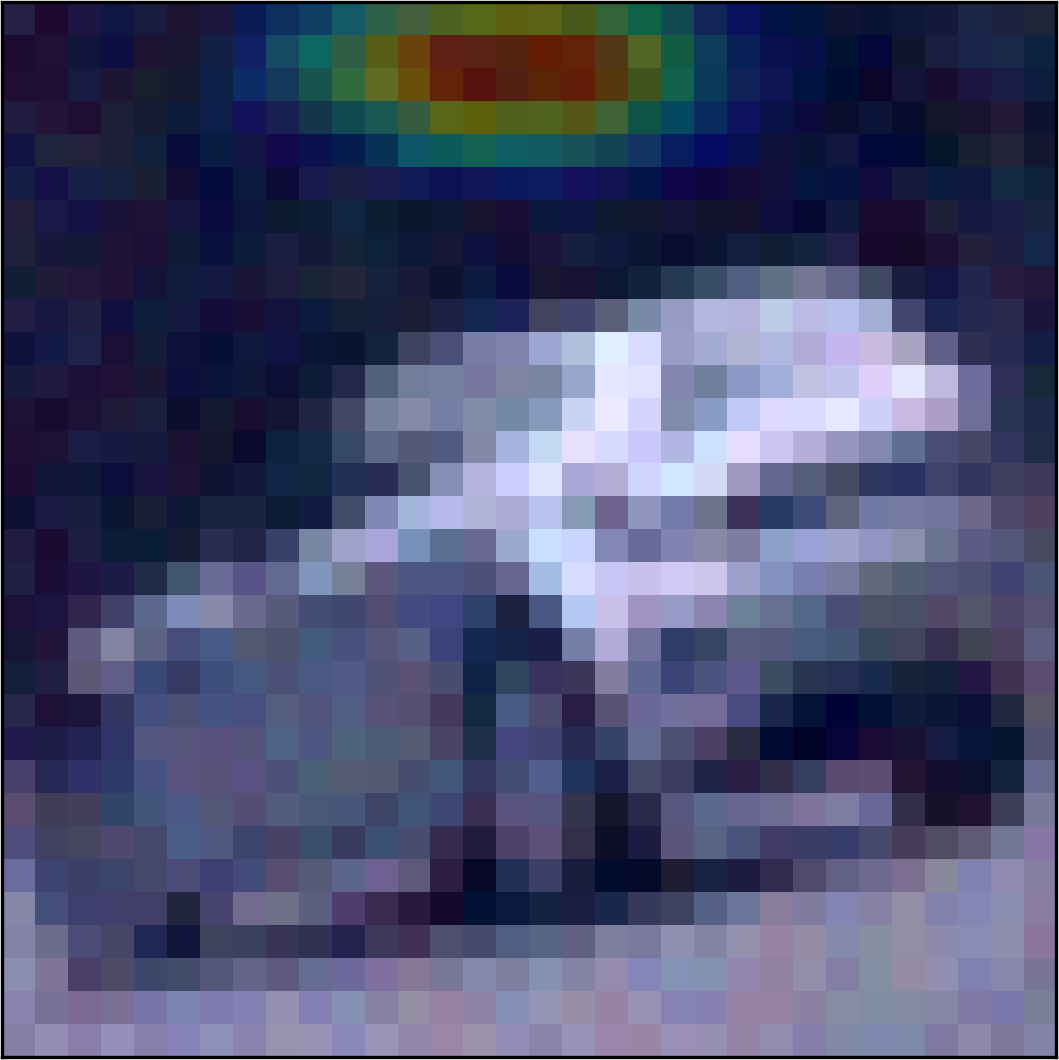} \\
    \caption{Overlayed Gradient-weighted Class Activation Maps with respect to true/correct class.}
    \label{cam_true}
\end{figure}

\begin{figure}[!htb]
    \centering
    \textbf{Pixel Attack}\\
    \vspace{0.1cm}
    \hrule
    \vspace{0.1cm}\small{\hspace{0.5cm} $x$ \hspace{1.2cm} $AM_{\hat{C}}$ \hspace{1.cm} $\hat{x}$\hspace{1.3cm} $\widehat{AM}_{\hat{C}}$}\\
    \includegraphics[width=0.2\columnwidth]{images/cropped_PixelAttack-test_72_cam_OI.png} 
    \includegraphics[width=0.2\columnwidth]{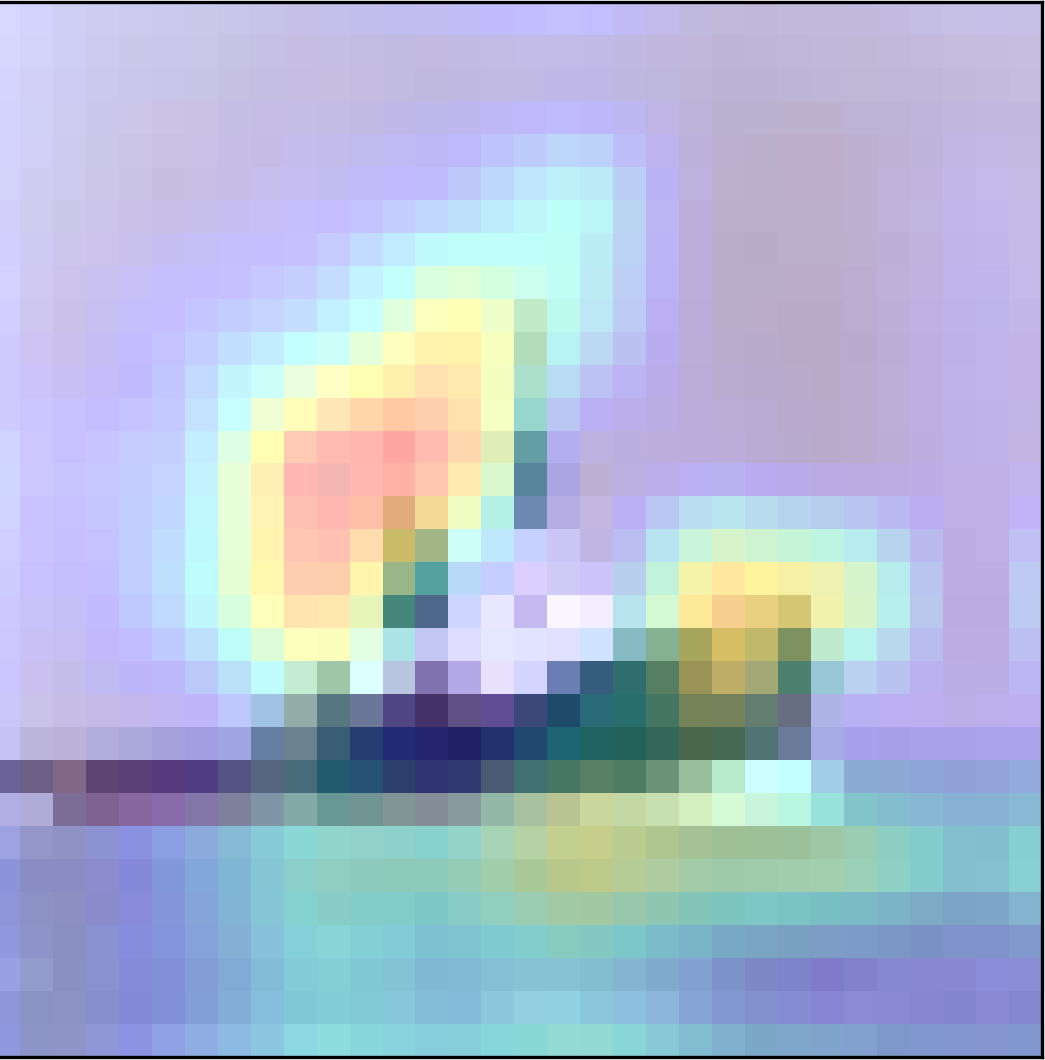} 
    \includegraphics[width=0.2\columnwidth]{images/cropped_PixelAttack-test_72_cam_AI.png} 
    \includegraphics[width=0.2\columnwidth]{images/cropped_PixelAttack-test_72_cam_AIM.png} \\
    \vspace{0.1cm}\small{\hspace{0.5cm} $x$ \hspace{1.2cm} $AM_{\hat{C}}$ \hspace{1.cm} $\hat{x}$\hspace{1.3cm} $\widehat{AM}_{\hat{C}}$}\\
    \includegraphics[width=0.2\columnwidth]{images/cropped_PixelAttack-test_87_cam_OI.png} 
    \includegraphics[width=0.2\columnwidth]{images/cropped_PixelAttack-test_87_cam_OSM.png} 
    \includegraphics[width=0.2\columnwidth]{images/cropped_PixelAttack-test_87_cam_AI.png} 
    \includegraphics[width=0.2\columnwidth]{images/cropped_PixelAttack-test_87_cam_AIM.png} \\
    \vspace{0.1cm}\small{\hspace{0.5cm} $x$ \hspace{1.2cm} $AM_{\hat{C}}$ \hspace{1.cm} $\hat{x}$\hspace{1.3cm} $\widehat{AM}_{\hat{C}}$}\\
    \includegraphics[width=0.2\columnwidth]{images/cropped_PixelAttack-test_46_cam_OI.png} 
    \includegraphics[width=0.2\columnwidth]{images/cropped_PixelAttack-test_46_cam_OSM.png} 
    \includegraphics[width=0.2\columnwidth]{images/cropped_PixelAttack-test_46_cam_AI.png} 
    \includegraphics[width=0.2\columnwidth]{images/cropped_PixelAttack-test_46_cam_AIM.png} \\
    \vspace{0.1cm}
    \textbf{Projected Gradient Descent Attack}\\
    \vspace{0.1cm}
    \hrule
    \vspace{0.1cm}\small{\hspace{0.5cm} $x$ \hspace{1.2cm} $AM_{\hat{C}}$ \hspace{1.cm} $\hat{x}$\hspace{1.3cm} $\widehat{AM}_{\hat{C}}$}\\
    \includegraphics[width=0.2\columnwidth]{images/cropped_PGD_12_cam_OI.png} 
    \includegraphics[width=0.2\columnwidth]{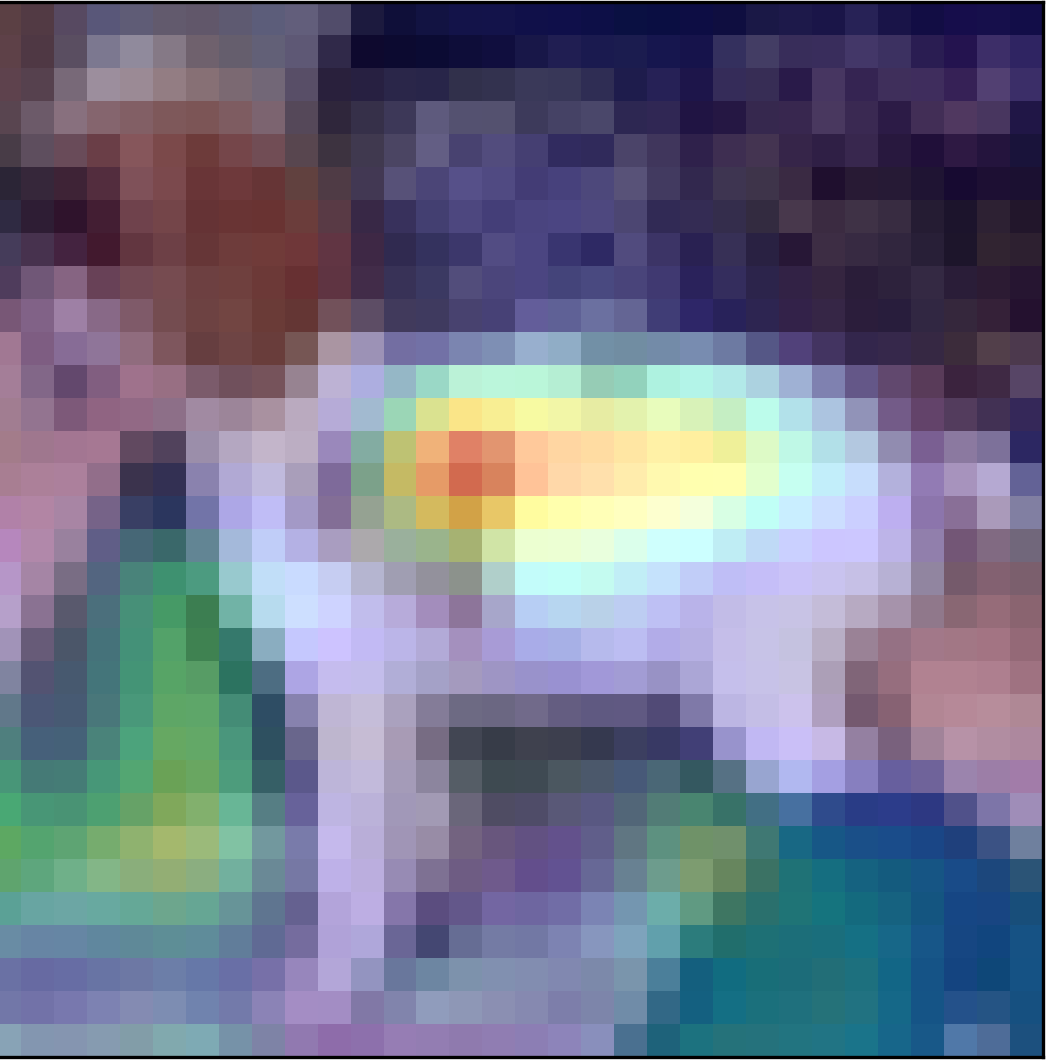} 
    \includegraphics[width=0.2\columnwidth]{images/cropped_PGD_12_cam_AI.png} 
    \includegraphics[width=0.2\columnwidth]{images/cropped_PGD_12_cam_AIM.png} \\
    \vspace{0.1cm}\small{\hspace{0.5cm} $x$ \hspace{1.2cm} $AM_{\hat{C}}$ \hspace{1.cm} $\hat{x}$\hspace{1.3cm} $\widehat{AM}_{\hat{C}}$}\\
    \includegraphics[width=0.2\columnwidth]{images/cropped_PGD_161_cam_OI.png} 
    \includegraphics[width=0.2\columnwidth]{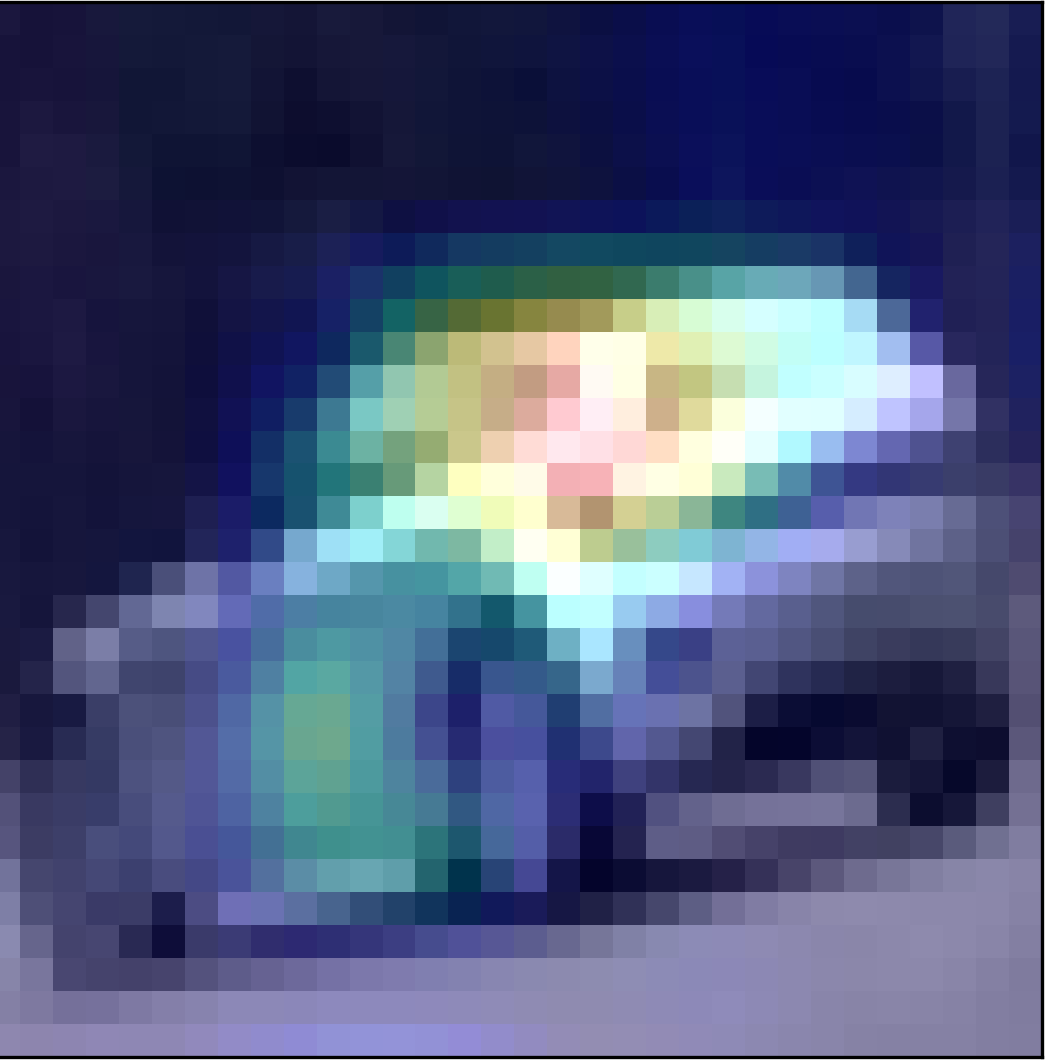} 
    \includegraphics[width=0.2\columnwidth]{images/cropped_PGD_161_cam_AI.png} 
    \includegraphics[width=0.2\columnwidth]{images/cropped_PGD_161_cam_AIM.png} \\

    \caption{Overlayed Gradient-weighted Class Activation Maps with respect to adversarial/incorrect class.}
    \label{cam_false} 
\end{figure}

We visualise the gradient-weighted class activation maps of original sample $(AM)$ and activation maps of adversarial sample $(\widehat{AM})$ with respect to predicted class of the model (Figure \ref{cam_pred}), true/correct class $(C)$ (Figure \ref{cam_true}), and adversarial/incorrect class $(\hat{C})$ (Figure \ref{cam_false}) for different attacks. 
For original sample predicted class is true/correct class $(C)$ where as for adversarial sample predicted class adversarial/incorrect class $(\hat{C})$.

Figure \ref{cam_pred} shows that the Projected Gradient Descent, a white-box attack, keeps the attention of the convolution layer in the same region using the gradients.
However, from Figure \ref{cam_true}, we can see that attention for the correct class is strongly distorted for adversarial image compared to the original.
Whereas the attention for the misclassified class for the adversarial image is distorted to bring it close to the attention for the correct class for the original image. 
This points out that white-box attacks find the perturbation keeping the region of interest intact and thus only effectively manipulating the activations inside the neural network (convolution layers) to call for misclassification.

From the activation maps of the last convolution layer for Pixel Attack (Figure \ref{cam_pred}), we observe that adversarial perturbations distort the attention of the convolution layer. 
This distortion either calls the attention towards or diverts the network's attention from the perturbed pixels. 
Similar to the characteristic shown by the Saliency Maps, Figures \ref{cam_true} and \ref{cam_false} show that this distortion in activation maps is mainly focused on adversarial class's attention, and the true class's attention is changed minimally.

An interesting observation of the activation maps is when the activation map of a particular class retracts from the perturbed pixel, and the activation map of another class gets intensely focused on the perturbed pixel.
This sheds light on the fact that some of the features learned are complementary to each other, and when the network focuses on one feature, it loses the attention on other complementary feature.
Further, the influence of perturbed pixels can be visualised through propagation maps \cite{vargas2019understanding}. 
As the perturbed pixels call attention towards another part of the image, another texture is analysed and weighted more.

We believe each adversarial attack has its inherent characteristic strategy to distort the saliency maps and activation maps of adversarial samples, as both Projected Gradient Descent Attack and Pixel Attack changes the saliency maps and activation maps differently. 
This sheds light on why some defences are successful against some attacks while remaining vulnerable to other attacks.
This also calls for a deeper understanding of the effect of adversarial samples and attacks in general on neural networks to propose effective defences by understanding how each attack affects the attention of the network. 

\section{Conclusions}

This paper analysed Saliency Maps (SM) and Gradient-weighted Class Activation Maps (Grad-CAM) for original images and adversarial images.
We analyse the distortion in the saliency of adversarial images compared to original images to verify the hypothesis of conflicting saliency.
We used adversarial images created by Pixel Attack and Projected Gradient Descent Attack for analysing the saliency.
Both the attacks differed the way adversarial perturbations are found (Pixel Attack being black-box and PGD being white box) and how adversarial perturbations are added to the image (Pixel Attack being $L_0$ norm attack while PGD being $L_\infty$ attack). 

Experimental results show that both black-box and white-box attacks, irrespective of the norm constraint, distorts the neural networks' saliency to make adversarial images misclassify, and adversarial attacks do not naively fool the neural networks.
Moreover, results reveal that both Pixel Attack and Projected Gradient Descent Attack distorts the saliency maps and activation maps differently. 
While Pixel Attack, a black-box attack call the image's saliency to perturb pixels or divert their saliency from them, effectively changing the region of interest for intermediate convolution layers of the neural network to change their region of attention.
The Projected Gradient Descent Attack, a white box attack, on the other hand, diffuses the saliency around the region of interest to induce misclassification.
Further, this diffusion of saliency causes intermediate convolution layers to lose attention for the correct class around the region of interest while gaining attention for the misclassified class on the intended region.

Thus, this paper analysed saliency for the adversarial images and shed light on the conflicting saliency hypothesis raised in \cite{vargas2019understanding}. 
It also opens up the understanding of adversarial attacks by analysing the effect of adversarial samples on neural networks. 
As we show, both the attacks evaluated differ in their strategy to distort the attention of the neural network. 
We believe this sheds light on why some adversarial defences mitigate some adversarial attacks while remaining vulnerable to others.
We hope this analysis will help the community understand the existence of adversarial samples and their effect on neural networks and help the community to develop more robust neural networks and, at the same time, develop better adversarial defences.

\section*{Acknowledgments}

This work was supported by JST, ACT-I Grant Number JP-50243 and JSPS KAKENHI Grant Number JP20241216.

\bibliographystyle{named}
\bibliography{adversarial_machine_learning}

\begin{thebibliography}{}

\bibitem[\protect\citeauthoryear{Athalye \bgroup \em et al.\egroup
  }{2018}]{athalye2018obfuscated}
Anish Athalye, Nicholas Carlini, and David Wagner.
\newblock Obfuscated gradients give a false sense of security: Circumventing
  defenses to adversarial examples.
\newblock In {\em Icml}, 2018.

\bibitem[\protect\citeauthoryear{Brown \bgroup \em et al.\egroup
  }{2017}]{brown2017adversarial}
Tom~B Brown, Dandelion Man{\'e}, Aurko Roy, Mart{\'\i}n Abadi, and Justin
  Gilmer.
\newblock Adversarial patch.
\newblock {\em arXiv preprint arXiv:1712.09665}, 2017.

\bibitem[\protect\citeauthoryear{Carlini and Wagner}{2017}]{carlini2017towards}
Nicholas Carlini and David Wagner.
\newblock Towards evaluating the robustness of neural networks.
\newblock In {\em 2017 ieee symposium on security and privacy (sp)}, pages
  39--57. Ieee, 2017.

\bibitem[\protect\citeauthoryear{Fawzi \bgroup \em et al.\egroup
  }{2018}]{fawzi2018empirical}
Alhussein Fawzi, Seyed-Mohsen Moosavi-Dezfooli, Pascal Frossard, and Stefano
  Soatto.
\newblock Empirical study of the topology and geometry of deep networks.
\newblock In {\em Proceedings of the IEEE Conference on Computer Vision and
  Pattern Recognition}, pages 3762--3770, 2018.

\bibitem[\protect\citeauthoryear{Goodfellow \bgroup \em et al.\egroup
  }{2014}]{goodfellow2014explaining}
Ian~J Goodfellow, Jonathon Shlens, and Christian Szegedy.
\newblock Explaining and harnessing adversarial examples.
\newblock {\em arXiv preprint arXiv:1412.6572}, 2014.

\bibitem[\protect\citeauthoryear{He \bgroup \em et al.\egroup
  }{2016}]{he2016deep}
Kaiming He, Xiangyu Zhang, Shaoqing Ren, and Jian Sun.
\newblock Deep residual learning for image recognition.
\newblock In {\em Proceedings of the IEEE Conference on Computer Vision and
  Pattern Recognition}, pages 770--778, 2016.

\bibitem[\protect\citeauthoryear{Itti \bgroup \em et al.\egroup
  }{1998}]{itti1998model}
Laurent Itti, Christof Koch, and Ernst Niebur.
\newblock A model of saliency-based visual attention for rapid scene analysis.
\newblock {\em IEEE Transactions on Pattern Analysis \& Machine Intelligence},
  (11):1254--1259, 1998.

\bibitem[\protect\citeauthoryear{Krizhevsky \bgroup \em et al.\egroup
  }{2009}]{krizhevsky2009learning}
Alex Krizhevsky, Geoffrey Hinton, et~al.
\newblock Learning multiple layers of features from tiny images.
\newblock Technical report, 2009.

\bibitem[\protect\citeauthoryear{Madry \bgroup \em et al.\egroup
  }{2018}]{madry2018towards}
Aleksander Madry, Aleksandar Makelov, Ludwig Schmidt, Dimitris Tsipras, and
  Adrian Vladu.
\newblock Towards deep learning models resistant to adversarial attacks.
\newblock In {\em International Conference on Learning Representations}, 2018.

\bibitem[\protect\citeauthoryear{Moosavi-Dezfooli \bgroup \em et al.\egroup
  }{2017}]{moosavi2017universal}
Seyed-Mohsen Moosavi-Dezfooli, Alhussein Fawzi, Omar Fawzi, and Pascal
  Frossard.
\newblock Universal adversarial perturbations.
\newblock In {\em Proceedings of the IEEE conference on computer vision and
  pattern recognition}, pages 1765--1773. Ieee, 2017.

\bibitem[\protect\citeauthoryear{Moosavi-Dezfooli \bgroup \em et al.\egroup
  }{2018}]{moosavi2018robustness}
Seyed-Mohsen Moosavi-Dezfooli, Alhussein Fawzi, Omar Fawzi, Pascal Frossard,
  and Stefano Soatto.
\newblock Robustness of classifiers to universal perturbations: A geometric
  perspective.
\newblock In {\em International Conference on Learning Representations}, 2018.

\bibitem[\protect\citeauthoryear{Nguyen \bgroup \em et al.\egroup
  }{2015}]{nguyen2015deep}
Anh Nguyen, Jason Yosinski, and Jeff Clune.
\newblock Deep neural networks are easily fooled: High confidence predictions
  for unrecognizable images.
\newblock In {\em Proceedings of the IEEE Conference on Computer Vision and
  Pattern Recognition}, pages 427--436, 2015.

\bibitem[\protect\citeauthoryear{Nicolae \bgroup \em et al.\egroup
  }{2018}]{art2018}
Maria-Irina Nicolae, Mathieu Sinn, Minh~Ngoc Tran, Beat Buesser, Ambrish Rawat,
  Martin Wistuba, Valentina Zantedeschi, Nathalie Baracaldo, Bryant Chen, Heiko
  Ludwig, Ian Molloy, and Ben Edwards.
\newblock Adversarial robustness toolbox v1.1.0.
\newblock {\em CoRR}, 1807.01069, 2018.

\bibitem[\protect\citeauthoryear{Papernot \bgroup \em et al.\egroup
  }{2016}]{papernot2016distillation}
Nicolas Papernot, Patrick McDaniel, Xi~Wu, Somesh Jha, and Ananthram Swami.
\newblock Distillation as a defense to adversarial perturbations against deep
  neural networks.
\newblock In {\em 2016 IEEE Symposium on Security and Privacy (SP)}, pages
  582--597. Ieee, 2016.

\bibitem[\protect\citeauthoryear{Selvaraju \bgroup \em et al.\egroup
  }{2017}]{selvaraju2017grad}
Ramprasaath~R Selvaraju, Michael Cogswell, Abhishek Das, Ramakrishna Vedantam,
  Devi Parikh, and Dhruv Batra.
\newblock Grad-cam: Visual explanations from deep networks via gradient-based
  localization.
\newblock In {\em Proceedings of the IEEE International Conference on Computer
  Vision}, pages 618--626, 2017.

\bibitem[\protect\citeauthoryear{Simonyan \bgroup \em et al.\egroup
  }{2013}]{simonyan2013deep}
Karen Simonyan, Andrea Vedaldi, and Andrew Zisserman.
\newblock Deep inside convolutional networks: Visualising image classification
  models and saliency maps.
\newblock {\em arXiv preprint arXiv:1312.6034}, 2013.

\bibitem[\protect\citeauthoryear{Springenberg \bgroup \em et al.\egroup
  }{2014}]{springenberg2014striving}
Jost~Tobias Springenberg, Alexey Dosovitskiy, Thomas Brox, and Martin
  Riedmiller.
\newblock Striving for simplicity: The all convolutional net.
\newblock {\em arXiv preprint arXiv:1412.6806}, 2014.

\bibitem[\protect\citeauthoryear{Su \bgroup \em et al.\egroup
  }{2019}]{su2019one}
Jiawei Su, Danilo~Vasconcellos Vargas, and Kouichi Sakurai.
\newblock One pixel attack for fooling deep neural networks.
\newblock {\em IEEE Transactions on Evolutionary Computation}, 23(5):828--841,
  2019.

\bibitem[\protect\citeauthoryear{Szegedy}{2014}]{szegedy2014intriguing}
Christian et~al. Szegedy.
\newblock Intriguing properties of neural networks.
\newblock In {\em In ICLR}. Citeseer, 2014.

\bibitem[\protect\citeauthoryear{Tram{\`e}r \bgroup \em et al.\egroup
  }{2018}]{tramer2018ensemble}
Florian Tram{\`e}r, Alexey Kurakin, Nicolas Papernot, Ian Goodfellow, Dan
  Boneh, and Patrick McDaniel.
\newblock Ensemble adversarial training: Attacks and defenses.
\newblock In {\em International Conference on Learning Representations}, 2018.

\bibitem[\protect\citeauthoryear{Vargas and Su}{2020}]{vargas2019understanding}
Danilo~Vasconcellos Vargas and Jiawei Su.
\newblock Understanding the one-pixel attack: Propagation maps and locality
  analysis.
\newblock In {\em Workshop on Artificial Intelligence Safety (AISafety 2020)},
  2020.

\bibitem[\protect\citeauthoryear{Zeiler and
  Fergus}{2014}]{zeiler2014visualizing}
Matthew~D Zeiler and Rob Fergus.
\newblock Visualizing and understanding convolutional networks.
\newblock In {\em European conference on computer vision}, pages 818--833.
  Springer, 2014.

\end{thebibliography}

\clearpage

\appendix

\onecolumn
\section{Extra Images of Saliency Maps For Pixel Attack}
\begin{figure*}[!b]
	\centering
		\includegraphics[width=0.48\columnwidth]{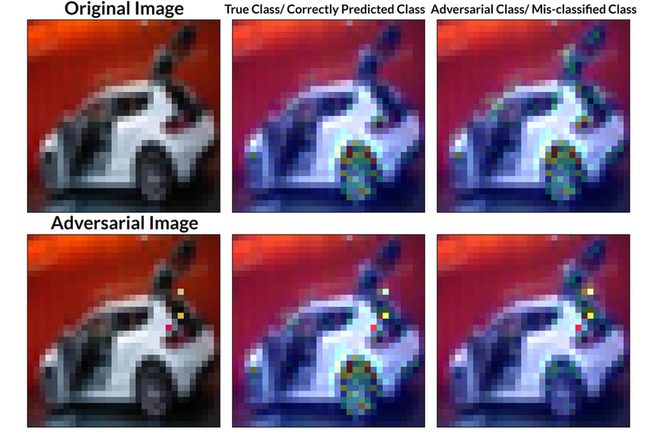} 
	    \includegraphics[width=0.48\columnwidth]{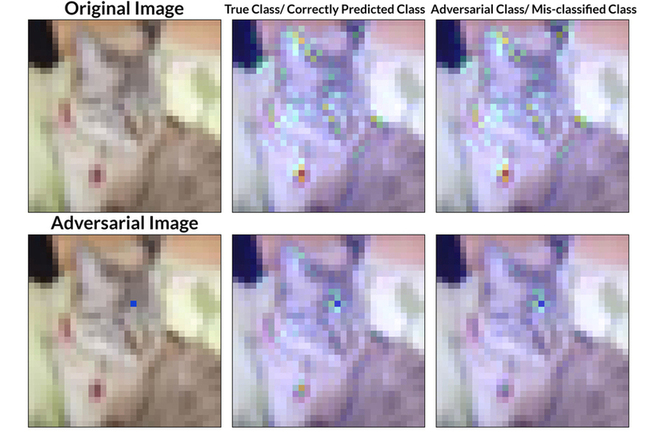}  \\
	    \includegraphics[width=0.48\columnwidth]{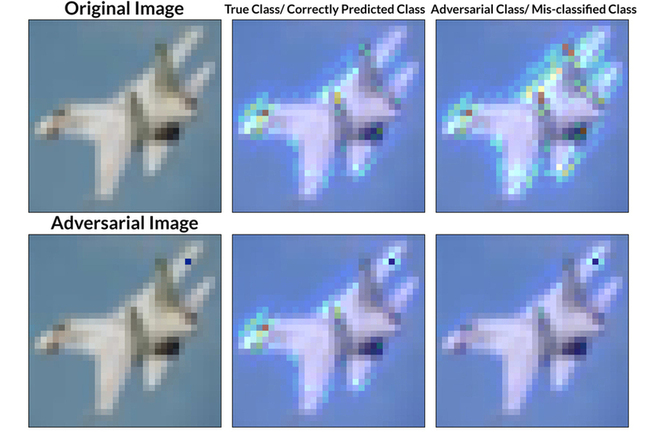}     
		\includegraphics[width=0.48\columnwidth]{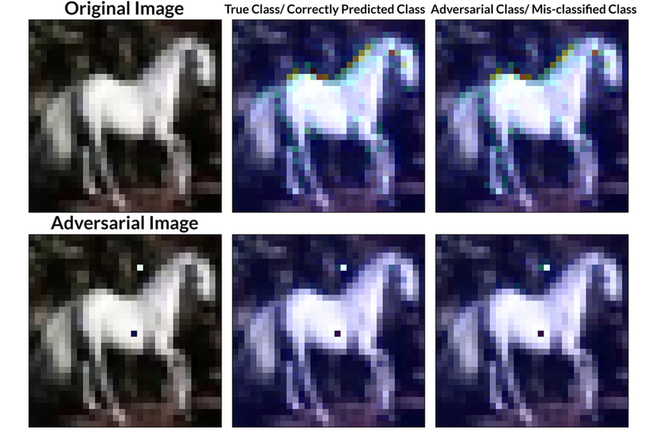} \\
	    \includegraphics[width=0.48\columnwidth]{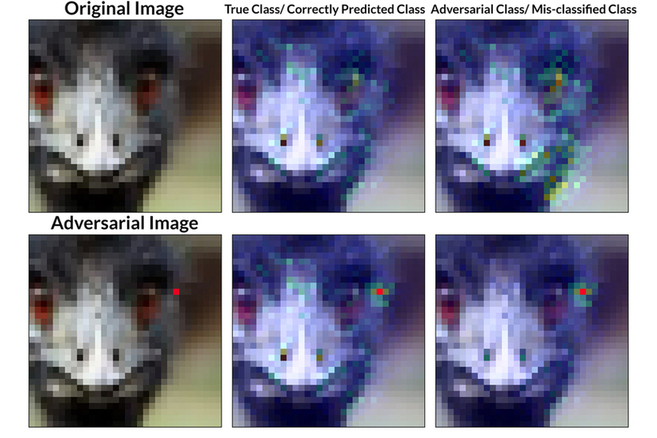}
	    \includegraphics[width=0.48\columnwidth]{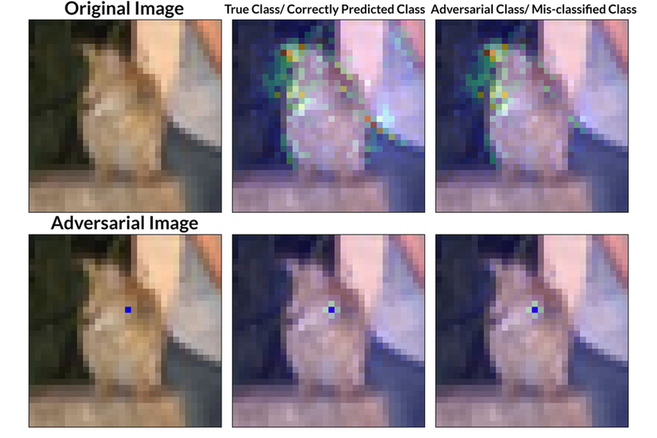}
	\caption{
	Overlayed Saliency Maps with respect to correctly predicted class and the misclassified class of Original Image and Adversarial Image generated by Pixel Attack. 
	}
	\label{saliency_1}
	\end{figure*}

\begin{figure*}[!b]
	\centering
		\includegraphics[width=0.48\columnwidth]{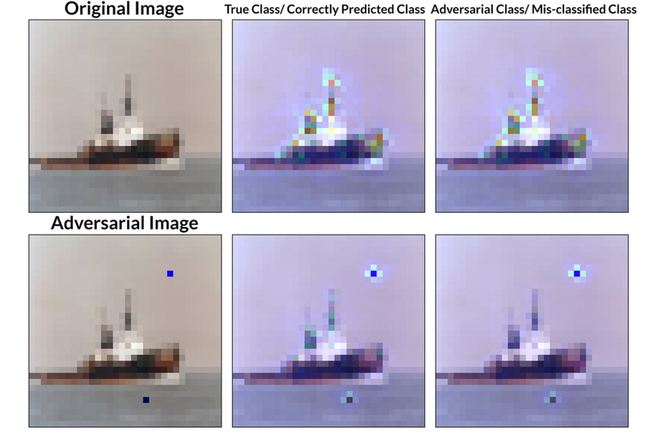} 
	    \includegraphics[width=0.48\columnwidth]{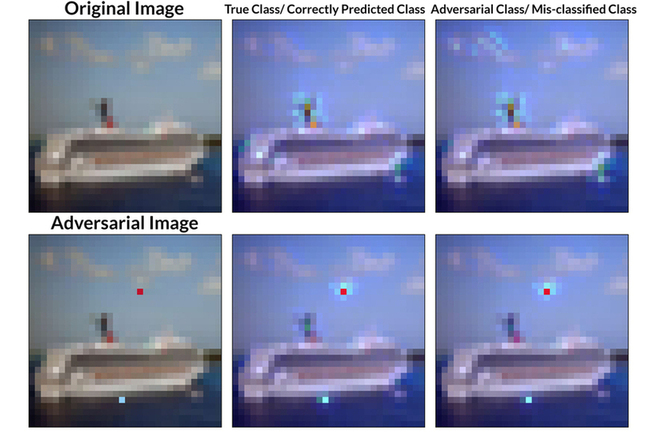} \\
	    \includegraphics[width=0.48\columnwidth]{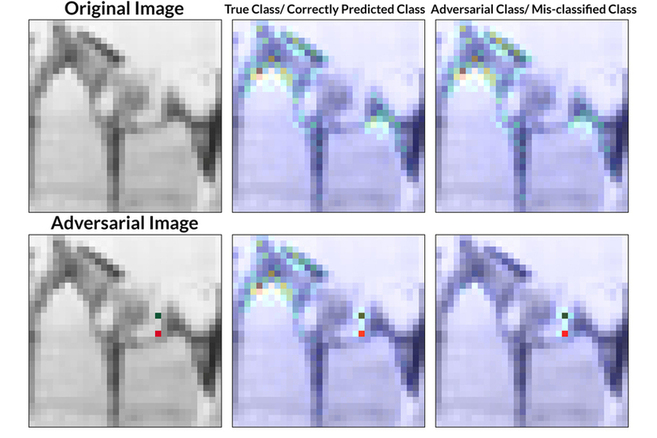} 
	    \includegraphics[width=0.48\columnwidth]{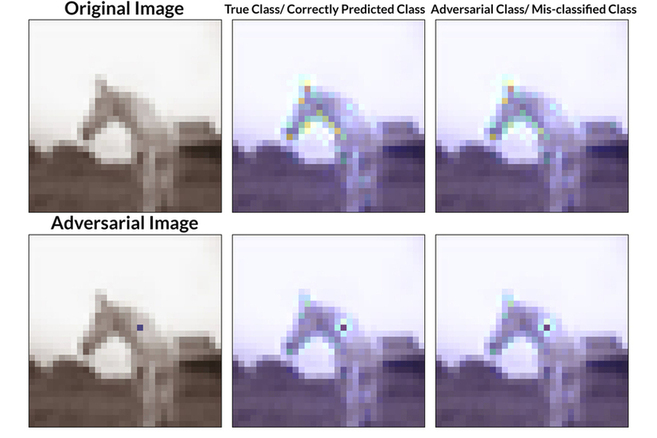} \\   
		\includegraphics[width=0.48\columnwidth]{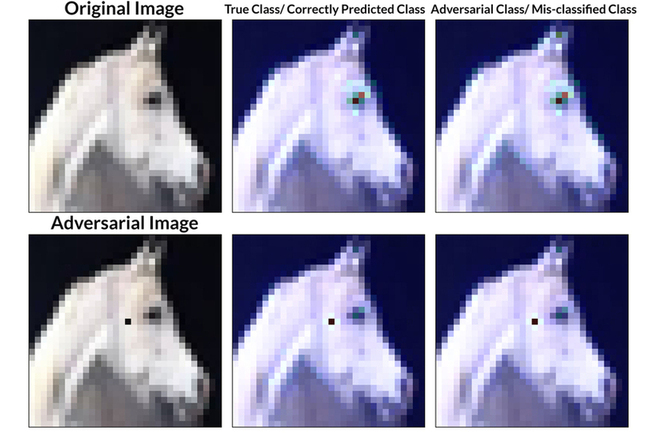} 
	    \includegraphics[width=0.48\columnwidth]{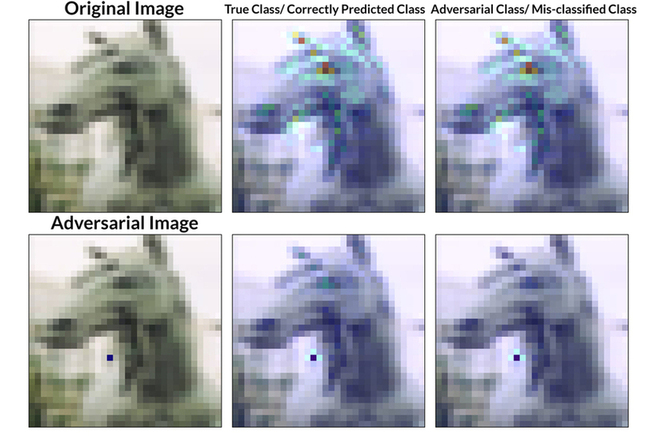} \\
	    \includegraphics[width=0.48\columnwidth]{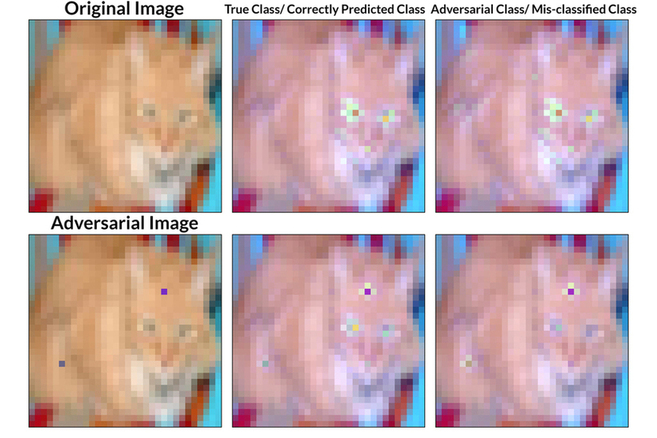}
	    \includegraphics[width=0.48\columnwidth]{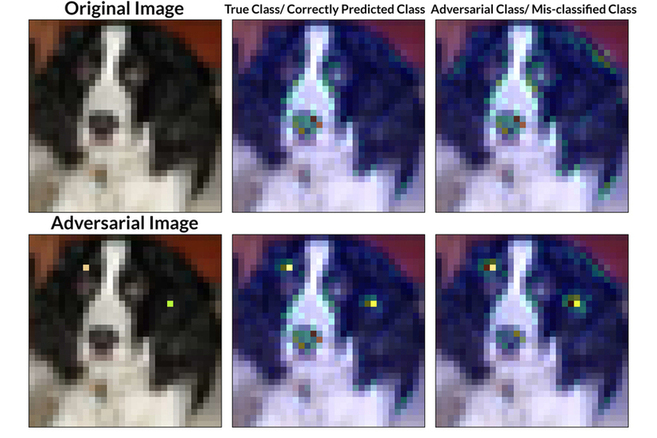}
	\caption{
	Overlayed Saliency Maps with respect to correctly predicted class and the misclassified class of Original Image and Adversarial Image generated by Pixel Attack. 
	}
	\label{saliency_2}
	\end{figure*}

	Figures \ref{saliency_1} and \ref{saliency_2} shows the Saliency Maps of original image and adversarial image with respect to correct class and misclassified class for Pixel Attack.
	First column shows the input image to the model. 
    Middle column shows activation maps concerning correctly predicted class $(C)$. 
    Rightmost column shows activation maps concerning misclassified class $(\hat{C})$.

\clearpage

\section{Extra Images of Gradient-weighted Class Activation Maps For Pixel Attack}
\begin{figure*}[!b]
	\centering
		\includegraphics[width=0.48\columnwidth]{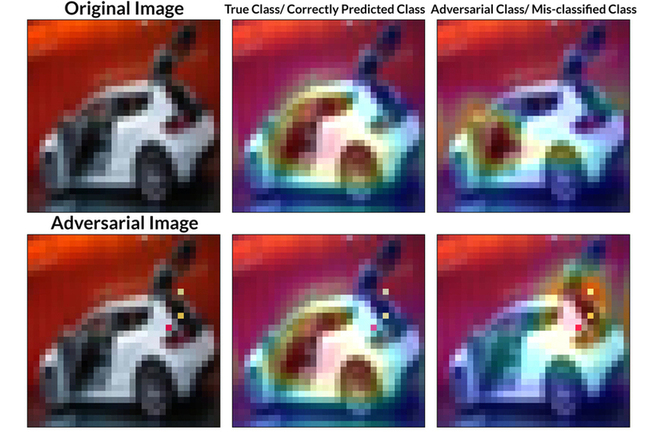} 
	    \includegraphics[width=0.48\columnwidth]{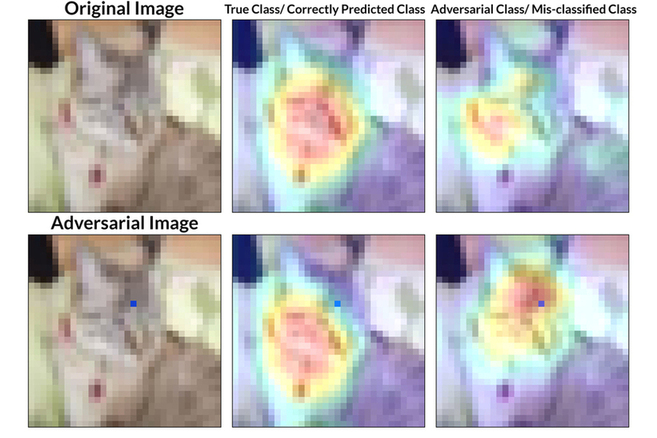}  \\
	    \includegraphics[width=0.48\columnwidth]{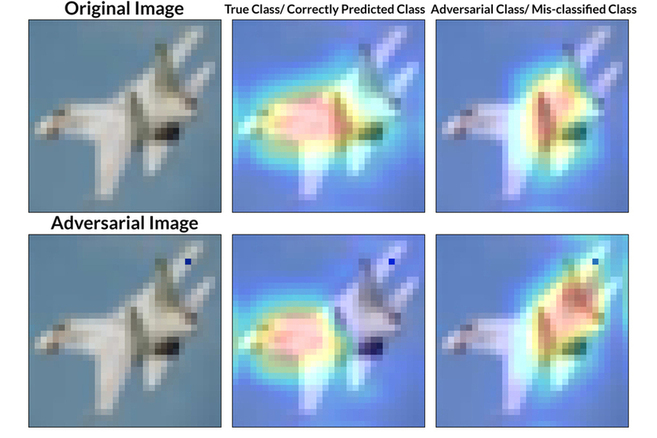}     
		\includegraphics[width=0.48\columnwidth]{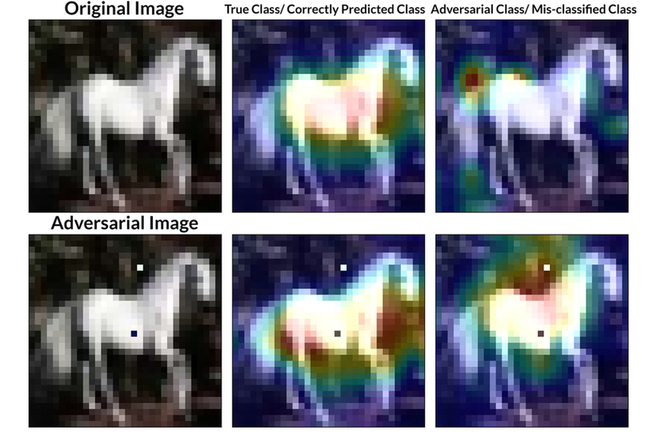} \\
	    \includegraphics[width=0.48\columnwidth]{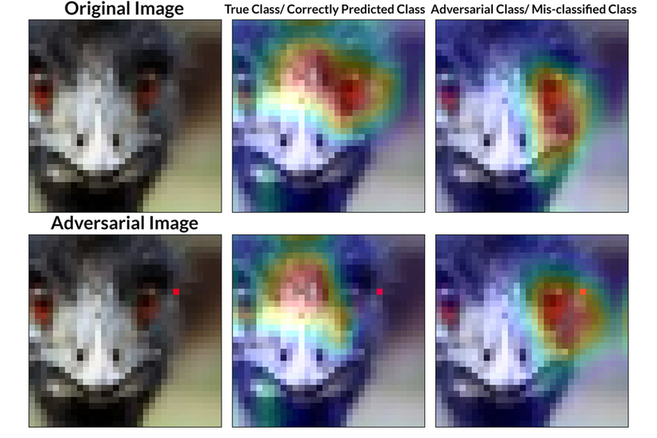}
	    \includegraphics[width=0.48\columnwidth]{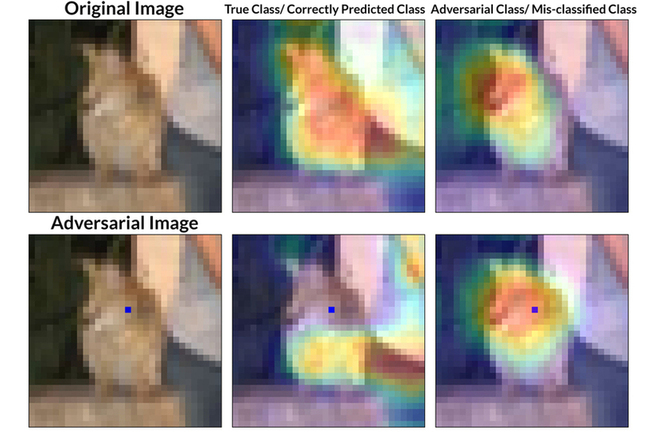}
	\caption{
	Overlayed Gradient-weighted Class Activation Maps with respect to correctly predicted class and the misclassified class of Original Image and Adversarial Image generated by Pixel Attack. 
	}
	\label{cam_1}
	\end{figure*}

\begin{figure*}[!b]
	\centering
		\includegraphics[width=0.48\columnwidth]{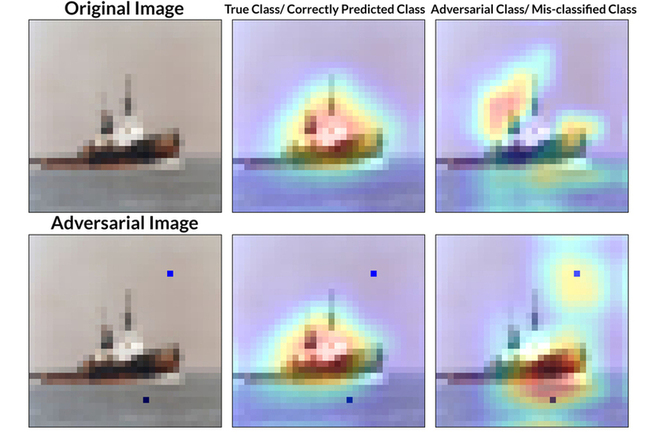} 
	    \includegraphics[width=0.48\columnwidth]{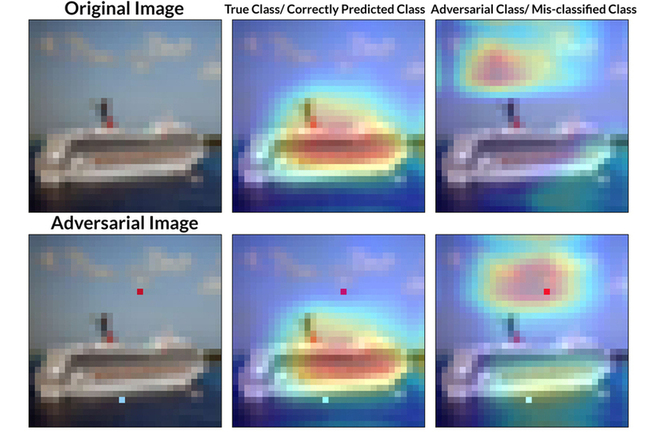} \\
	    \includegraphics[width=0.48\columnwidth]{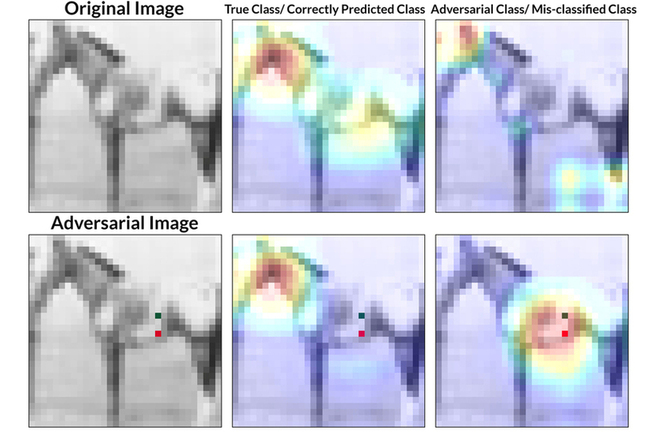} 
	    \includegraphics[width=0.48\columnwidth]{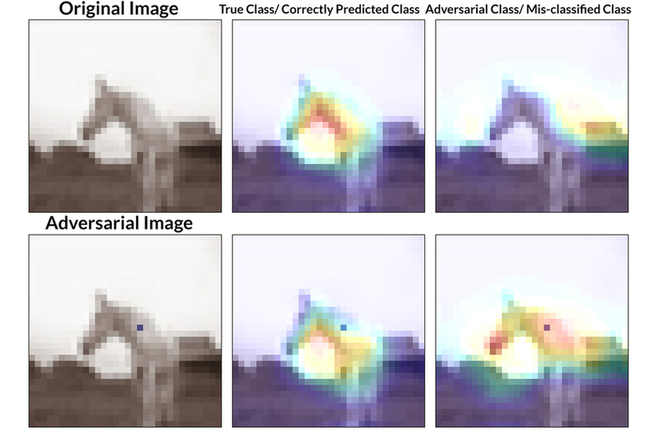} \\   
		\includegraphics[width=0.48\columnwidth]{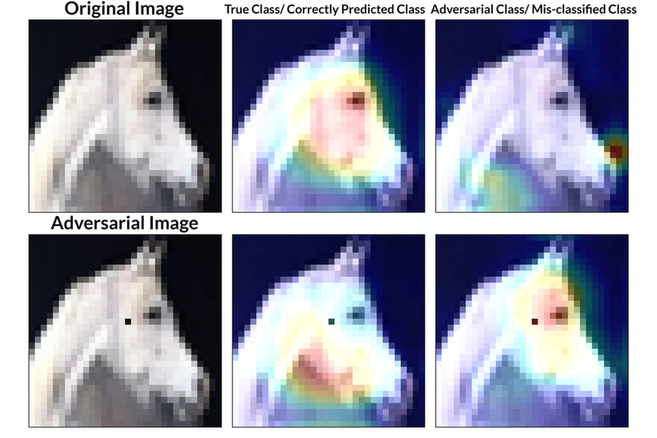} 
	    \includegraphics[width=0.48\columnwidth]{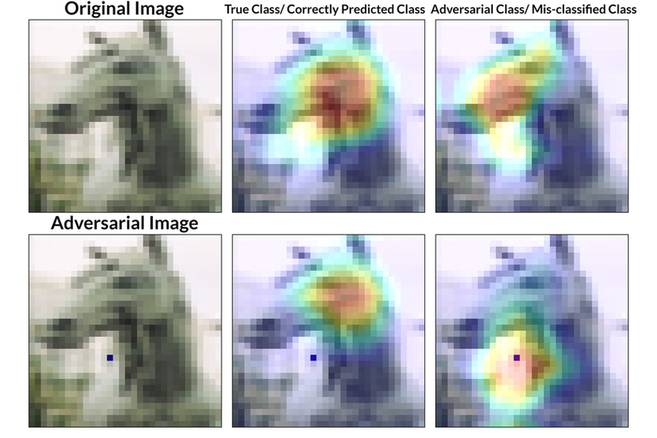} \\
	    \includegraphics[width=0.48\columnwidth]{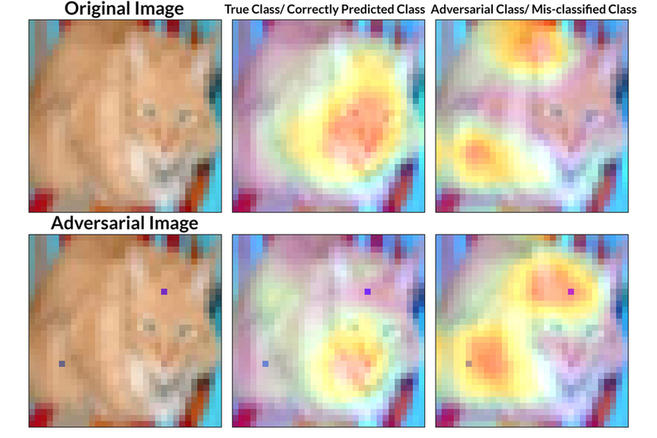}
	    \includegraphics[width=0.48\columnwidth]{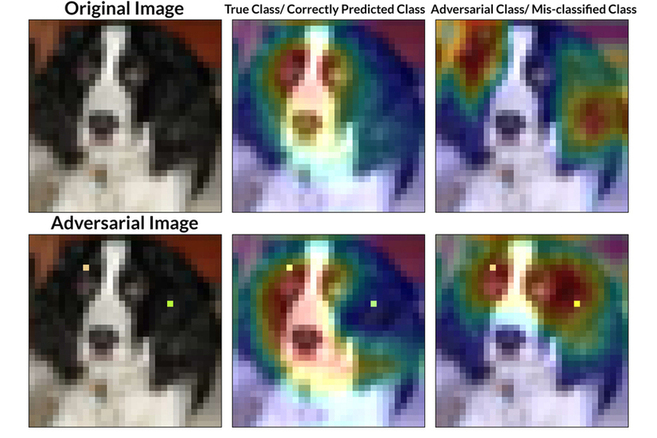}
	\caption{
	Overlayed Gradient-weighted Class Activation Maps with respect to correctly predicted class and misclassified class of Original Image and Adversarial Image generated by Pixel Attack. 
	}
	\label{cam_2}
	\end{figure*}

	Figures \ref{cam_1} and \ref{cam_2} shows the Gradient-Weighted Class Activation Maps of original image and adversarial image with respect to correct class and misclassified class for Pixel Attack.
	First column shows the input image to the model. 
    Middle column shows activation maps concerning correctly predicted class $(C)$. 
    Rightmost column shows activation maps concerning misclassified class $(\hat{C})$.

\clearpage

\section{Extra Images of Saliency Maps For Projected Gradient Descent Attack}
\begin{figure*}[!b]
	\centering
		\includegraphics[width=0.48\columnwidth]{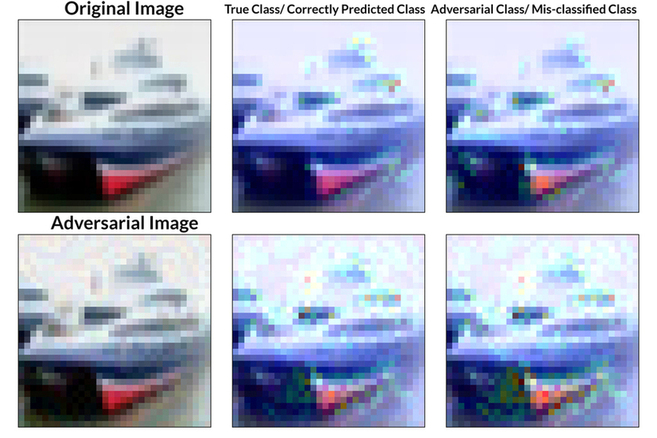} 
		\includegraphics[width=0.48\columnwidth]{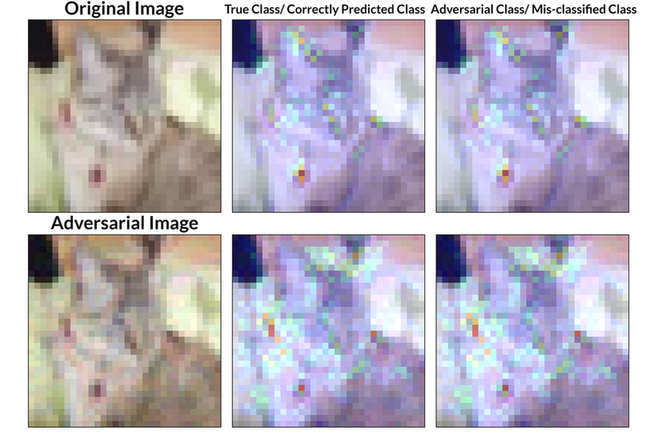} \\
		\includegraphics[width=0.48\columnwidth]{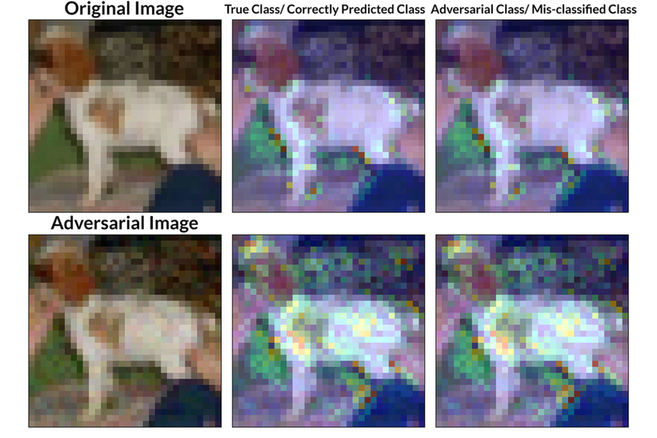}     
		\includegraphics[width=0.48\columnwidth]{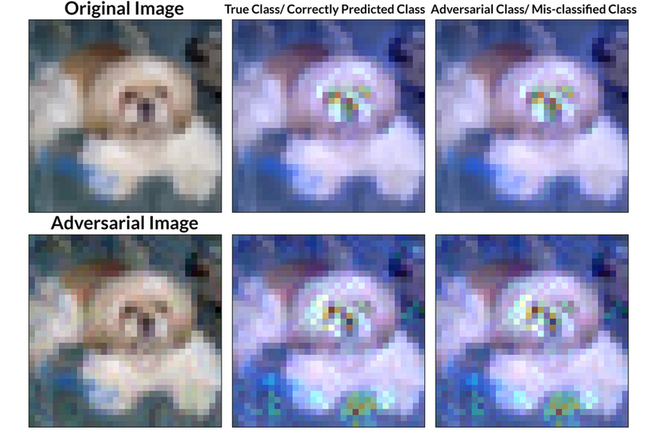} \\
	    \includegraphics[width=0.48\columnwidth]{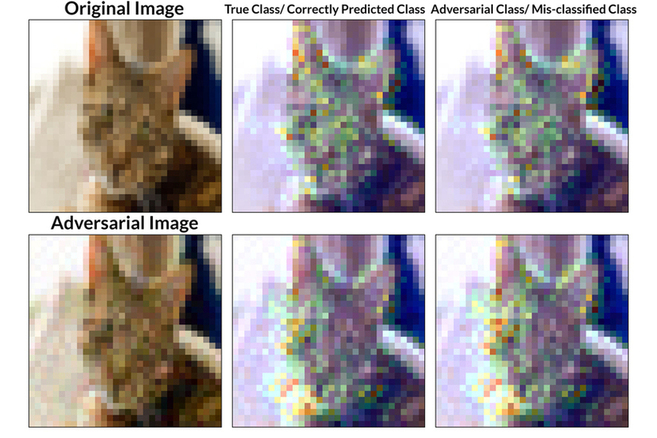}
	    \includegraphics[width=0.48\columnwidth]{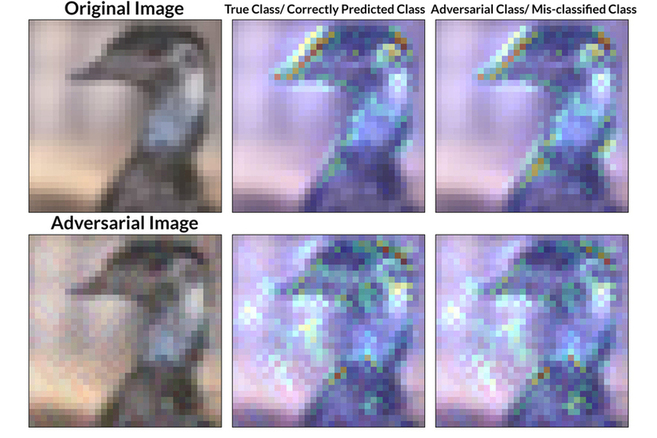} \\
	\caption{
	Overlayed Saliency Maps with respect to correctly predicted class and the misclassified class of Original Image and Adversarial Image generated by Projected Gradient Descent Attack. 
	}
	\label{saliency_3}
	\end{figure*}

\begin{figure*}[!t]
	\centering
		\includegraphics[width=0.48\columnwidth]{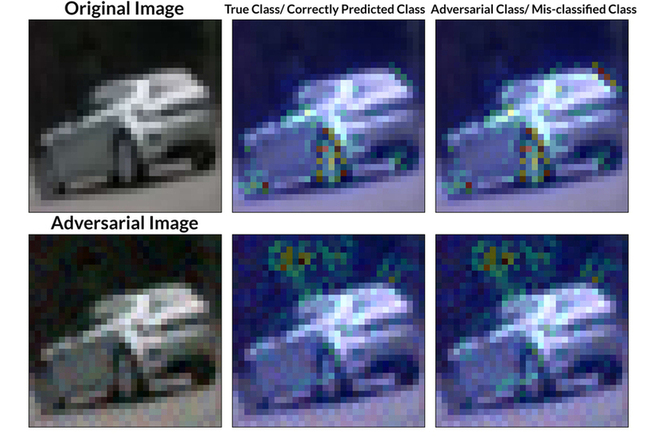}
		\includegraphics[width=0.48\columnwidth]{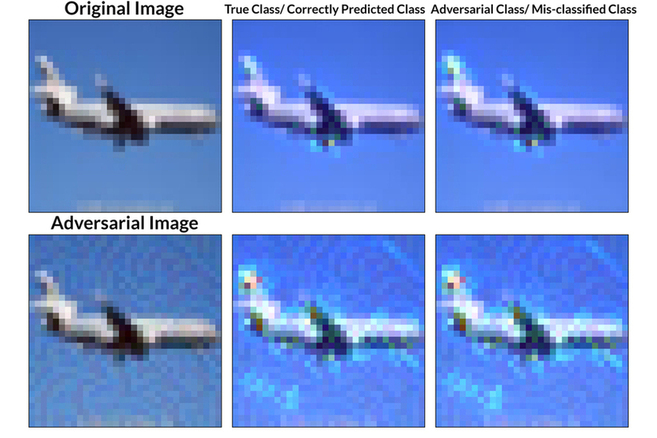}
	\caption{
	Overlayed Saliency Maps with respect to correctly predicted class and the misclassified class of Original Image and Adversarial Image generated by Projected Gradient Descent Attack. 
	}
	\label{saliency_4}
	\end{figure*}

	Figures \ref{saliency_3} and \ref{saliency_4} shows the Saliency Maps of original image and adversarial image with respect to correct class and misclassified class for Projected Gradient Descent Attack.
	First column shows the input image to the model. 
	Middle column shows activation maps concerning correctly predicted class $(C)$. 
	Rightmost column shows activation maps concerning misclassified class $(\hat{C})$.

\clearpage

\section{Extra Images of Gradient-weighted Class Activation Maps For Projected Gradient Descent Attack}
\begin{figure*}[!h]
	\centering
		\includegraphics[width=0.48\columnwidth]{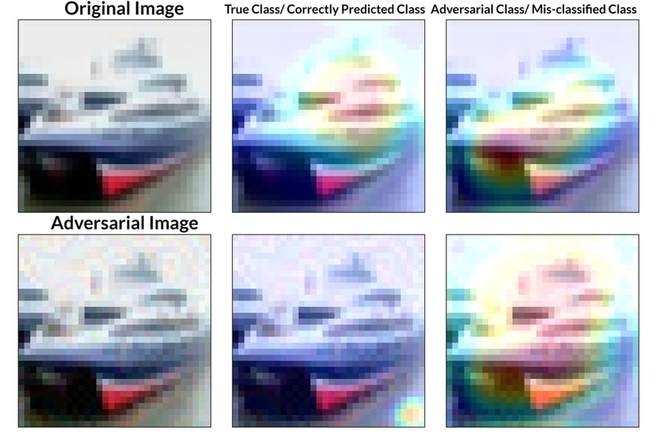} 
		\includegraphics[width=0.48\columnwidth]{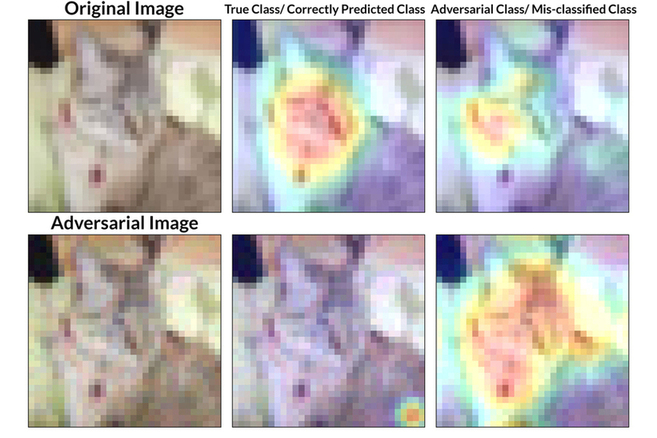}
	\caption{
	Overlayed Gradient-weighted Class Activation Maps with respect to correctly predicted class and the misclassified class of Original Image and Adversarial Image generated by Projected Gradient Descent Attack. 
	}
	\label{cam_3}
	\end{figure*}

\begin{figure*}[!h]
	\centering
		\includegraphics[width=0.48\columnwidth]{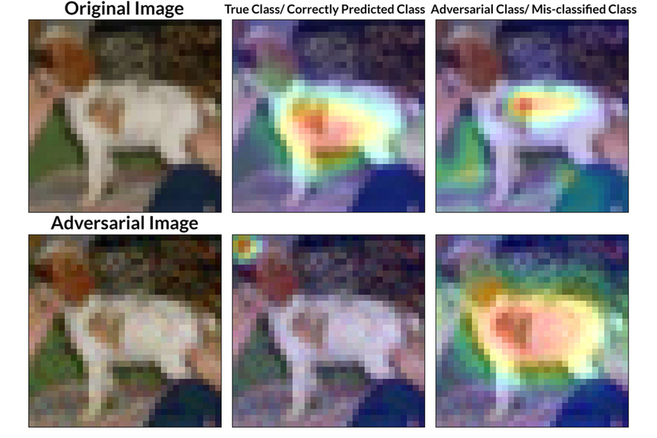}     
		\includegraphics[width=0.48\columnwidth]{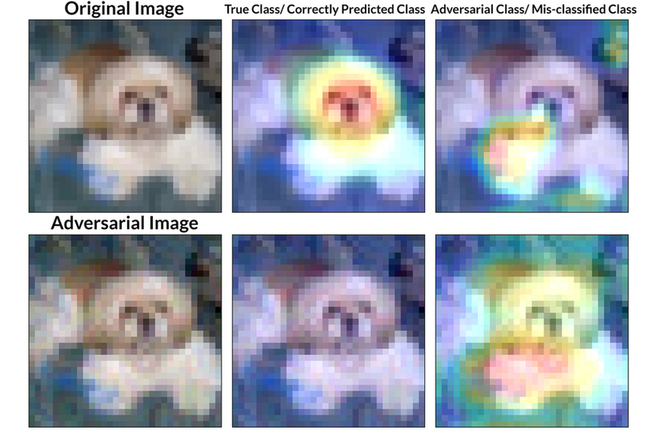} \\
		\includegraphics[width=0.48\columnwidth]{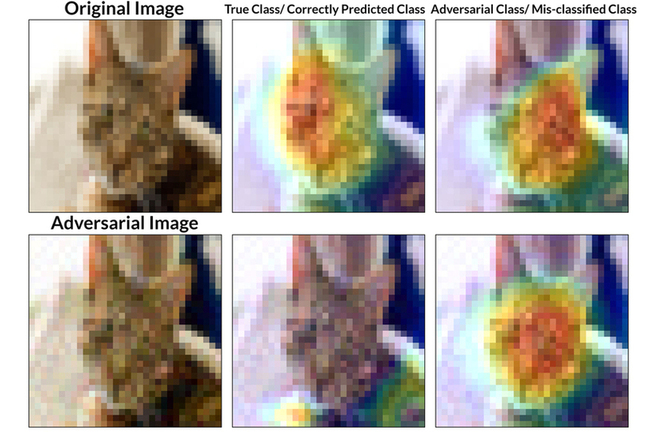}
	    \includegraphics[width=0.48\columnwidth]{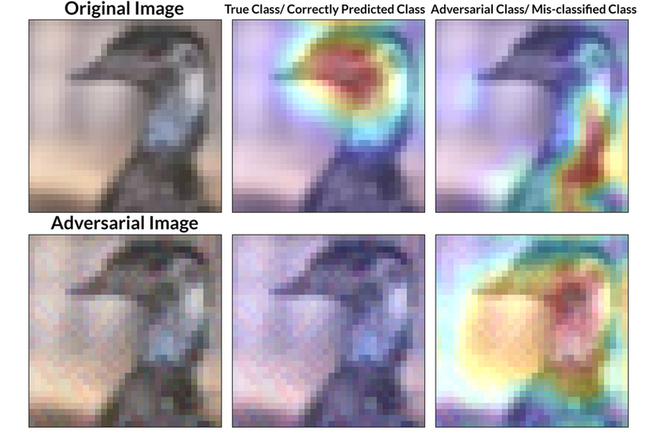} \\
		\includegraphics[width=0.48\columnwidth]{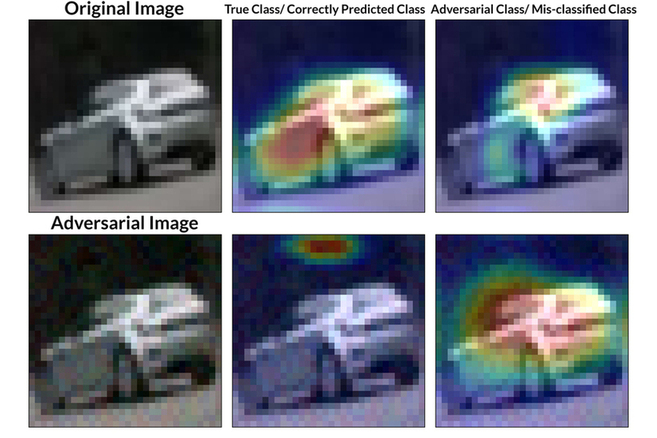} 
		\includegraphics[width=0.48\columnwidth]{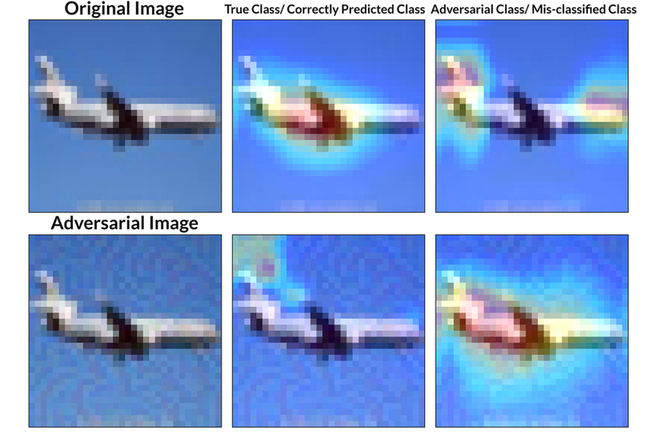}
	\caption{
	Overlayed Gradient-weighted Class Activation Maps with respect to correctly predicted class and the misclassified class of Original Image and Adversarial Image generated by Projected Gradient Descent Attack. 
	}
	\label{cam_4}
	\end{figure*}

	Figures \ref{cam_3} - \ref{cam_4} shows the Gradient-Weighted Class Activation Maps of original image and adversarial image with respect to correct class and misclassified class for Projected Gradient Descent Attack.
	First column shows the input image to the model. 
    Middle column shows activation maps concerning correctly predicted class $(C)$. 
    Rightmost column shows activation maps concerning misclassified class $(\hat{C})$.

\clearpage

\end{document}